\documentclass{article}

\usepackage{microtype}
\usepackage{graphicx}
\usepackage{subcaption}
\usepackage{booktabs}
\usepackage{hyperref}
\usepackage{titletoc}

\usepackage[accepted]{icml2026}
\usepackage{amsmath}
\usepackage{amssymb}
\usepackage{mathtools}
\usepackage{amsthm}
\usepackage[capitalize,noabbrev]{cleveref}
\usepackage{pifont}
\usepackage{enumitem}
\usepackage{xurl}
\usepackage{makecell}
\usepackage{multirow}
\usepackage{tcolorbox}
\tcbuselibrary{breakable}
\theoremstyle{plain}

\theoremstyle{definition}

\theoremstyle{remark}

\usepackage[table]{xcolor}

\definecolor{darksalmon}{rgb}{0.91, 0.59, 0.48}
\definecolor{emerald}{rgb}{0.31, 0.78, 0.47}
\definecolor{green(pigment)}{rgb}{0.0, 0.65, 0.31}
\definecolor{amaranth}{rgb}{0.9, 0.17, 0.31}
\definecolor{iris}{rgb}{0.35, 0.31, 0.81}
\definecolor{uu}{rgb}{0.95, 0.51, 0.51}
\definecolor{spirodiscoball}{rgb}{0.06, 0.75, 0.99}

\icmltitlerunning{AtelierEval}
\begin{document}

\twocolumn[
  \icmltitle{\texttt{AtelierEval}: Agentic Evaluation of Humans \& LLMs as \\Text-to-Image Prompters}

  \icmlsetsymbol{equal}{*}

  \begin{icmlauthorlist}
    \icmlauthor{Hanjun Luo}{nyuad,equal}
    \icmlauthor{Zhimu Huang}{nyuad,equal}
    \icmlauthor{Sylvia Chung}{zju}
    \icmlauthor{Yiran Wang}{uestc}
    \icmlauthor{Yingbin Jin}{polyu}
    \icmlauthor{Jialin Li}{nyuad}
    \icmlauthor{Jiang Li}{nyuad}
    \icmlauthor{Xinfeng Li}{ntu}
    \icmlauthor{Hanan Salam}{nyuad}
  \end{icmlauthorlist}

  \icmlaffiliation{nyuad}{New York University Abu Dhabi}
  \icmlaffiliation{ntu}{Nanyang Technological University}
  \icmlaffiliation{zju}{Zhejiang University}
  \icmlaffiliation{uestc}{University of Electronic Science and Technology of China}
  \icmlaffiliation{polyu}{The Hong Kong Polytechnic University}
  \icmlcorrespondingauthor{Hanjun Luo}{hl6266@nyu.edu}

  \icmlkeywords{Machine Learning, ICML}

  \vskip 0.3in
]

\printAffiliationsAndNotice{\icmlEqualContribution}

\begin{abstract}

Text-to-image (T2I) systems increasingly rely on upstream \emph{prompters}, either humans or multimodal large language models (MLLMs), to translate user intent into detailed prompts. Yet current benchmarks fix the prompt and only evaluate T2I models, leaving the \emph{prompting proficiency} of this upstream component entirely unmeasured. We introduce \textbf{\texttt{AtelierEval}}, the first unified benchmark that quantifies prompting proficiency across 360 expert-crafted tasks. Grounded in a cognitive view, it spans three task categories and instantiates tasks using a taxonomy of real-world challenges, with interfaces for both humans and MLLMs. To enable scalable and reliable evaluation, we propose AtelierJudge, a skill‑based, memory‑augmented agentic evaluator. It produces subjective and objective scores for prompt–image pairs, achieving a Spearman correlation of \textbf{0.81} with human experts, approaching human performance. Extensive experiments benchmark \textbf{8} MLLMs against \textbf{48} human users across \textbf{4} T2I backends, validate \textbf{\texttt{AtelierEval}} as a robust diagnostic tool, and reveal the superiority of mimicry over planning, advocating for an image-augmented direction for future prompters. Our work is \href{https://github.com/Astarojth/AtelierEval}{\textcolor{cyan}{released}} to support future research.

\end{abstract}

\section{Introduction}

The rapid advancement of text-to-image (T2I) models is reshaping creative workflows, enabling the efficient production of complex visual content from commercial illustrations to academic diagrams \cite{ko2023large,esser2024scaling,jaiprakash2025exploring,yang2025text,sordo2025review}. As generation quality and controllability improve, T2I systems increasingly depend on the \textbf{Prompting Proficiency} of upstream \emph{prompters}, i.e., the ability to transform user intent into executable prompts that reliably produce desired outputs \cite{liu2022design,canossa2025algorithmic}.

{
\setlength{\intextsep}{5pt}
\setlength{\belowcaptionskip}{-5pt}
\begin{figure}[H]
    \centering
    \includegraphics[width=\columnwidth]{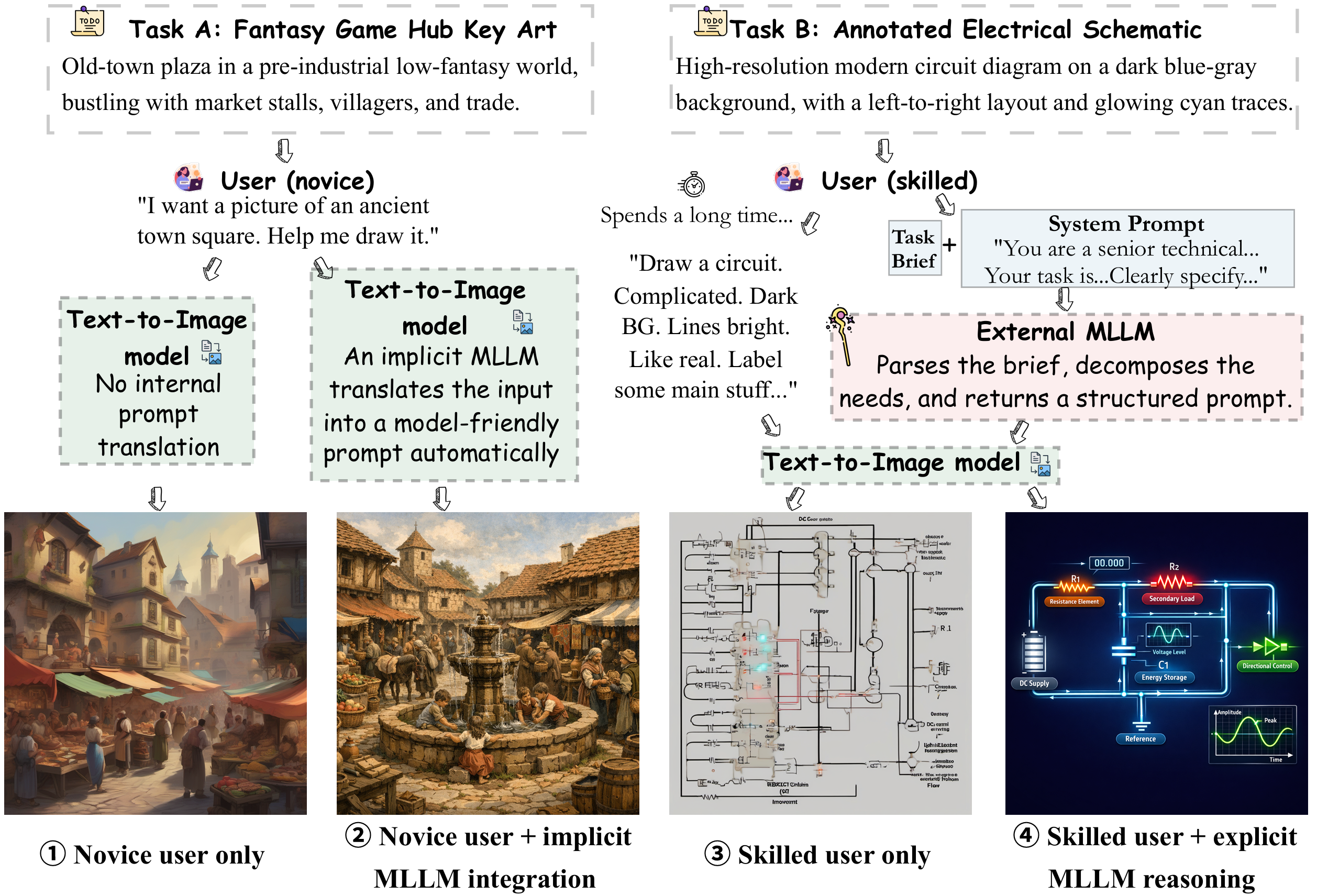}
    \caption{MLLMs act as prompters in diverse T2I workflows, translating user intent into effective prompts.}
    \label{fig:compare}
\end{figure}
}

This proficiency remains an important bottleneck in practice, as effective prompting requires substantial expertise to encode semantics, constraints, and stylistic intent in a single prompt \cite{cao2023beautifulprompt}. In response, contemporary T2I workflows rely on multimodal large language models (MLLMs) to support human prompters. These workflows typically follow two integration patterns as illustrated in Figure~\ref{fig:compare}: (i) Leading platforms (e.g., ChatGPT \cite{openai2025gptimage15}, Gemini \cite{citron2024new}, and Doubao \cite{doubao2025texttoimage}) adopt an \emph{implicit integration} pattern, embedding an internal MLLM middleware to automatically translate user inputs into model-friendly prompts. (ii) Advanced creators employ \emph{explicit reasoning} workflows, where state-of-the-art (SOTA) MLLMs are utilized as external prompting assistants \cite{gani2023llm,xiang2025promptsculptor}. By employing a general-purpose yet carefully designed 
system prompt, they leverage the visual grounding and reasoning capabilities of MLLMs to decompose complex visual intent into detailed, parameterized prompts that meet production-level needs.

Prompting-proficiency evaluation targets the input end of T2I pipelines and is complementary to conventional benchmarks that assess generative models given fixed prompts. Nonetheless, existing evaluation practices have not kept pace with this evolution: there is still no standardized way to quantify the prompting proficiency of humans and MLLMs \cite{gu2023systematic,schulhoff2025prompt}. Research on human prompting remains confined to fragmented qualitative user studies \cite{sanchez2023examining,holzner2025generative,heigl2025generative}, while standardized benchmarks focus solely on T2I models \cite{bakr2023hrs,zhang2024text}, systematically overlooking upstream instruction builders. This evaluation gap can significantly mislead research on prompt optimizers. Typically validated on T2I benchmarks that presuppose executable inputs, these methods prioritize model-specific prompt polishing for visual quality or alignment \cite{pavlichenko2023best,yeh2024tipo,you5178787enhancing}, leaving prompting proficiency—the model-agnostic translation from intent to executable prompts—largely unexplored. This gap represents a key bottleneck for the democratization of T2I systems \cite{mahdavi2024ai} and for recognizing the contributions of prompters \cite{chong2025prompting}, yet remains unmeasured, limiting the advancement of prompting, both as a human skill and as a scalable technology \cite{hartwig2025survey}.

In light of this gap, we introduce \textbf{\texttt{AtelierEval}}, \emph{the first} unified benchmark to evaluate both humans and LLMs quantitatively in their role as prompters. Inspired by Structure of Intellect (SI) theory \cite{guilford1967nature}, we operationalize 360 expert-crafted tasks into three cognitive categories across two modalities: \textit{Open-ended}, \textit{Constrained}, and \textit{Imitation}. Drawing on prior analyses \cite{huang2023t2i,chen2025r2ibench}, we taxonomize common T2I generation failures into semantic interpretation and constraint realization, yielding ecologically grounded, execution-challenging tasks to expose prompting proficiency. Furthermore, \textbf{\texttt{AtelierEval}} offers a unified protocol with an intuitive user interface (UI) for humans and a standardized toolkit for MLLMs, enabling rigorous comparisons between humans and MLLMs in the T2I domain. To empower \textbf{\texttt{AtelierEval}} with scalable precision, we propose AtelierJudge, a cognitive-mimetic Agent-as-a-Judge system orchestrating modular skills and 5 task-specific memories. Grounded in Dual-Process Theory \cite{kahneman2011thinking}, the agent routes skills to decouple evaluation into two parallel processes: \ding{182} subjective scoring via retrieval-augmented generation (RAG) \cite{lewis2020retrieval}, and \ding{183} objective constraint checks. This synergy of objective logic and calibration via human-curated exemplars not only mitigates self-preference bias but also architecturally mirrors human cognition. Empirically, AtelierJudge achieves a Spearman correlation of \textbf{0.81}, nearing the human agreement (\textbf{0.83}) while surpassing baselines (\textbf{0.55}) in subjective scoring, along with a robust \textbf{95.5\%} accuracy on objective constraints.

Leveraging \textbf{\texttt{AtelierEval}}, we benchmark a stratified spectrum of \textbf{8} MLLMs, ranging from lightweight open-source models to SOTA models, against \textbf{48} users (24 novices, 24 skilled users). Prompters are evaluated across \textbf{4} T2I models, representing varying levels of LLM middleware intervention. The results show \textbf{\texttt{AtelierEval}} effectively quantifies the model-agnostic prompting proficiency gaps between different tiers of humans and MLLMs, as evidenced by consistent prompter rankings across all evaluated T2I backends. Moreover, our experiments reveal several key insights. In particular, while advanced T2I backends with MLLM middleware tend to homogenize subjective quality across prompters, they also induce logical conflicts with external MLLM reasoning in several settings, leading to objective performance degradation. In contrast, imitation prompting consistently avoids these conflicts, motivating a shift from symbolic planning to image-augmented prompting for the development of future agents. Our core contributions are summarized as follows:

\begin{itemize}[
    leftmargin=*,     
    topsep=-0.5em,       
    partopsep=0pt,    
    parsep=0pt,      
    itemsep=0pt       
]
    \item[\ding{182}] \textbf{\textit{Unified Benchmark.}} We introduce \textbf{\texttt{AtelierEval}}, the first systematic benchmark to quantify T2I prompting proficiency. Grounded in cognitive science, it comprises 360 expert-crafted tasks spanning modalities and features a unified interaction protocol for both humans and MLLMs.
    
    \item[\ding{183}] \textbf{\textit{Agentic Evaluator.}} We propose AtelierJudge, a cognitive-mimetic and skill-based agent with memory for retrieval-augmented evaluation. It achieves superior alignment with human experts compared to baselines, offering a principled blueprint for future T2I evaluators.

    \item[\ding{184}] \textbf{\textit{Validation \& Insights.}} We validate our framework's effectiveness in evaluating prompting proficiency. We also reveal key insights and provide open-source infrastructure, facilitating both prompt engineering education and the development of prompting agents.
\end{itemize}

\section{Related Work}

\textbf{Prompting Proficiency Benchmarking.}
Current T2I benchmarks offer comprehensive coverage, ranging from general quality and alignment \cite{saharia2022photorealistic,wu2023human} through complex metrics like text rendering and spatial relationships \cite{bosheah2025challenges} to specific applications \cite{chang2025sridbench,tao2025designweaver} and safety metrics \cite{luo2024faintbench,luo2024bigbench,luo2026biasig,chinchure2025tibet}, yet they primarily focus on T2I models themselves. By treating the inputs as static prompts, they systematically overlook the role of prompters. Although numerous prompt optimizers have emerged, their validation often directly applies the aforementioned T2I benchmarks \cite{hao2023optimizing,manas2024opt2i,zhang2024ak4prompts,luo2024versusdebias}. This evaluation paradigm confines prompting proficiency to model-specific, metric-specific, and fragmented parametric adaptation \cite{mo2024dynamic,li2024promptist}, rather than a generalized request-to-prompt translation capability. While research on human prompting has evolved from simple prompt refinement \cite{feng2023promptmagician,hei2024user,wang2024promptcharm} to analyzing users' request-to-prompt performance \cite{liu2025tamingtexttoimagesynthesisnovices,huang2025promptnavi}, this body of work remains confined to small-scale and qualitative studies \cite{tsao2025genai,gulzar2025designeducation}. Consequently, this gap urgently calls for a standardized benchmark that can isolate prompting proficiency from model execution and support systematic comparison across prompters.

\textbf{Agentic Evaluation Paradigms.}
T2I evaluation has long relied on automated metrics such as CLIP Score, with corresponding evaluators \cite{heusel2017gans,hessel2021clipscore}. However, recent studies indicate that these coarse-grained metrics exhibit poor correlation with human perception, particularly for complex spatial structures, logical consistency, and subtle aesthetic nuances \cite{ghosh2023geneval,li2024genaibench}. This discrepancy has compelled the community to resort to unscalable and unreproducible manual evaluation \cite{lee2023holistic,wiles2025revisiting}. In response, the \textit{LLM/MLLM-as-a-Judge} paradigm—leveraging LLMs to simulate human judgment—has been widely adopted \cite{zheng2023judging,li2024llms}. Recent advancements have introduced the \textit{Agent-as-a-Judge}, which significantly enhances evaluation performance \cite{zhuge2024agent}. Nevertheless, this shift remains underexplored in the T2I domain. Current MLLM-based scorers predominantly perform static visual question answering (VQA) \cite{hu2023tifa,sun2025t2ireasonbench}. Not only do they struggle with the dual challenge of ``objective logical constraints'' and ``subjective aesthetic appreciation'' when evaluating prompts \cite{fu2024commonsenset2i,jin2025compose}, but they are also prone to severe model biases and self-preferences \cite{chen2024mllmasajudge}.

\section{\textbf{\texttt{AtelierEval}}: Benchmark Design}

For systematic evaluation, \textbf{\texttt{AtelierEval}} operationalizes prompting proficiency as a structured cognitive process. Section \ref{sec:problem} formalizes the prompter-model interaction as an optimization problem, defining prompting proficiency rigorously. Drawing on Guilford’s SI theory, Section \ref{sec:categories} dissects this process into three distinct cognitive categories to ensure comprehensive capability coverage. Section \ref{sec:dataset} details the construction of our 360-task dataset based on taxonomic challenge primitives. Section \ref{sec:protocol} establishes a unified protocol to benchmark humans and MLLMs.

\subsection{Problem Formulation}
\label{sec:problem}

Based on established research on prompt engineering \cite{zamfirescu2023johnny}, we conceptualize prompting as the process of bridging the gap between user intent and executable system actions (\textit{Gulf of Execution}) \cite{norman2013design}. We formally distinguish prompting proficiency from existing evaluation paradigms by modeling the generation process $x = \mathcal{M}(p)$ with two distinct alignment targets: the \textbf{Intent Representation} $I$ (the user's ground truth goal in a pre-execution form) and the \textbf{Literal Intent} $I_p$ (the explicit semantics and implied aesthetics of prompt $p$). Notably, $I$ may be abstract, structured, or perceptual, provided that it is not directly executable by T2I models. Let $\pi: I \rightarrow p$ be the prompter's policy and $\mathcal{M}$ be the T2I model.

{
\setlength{\intextsep}{3pt}
\setlength{\belowcaptionskip}{-5pt}
\begin{figure}[H]
    \centering
    \includegraphics[width=0.9\columnwidth]{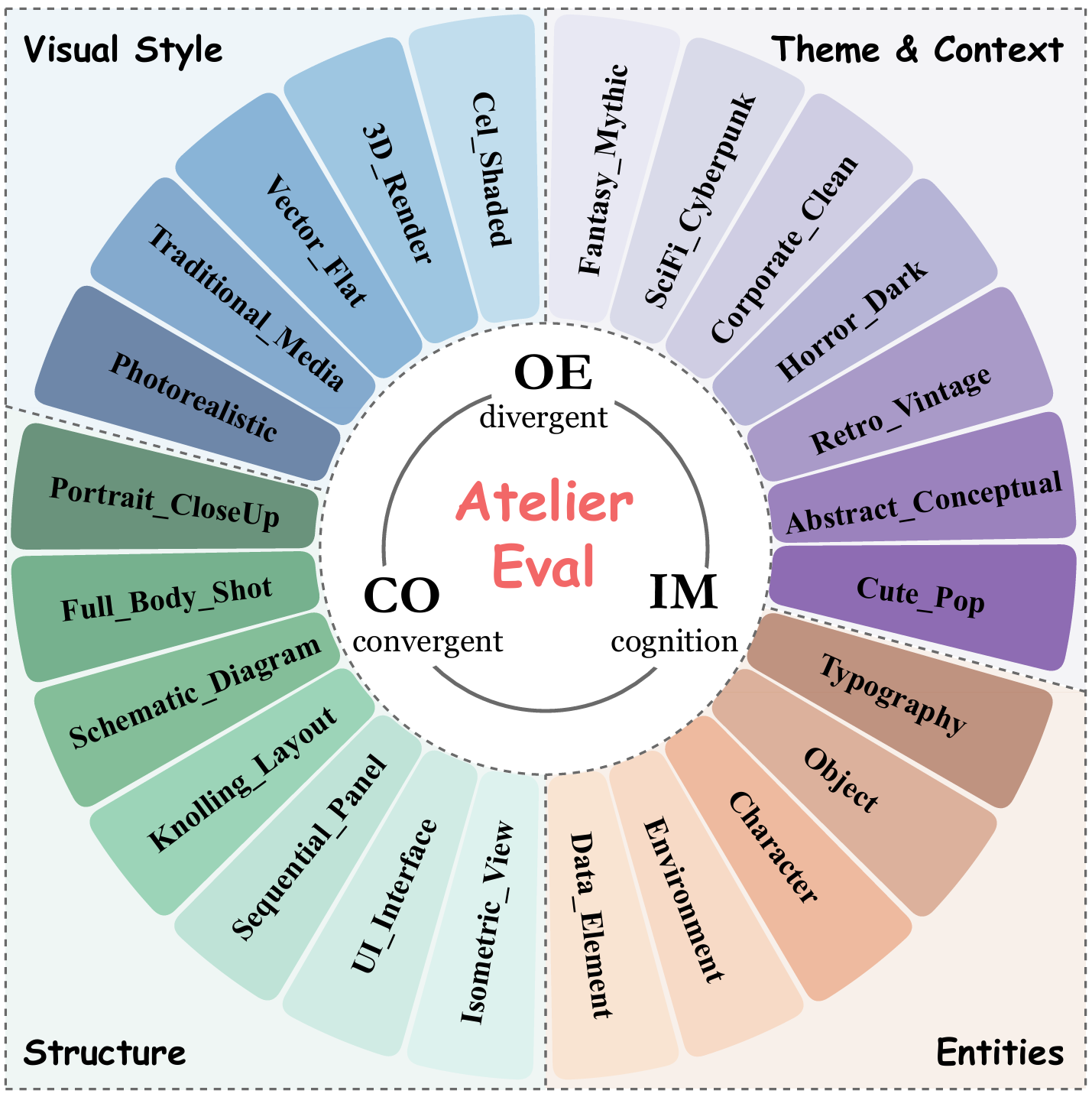}
    \caption{Three cognitively grounded categories form a complete task partition across 4 application context dimensions and 24 tags. Details of the application contexts are presented in Appendix \ref{app:category}.}
    \label{fig:stat}
\end{figure}
}

\textbf{Paradigm 1: Model Benchmarking.}
Mainstream benchmarks fix $p$ to evaluate the generative capabilities of a set of candidate models $\mathbb{M}$. They assume $p$ is the ground truth, defining the optimal model $\mathcal{M}^* \in \mathbb{M}$ that maximizes the expected score $S$ over a fixed prompt distribution $\mathcal{D}$:
{
\setlength{\abovedisplayskip}{3pt} 
\setlength{\belowdisplayskip}{0pt}
\begin{equation}
\label{eq:model_bench}
    \mathcal{M}^* = \operatorname*{argmax}_{\mathcal{M}} \mathbb{E}_{p \in \mathcal{D}} \left[ S(\mathcal{M}(p), I_p) \right],
\end{equation}
}

\textbf{Paradigm 2: Prompt Optimization.}
Optimizers (e.g., token searchers, LLM-based rewriters) refine an initial prompt $p_{init}$ into an optimized $p^*$ within a semantic neighborhood $\mathcal{N}(p_{init})$ to boost generation quality or alignment:
{
\setlength{\abovedisplayskip}{3pt} 
\setlength{\belowdisplayskip}{3pt}
\begin{equation}
\label{eq:prompt_opt}
    p^* = \operatorname*{argmax}_{p \in \mathcal{N}(p_{init})} S(\mathcal{M}(p), I).
\end{equation}
}This \textit{utility-oriented} paradigm prioritizes instance-specific maximizing $S$, reduces prompting to polishing the explicit $p_{init}$, and neglects the fundamental translation of $I$.

\textbf{Paradigm 3: Prompting Proficiency.}
Unlike Paradigm 2, the paradigm of prompting proficiency is \textit{capability-oriented}. We evaluate the prompter's policy $\pi$ itself. We fix the distribution of intents $\mathcal{D}$ and evaluate $\pi$'s ability to structurally decode $I$ across varying models:
 {
\setlength{\abovedisplayskip}{3pt} 
\setlength{\belowdisplayskip}{3pt}
\begin{equation}
\label{eq:prompting_prof}
    \pi^* = \operatorname*{argmax}_{\pi} \mathbb{E}_{I \in \mathcal{D}} \left[ \mathbb{E}_{\mathcal{M}} [ S(\mathcal{M}(\pi(I)), I) ] \right].
\end{equation}
}Prompting proficiency aims to quantify the prompter's ability to generalize, translate, and adhere to intents, establishing a metric for the prompter's intrinsic proficiency.

\subsection{Task Categorization \& Cognitive Mapping}
\label{sec:categories}

Equation \eqref{eq:prompting_prof} provides a principled way to evaluate prompters. However, our goal is a systematic and diagnostic evaluation that avoids reliance on fragmented, ad-hoc task collections and reveals the specific capability dimensions that distinguish different prompters, thereby informing both human prompting education and prompting agent development. Currently, prompt engineering is largely treated as a trial-and-error process, with evaluation confined to overall outcome quality \cite{strobelt2022interactive,knoth2024ai}. To move beyond such black-box evaluation, a structured decomposition of the policy $\pi$ is necessary. To prevent ad-hoc or arbitrary decompositions, we adopt the \emph{Operations} dimension of SI theory and map all five operations to our framework. Specifically, the analysis-oriented \emph{memory} and \emph{evaluation} operations are delegated to AtelierJudge (Section~\ref{sec:judge}), while task design focuses on the three constructive operations—\emph{divergent production}, \emph{convergent production}, and \emph{cognition}. Accordingly, as shown in Figure \ref{fig:stat}, we partition the task space into the following three categories:

\begin{itemize}[leftmargin=*,
    topsep=-0.5em,
    partopsep=0pt,
    parsep=0pt,
    itemsep=0pt
    ]

\item[\ding{224}] \textbf{Open-ended Creation (OE)}
tasks correspond to \emph{divergent production} and focus on translating abstract, scenario-driven natural language requests—typically information-sparse and containing narrative noise—into executable prompts. These tasks emphasize thematic, atmospheric, and stylistic intent without providing explicit execution constraints. Prompters must extract key information, complete and organize elements, and translate implicit intent into a complete prompt.

\item[\ding{224}] \textbf{Constrained Creation (CO)}
tasks correspond to \emph{convergent production} and focus on constructing prompts under explicit rules. Such tasks provide structured constraints across multiple dimensions, which are semantically clear yet difficult to execute directly. Prompters must jointly integrate complex constraints into a single prompt that is more likely to satisfy all conditions during generation.

\item[\ding{224}] \textbf{Imitation (IM)}
tasks correspond to \emph{cognition}. Given a target image, prompters identify key features and translate perceptual information into prompts to reproduce similar visual results. These tasks reflect the information encoding and representation processes involved in \emph{cognition}.
\end{itemize}

Notably, \textbf{\texttt{AtelierEval}} is restricted to single-turn, pure text-to-image settings. From an information-theoretic perspective \cite{cover1999elements}, OE, CO, and IM correspond to expansion, convergence, and fidelity-preserving encoding, respectively, forming a complete task partition for this setting. Formal justification is provided in Appendix \ref{app:justify}.

\subsection{Dataset Construction}
\label{sec:dataset}

To focus our evaluation on prompting proficiency, tasks are designed to involve intents difficult for the model to satisfy directly, but whose satisfaction probability can be improved through effective translation. Drawing on prior studies \cite{ribeiro2020checklist,oppenlaender2024taxonomy}, we abstract failure modes from such intents into two dimensions, semantic interpretation $\{S_i\}$ and constraint realization $\{C_j\}$, and define two corresponding sets of \emph{challenge primitives} for task construction (Table \ref{tab:challenge_primitives}). Tasks are instantiated by experts via cross-combining subsets of these primitives and grounding them in common T2I application contexts, with task complexity emerging from their composition. The details of task construction process and expert involvement are provided in Appendix \ref{app:data_collect} \& \ref{app:experts}. The three task categories are instantiated as follows, each with 120 tasks:

\begin{itemize}[leftmargin=*,
    topsep=-0.5em,
    partopsep=0pt,
    parsep=0pt,
    itemsep=0pt
    ]

\item[\ding{224}] \textbf{OE Tasks}
 instantiate $\{S_i\}$ under high narrative noise. These tasks focus on semantic translation, without introducing explicit $\{C_j\}$.

\item[\ding{224}] \textbf{CO Tasks}
emphasize $\{C_j\}$ under low-noise, structured specifications. $\{S_i\}$ are still present but expressed explicitly with limited ambiguity to provide constraint context.

\item[\ding{224}] \textbf{IM Tasks}
instantiate visual counterparts of challenge primitives through image inputs. Primitives that cannot be reliably inferred from images, including $S_2$, $S_4$, and $C_5$, are excluded as stated in Appendix \ref{apps:data_im}.

\end{itemize}

{
\setlength{\intextsep}{8pt}
\begin{table}[H]
\centering
\caption{Challenge Primitives of \textbf{\texttt{AtelierEval}}. 
Primitive decomposition and task instances are provided in Appendix~\ref{app:task_instances}.}
\label{tab:challenge_primitives}
\resizebox{\linewidth}{!}{%
\begin{tabular}{@{}ll@{}}
\toprule
\textbf{Primitive} & \textbf{Definition} \\ \midrule

\multicolumn{2}{@{}l}{\textit{Semantic Challenge Primitives}} \\[2pt]
\quad $\mathbf{S_1}$ \textbf{Abstract Intent} 
& Abstract or affective intent requiring visualization. \\
\quad $\mathbf{S_2}$ \textbf{Audience Intent} 
& Intent specified via target audience preferences. \\
\quad $\mathbf{S_3}$ \textbf{Implicit Style} 
& Style or medium implied but not explicitly stated. \\
\quad $\mathbf{S_4}$ \textbf{Semantic Negation} 
& What should not be generated at the semantic level. \\[4pt]
\midrule
\multicolumn{2}{@{}l}{\textit{Constraint Challenge Primitives}} \\[2pt]
\quad $\mathbf{C_1}$ \textbf{Attribute Binding} 
& Binding attributes to the correct entities. \\
\quad $\mathbf{C_2}$ \textbf{Spatial Relation} 
& Explicit spatial or layout relationships. \\
\quad $\mathbf{C_3}$ \textbf{Quantity} 
& Exact numerical constraints on object counts. \\
\quad $\mathbf{C_4}$ \textbf{Text} 
& Exact text content and spelling. \\
\quad $\mathbf{C_5}$ \textbf{Hard Constraint} 
& Global, non-relaxable constraints on generation. \\

\bottomrule
\end{tabular}%
}
\end{table}
}

{
\setlength{\dbltextfloatsep}{3pt}
\setlength{\belowcaptionskip}{-10pt}
\begin{figure*}[t]
\centering
\includegraphics[width=0.99\textwidth]{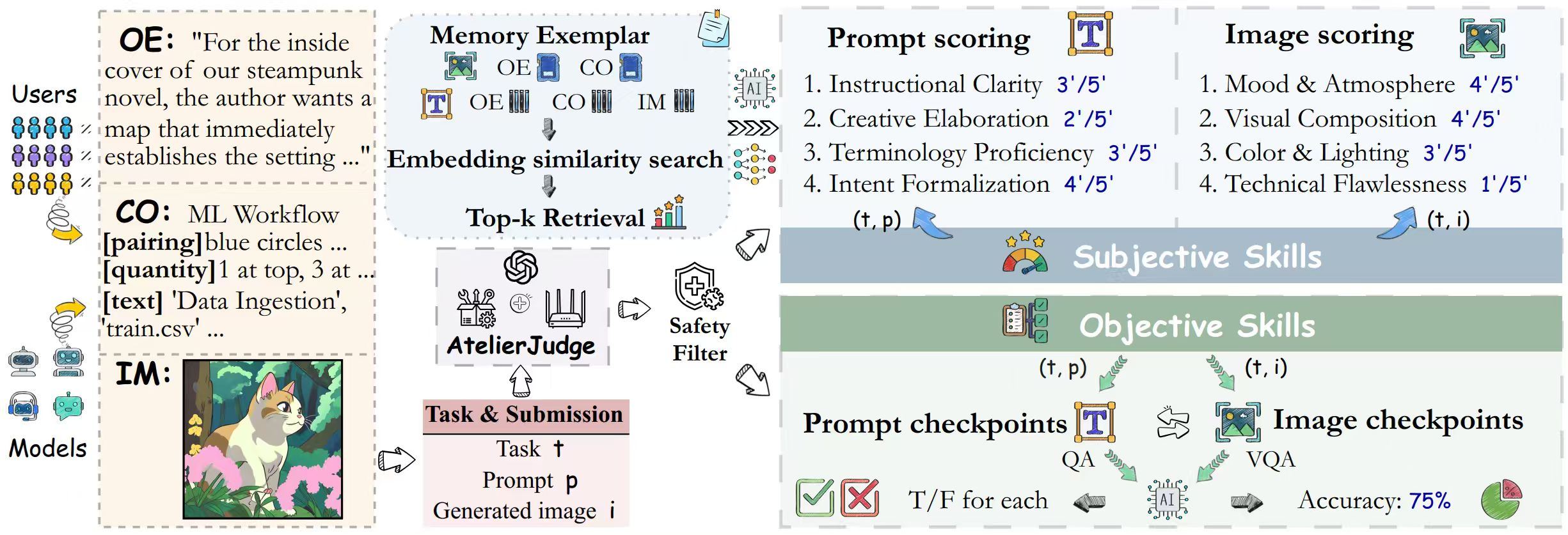}
\caption{Illustration of how AtelierJudge works in \textbf{\texttt{AtelierEval}}. It decouples evaluation into two parallel processes independently applied to prompt and image: a \emph{subjective} branch that performs memory-augmented quality evaluation, and an \emph{objective} branch to verify constraint adherence. Both processes are executed through a skill library, from which the evaluator selects a task-conditioned sequence.}
\label{fig:arch}
\end{figure*}
}

\subsection{Unified Interaction Protocol}
\label{sec:protocol}

We design a unified and comparable interaction protocol for humans and MLLMs to evaluate their prompting policy $\pi: I \rightarrow p$ under controlled conditions. The protocol adopts a single-turn, text-only input paradigm (stated in Section \ref{sec:categories}) without immediate generation feedback to isolate and measure the prompter’s ability to translate $I$ into $p$ in one decision, excluding feedback-driven iterative optimization. By uniformly constraining both humans and MLLMs, the protocol enables a direct and comparable assessment of prompting proficiency. At the implementation level, we provide dual interfaces for humans and MLLMs with identical semantics. The UI follows the interaction pattern of SD-WebUI \cite{AUTOMATIC1111StableDiffusionWebUI}, a widely adopted interface in current T2I tools, and is implemented as a web interface built with Gradio \cite{abid2019gradio} and deployed on HF Spaces \cite{wolf2020transformers}. The interface is intentionally kept minimal, retaining only the task description and text input areas to avoid introducing extraneous variables. Correspondingly, MLLMs receive the same task specification through a standardized API and output prompts. Prompt--image pairs generated from both interfaces are processed through the same pipeline to ensure fairness and reproducibility. Complete workflow and examples are provided in Appendix \ref{app:assessment_platform}.

\section{AtelierJudge: Agentic Evaluation}
\label{sec:judge}

\textbf{Dual-Process Theory.}
Evaluating prompting proficiency inherently involves both assessing subjective aspects (e.g., aesthetics) and verifying explicit task requirement satisfaction. These two aspects are complementary but require fundamentally different evaluation mechanisms. To coherently integrate these distinct aspects into a unified evaluation framework, we draw inspiration from \emph{Dual-Process Theory}, which characterizes reasoning as comprising two complementary systems, commonly referred to as \emph{System~1} (S1) and \emph{System~2} (S2). Conceptually, S1 is associated with holistic, experience-driven judgment, whereas S2 emphasizes analytic reasoning over explicit structures and conditions.

\textbf{Our Design.}
In our setting, these systems are adopted as a functional abstraction. S1 supports subjective assessment by grounding judgments in gold exemplar memory. S2 supports objective verification by decomposing task alignment into explicit, independent checkpoints. This formulation provides a principled explanation for our design: perceptual impressions about quality are stabilized through memory-based calibration, while analytic decomposition prevents individual constraint violations from being influenced by overall judgments, consistent with established dual-process accounts of evaluation in cognitive science \cite{evans2008dual,evans2013dual}. Based on this abstraction, as illustrated in Figure~\ref{fig:arch}, AtelierJudge adopts a modular skill library that operationalizes the separation between subjective and objective evaluation. Each skill encapsulates a focused evaluation dimension, operates over either the prompt or the generated image, and is flexibly composed to support different task categories. This structured library enables an agentic evaluation process that systematically composes subjective and objective skills across the prompt and image modalities, resulting in comprehensive and interpretable evaluation of prompting proficiency.

\subsection{System 1: Subjective Evaluation}

AtelierJudge executes subjective evaluation skills for degree-based assessment to characterize the perceptual quality of submitted results. The design draws from the memory-augmented evaluation paradigm, shown to improve LLM judgment consistency and human alignment in complex subjective tasks~\cite{luo2025agentauditor,huang2026rethinkingmemorymechanismsfoundation}.

The subjective evaluation is implemented as five modular skills, each paired with an expert-curated exemplar memory for calibration: three prompt-level skills across all task categories, and two image-level skills for OE and CO tasks. Subjective evaluation for images is not applied for IM tasks which aims to reproduce a target image. Each memory contains 120 representative exemplars spanning multiple tasks and quality levels. All exemplars are annotated with a 1–5 Likert score \cite{likert1932technique} for each subjective dimension, together with a brief rationale justifying the assigned score. Prompt dimensions include \emph{Instructional Clarity}, \emph{Creative Elaboration}, \emph{Terminology Proficiency}, and \emph{Intent Formalization}. Image dimensions include \emph{Mood \& Atmosphere}, \emph{Visual Composition}, \emph{Color \& Lighting}, and \emph{Technical Flawlessness}. These dimensions are derived from prior studies on text-to-image generation and human perceptual evaluation~\cite{otani2023toward,liang2024rich,chen2025unleashing}, and definitions are detailed in Appendix~\ref{app:subjective_dimensions}. Details of memory construction are described in Appendix~\ref{app:exemplar}.

During evaluation, AtelierJudge first routes each submission to the corresponding skills based on the task category. The prompt and the generated image are then decoupled and evaluated independently. For each modality, the selected skill retrieves the Top-$K$ exemplars from its associated memory based on cosine similarity in modality-specific embedding spaces, with each exemplar annotated with human-curated scores and rationales. Conditioned on the predefined scoring criteria and retrieved exemplars, AtelierJudge then produces scores for each subjective dimension.

\subsection{System 2: Objective Check}
\label{sec: S2}

AtelierJudge executes objective evaluation skills for binary verification, determining whether results satisfy task constraints that are inherently T/F in semantics. For each task, evaluation is formulated as an expert-defined, fixed checklist of constraints. The objective skills apply zero-shot QA/VQA to perform constraint checks separately on the prompt and the generated image. Each constraint corresponds to a pair of checkpoints: a prompt checkpoint for explicit specification of constraints, and an image checkpoint for its actual realization. This paired design decouples the prompter’s intent specification from the model’s execution, enabling fine-grained analysis of prompting proficiency. The sources of task constraints vary across task types. CO task constraints are given as explicit, structured specifications. OE task constraints are derived from descriptions with clear T/F requirements. IM task constraints are derived from perceptible attributes of the target image that admit binary judgment. These checklists evaluate attribute-level semantic consistency rather than pixel-level reconstruction, allowing multiple valid realizations that preserve the target image's key objects, relations, and style cues. Appendix~\ref{app:task_instances} provides representative checklist examples.

\subsection{Skill Routing}

AtelierJudge decomposes evaluation into five composable skills~\cite{zhang2026evoskillsselfevolvingagentskills}, and each skill encapsulates a single scoring or verification logic. In a complete run, AtelierJudge first executes \emph{safety filter skills} (see Appendix~\ref{app:safety} for details) to filter out submissions with severe safety risks. For passed submissions, the system concurrently schedules skills along the subjective and objective branches, and within each branch further executes the corresponding skills on the prompt and image modalities, forming a $2\times2$ parallel evaluation process. All skills execute independently, without shared intermediate state or ordering dependencies. After skill execution, the system aggregates outputs from different skills. Multi-dimensional scores from subjective evaluation skills are directly retained, while binary outputs from objective evaluation skills are accumulated to compute a task-level constraint satisfaction rate in $[0,1]$. Subjective and objective metrics remain strictly decoupled, providing interpretable and orthogonal evaluation signals for further analysis.

\section{Experiments}

\subsection{General Experimental Settings}
\label{sec:exp_set}

\textbf{MLLM Selection.}
We evaluate the prompting proficiency of \textbf{8} MLLMs, selected to form a structured capability spectrum. Models are grouped into three capability tiers ($T_0$–$T_2$), namely SOTA, mid-tier, lightweight, with comparable benchmark performance within each tier to control for capacity effects in analyses \cite{hendrycks2020measuring,yue2024mmmu,rein2024gpqa}. Specifically, $T_0$ includes GPT-5.2 \cite{openai_gpt52_system_card}, Claude-4.5-Sonnet (Cl-4.5) \cite{anthropic2025sonnet45}, and Gemini-3-Pro-preview (Gem-3) \cite{google_gemini3pro_model_card}; $T_1$ includes GPT-4.1 \cite{openai2025gpt41}, Qwen-3-VL-235B-A22B (Qwen-L) \cite{yang2025qwen3technicalreport}, and Gemini-2.0-Flash (Gem-2) \cite{google_gemini20_flash_model_card}; $T_2$ includes GPT-4.1-Nano (GPT-4n) and Qwen-3-VL-8B (Qwen-S). GPT models enable controlled analysis along a single model lineage. Qwen models represent widely adopted open-source MLLMs~\cite{guo2025quantized}. GPT-5.4 \cite{openai_gpt54_system_card}, GPT-5.2, Cl-4.5, and Gem-3 are used for evaluator selection.

\textbf{T2I Backends.}
We evaluate prompting proficiency on \textbf{4} widely applied T2I backends, representing typical patterns in architectures and MLLM involvement: Gemini-3-Pro-Image-Preview (nBanana) \cite{google2025gemini3proimagecard}, GPT-Image-1-All (GI-1) \cite{openai2024gptimage1}, and Flux.1 Pro \cite{labs2025flux1kontextflowmatching} as commercial models; SDXL \cite{podell2023sdxl} as the open-source model. Specifically, GI-1 incorporates an MLLM middleware to rewrite user inputs into structured internal prompts before image generation, while nBanana is built on Gem-3 and supports internal reasoning via undisclosed mechanisms. Flux Pro and SDXL do not include MLLM middleware. Flux Pro adopts a Transformer-based architecture with joint modeling of text conditions, whereas SDXL employs a diffusion-based architecture in which text conditions are injected via a relatively independent encoder \cite{radford2021learning}. As Flux Pro and GI-1 achieve comparable performance on existing benchmarks \cite{huang2025t2i}, we use them as a roughly controlled pair when analyzing the impact of middleware. Collectively, they span the spectrum of text processing paradigms in T2I, enabling holistic analysis of prompting dynamics across diverse systems. Detailed settings for all MLLMs and T2I backends are provided in Appendix~\ref{app:para}.

\textbf{Human Study Design.}
We recruit 48 human participants as prompters into two predefined groups: 24 novices and 24 skilled users (see Appendix~\ref{app:demographic} for criteria and statistics). Each participant completes 30 tasks, with 10 tasks per category, randomly assigned via a balanced Latin square design. Each task is completed twice per group, and scores are averaged at the group level. To reduce ordering effects and fatigue, tasks are organized into 6 randomized rounds, each containing 5 tasks of the same category. Participants interact through our UI under the same protocol as MLLMs (Section~\ref{sec:protocol}), submit only natural-language prompts (structured formats such as lists are allowed), and are not informed of the underlying T2I backend. The use of LLM-based tools is prohibited. Informed consent and participant instructions are provided in Appendix~\ref{app:consent}.

\subsection{Meta-Evaluation of AtelierJudge}
\label{meta_evaluation}

\textbf{Setup.}
We validate AtelierJudge on an expert-annotated set of 360 stratified prompt--image pairs (one per task), detailed in Appendix~\ref{app:valid_set}. Text and image retrieval are implemented with \textit{Nomic-Embed-Text-V1.5} (dim=512) \cite{nussbaum2024nomic} and \textit{DINO-V2-Giant} \cite{oquab2023dinov2}, respectively, with a retrieval count of $K{=}3$ for both modalities. For subjective metrics, we calculate results per subjective dimension and modality, and report the macro-averaged mean absolute error (MAE), Within-1 accuracy (W1-A), and Spearman correlation ($\rho$) \cite{spearman1961proof}. For objective metrics, we report the accuracy (Acc) and F1-score (F1), micro-averaged globally across all checkpoints. Design justifications, including ablations, are detailed in Appendix~ \ref{app:ablation}.

{
\setlength{\belowcaptionskip}{-5pt}
\begin{table}[H]
\centering
\caption{Subjective verification results. Human performance is estimated via leave-one-out cross-validation among three experts}
\label{tab:subj_performance}
\renewcommand{\arraystretch}{1.1}
\setlength{\tabcolsep}{3pt}
\resizebox{0.99\linewidth}{!}{
\begin{tabular}{l|cc|cc|cc}
\toprule

\multirow{2}{*}{\textbf{Model}} 

& \multicolumn{2}{c|}{\textbf{MAE} {\boldmath$\downarrow$}} 
& \multicolumn{2}{c|}{\textbf{W1-A} {\boldmath$\uparrow$}} 
& \multicolumn{2}{c}{{\boldmath$\rho \uparrow$}} \\ 
\cmidrule(lr){2-3} \cmidrule(lr){4-5} \cmidrule(lr){6-7}
 & Base & Ours$_{\Delta (\%)}$ & Base & Ours$_{\Delta (\%)}$ & Base & Ours$_{\Delta (\%)}$ \\
\midrule
GPT-5.4 & 0.74 & $\textbf{0.33}_{\textcolor{red}{\downarrow 55.4}}$ & 0.67 & $\textbf{0.95}_{\textcolor{red}{\uparrow 41.8}}$ & 0.55 & $\textbf{0.81}_{\textcolor{red}{\uparrow 47.3}}$ \\
GPT-5.2 & 0.72 & $0.34_{\textcolor{red}{\downarrow 52.8}}$ & 0.64 & $0.93_{\textcolor{red}{\uparrow 45.3}}$ & 0.56 & $0.79_{\textcolor{red}{\uparrow 41.1}}$ \\
Gem-3 & 0.65 & $0.35_{\textcolor{red}{\downarrow 46.2}}$ & 0.68 & $0.93_{\textcolor{red}{\uparrow 36.8}}$ & 0.51 & $0.77_{\textcolor{red}{\uparrow 51.0}}$ \\
Cl-4.5 & 0.78 & $0.37_{\textcolor{red}{\downarrow 52.6}}$ & 0.61 & $0.91_{\textcolor{red}{\uparrow 49.2}}$ & 0.48 & $0.73_{\textcolor{red}{\uparrow 52.1}}$ \\
\midrule
\textbf{Human} & \multicolumn{2}{c|}{\textbf{0.29}} & \multicolumn{2}{c|}{\textbf{0.97}} & \multicolumn{2}{c}{\textbf{0.83}} \\
\bottomrule
\end{tabular}
}
\end{table}
}

\textbf{Subjective.}
As shown in Table~\ref{tab:subj_performance}, integrating AtelierJudge yields significant improvements across all baselines, with GPT-5.4 achieving the strongest alignment. Its MAE decreases to \textbf{0.33}, W1-A rises to \textbf{0.95}, demonstrating exceptional absolute calibration precision. Crucially, on $\rho$, which measures ranking correctness, it reaches \textbf{0.81}, narrowing the gap to human performance to just \textbf{0.02}. This breakthrough stems from correcting the core defects of the baselines. Our qualitative analysis reveals that all MLLM baselines tend to up-shift a large number of samples, resulting in an inflated distribution of scores 4 and 5. This makes it difficult for the models to distinguish between ``good'' and ``perfect.'' Regarding the models, Gem-3 tends to assign more balanced scores, GPT-5.2 is slightly closer to humans in relative ranking, and GPT-5.4 shows higher coarse-grained agreement but also a more lenient judging tendency. AtelierJudge addresses this failure by utilizing memory-augmented evaluation, recovering the critical gradients between scores 4/5 or 3/4. This calibration capability enables MLLMs to adjust their judgments according to human experts, establishing a decisive advantage in discriminative capability.

\textbf{Objective.}
Table~\ref{tab:obj_verif} demonstrates AtelierJudge's exceptional reliability. GPT-5.4 achieves the best overall performance (95.5\% Acc and 93.9\% F1), combining strong prompt-side verification with the highest image accuracy in the most challenging visual dimension. Meanwhile, Gem-3 retains the highest prompt-side accuracy, and Cl-4.5 achieves the strongest prompt-side F1, indicating complementary strengths across textual constraint parsing and binary decision calibration. Overall, the high-precision performance, with prompt accuracy consistently exceeding 95\% and image accuracy surpassing 90\%, confirms that our design meets the strict requirements for automated evaluation.

{
\setlength{\intextsep}{3pt}
\begin{table}[H]
\centering
\caption{Objective verification results by modality.}
\label{tab:obj_verif}
\small
\renewcommand{\arraystretch}{1.1}
\setlength{\tabcolsep}{3pt}
\resizebox{0.99\linewidth}{!}{
\begin{tabular}{l|cc|cc|cc}
\toprule
\multirow{2}{*}{\textbf{Model}} & \multicolumn{2}{c|}{\textbf{Prompt}} & \multicolumn{2}{c|}{\textbf{Image}} & \multicolumn{2}{c}{\textbf{Overall}} \\
\cmidrule(lr){2-3} \cmidrule(lr){4-5} \cmidrule(lr){6-7}
 & Acc (\%) & F1 (\%) & Acc (\%) & F1 (\%) & Acc (\%) & F1 (\%) \\
\midrule
GPT-5.4 & 97.6 & 97.2 & \textbf{93.4} & \textbf{90.6} & \textbf{95.5} & \textbf{93.9} \\
GPT-5.2 & 96.6 & 95.8 & 91.8 & \textbf{90.6} & 94.2 & 93.1 \\
Gem-3 & \textbf{97.8} & 97.0 & 90.7 & 89.6 & 94.3 & 93.3 \\
Cl-4.5 & 97.4 & \textbf{98.0} & 86.1 & 86.6 & 91.7 & 92.3 \\
\bottomrule
\end{tabular}
}
\end{table}
}

\begin{table*}[t]
\centering
\caption{Experimental results averaged by task categories. For both humans and MLLMs, each prompt is executed to produce 4 images using each T2I backend, and the highest-scored image is retained. This setting approximates limited retries in practical workflows. In Appendix~\ref{app:scale}, our empirical analysis shows that 4 samples are sufficient for stable evaluation with negligible variance at large scales. Novice and Skilled denote prompting strategies for MLLMs and user groups for humans.}
\label{tab:main_results}
\setlength{\tabcolsep}{2pt}
\renewcommand{\arraystretch}{1.1}
\resizebox{\textwidth}{!}{
\begin{tabular}{l@{\hspace{2pt}}c|ccc|ccc|cc|c|ccc|ccc|cc|c}
\toprule
\multicolumn{2}{c|}{\multirow{2}{*}{}} & \multicolumn{9}{c|}{\textbf{Novice}} & \multicolumn{9}{c}{\textbf{Skilled}} \\
\cmidrule(lr){3-11} \cmidrule(lr){12-20}
\multicolumn{2}{c|}{} & \textbf{GPT-5.2} & \textbf{Gem-3} & \textbf{Cl-4.5} & \textbf{GPT-4.1} & \textbf{Gem-2} & \textbf{Qwen-L} & \textbf{GPT-4n} & \textbf{Qwen-S} & \textbf{Human} & \textbf{GPT-5.2} & \textbf{Gem-3} & \textbf{Cl-4.5} & \textbf{GPT-4.1} & \textbf{Gem-2} & \textbf{Qwen-L} & \textbf{GPT-4n} & \textbf{Qwen-S} & \textbf{Human} \\
\midrule
\multirow{2}{*}{\textbf{Prompt}} & \textbf{Obj.} & 63.3 & 59.2 & 54.5 & 50.7 & 43.5 & 54.3 & 47.8 & 50.6 & 56.5 & 63.7 & 65.5 & 60.1 & 56.1 & 48.2 & 58.8 & 53.2 & 52.7 & 80.6 \\
 & \textbf{Subj.} & 4.05 & 3.81 & 3.45 & 3.46 & 2.96 & 3.65 & 3.27 & 3.15 & 2.90 & 4.10 & 4.01 & 3.77 & 3.72 & 3.23 & 3.92 & 3.48 & 3.38 & 3.88 \\
\midrule
\multirow{2}{*}{\textbf{Image}} & \textbf{Obj.} & 65.7 & 64.7 & 64.1 & 63.0 & 60.9 & 61.3 & 60.8 & 57.8 & 56.9 & 65.4 & 65.4 & 65.1 & 63.9 & 62.9 & 63.1 & 62.8 & 57.2 & 76.7 \\
 & \textbf{Subj.} & 3.96 & 4.01 & 3.88 & 3.92 & 3.82 & 3.99 & 3.89 & 3.90 & 3.01 & 3.96 & 4.00 & 3.92 & 3.99 & 3.89 & 3.98 & 3.92 & 3.87 & 3.97 \\
\midrule
\multirow{2}{*}{\textbf{nBanana}} & \textbf{Obj.} & 73.2 & 70.6 & 70.0 & 67.6 & 66.0 & 67.1 & 64.7 & 64.1 & 58.2 & 73.5 & 73.9 & 72.5 & 70.3 & 68.7 & 70.9 & 67.7 & 65.4 & 84.9 \\
 & \textbf{Subj.} & 4.14 & 4.10 & 4.02 & 4.07 & 3.95 & 4.13 & 4.08 & 4.05 & 3.11 & 4.17 & 4.15 & 4.08 & 4.17 & 4.06 & 4.12 & 4.08 & 4.01 & 4.11 \\
\midrule
\multirow{2}{*}{\textbf{GI-1}} & \textbf{Obj.} & 71.9 & 69.8 & 70.3 & 67.6 & 66.2 & 66.7 & 65.7 & 64.9 & 66.4 & 73.4 & 72.5 & 72.0 & 69.4 & 68.8 & 69.8 & 68.2 & 64.2 & 83.5 \\
 & \textbf{Subj.} & 4.21 & 4.22 & 4.13 & 4.18 & 4.12 & 4.20 & 4.17 & 4.14 & 3.25 & 4.22 & 4.24 & 4.17 & 4.25 & 4.13 & 4.19 & 4.12 & 4.10 & 4.01 \\
\midrule
\multirow{2}{*}{\textbf{Flux Pro}} & \textbf{Obj.} & 64.1 & 63.8 & 61.1 & 62.3 & 57.7 & 59.5 & 59.3 & 54.4 & 56.8 & 62.2 & 63.2 & 62.1 & 63.6 & 60.7 & 60.5 & 60.7 & 54.0 & 73.1 \\
 & \textbf{Subj.} & 3.92 & 3.98 & 3.83 & 3.88 & 3.68 & 3.92 & 3.78 & 3.81 & 2.94 & 3.90 & 4.00 & 3.84 & 3.93 & 3.79 & 3.94 & 3.85 & 3.82 & 3.93 \\
\midrule
\multirow{2}{*}{\textbf{SDXL}} & \textbf{Obj.} & 53.6 & 54.5 & 54.8 & 54.6 & 53.8 & 51.8 & 53.6 & 48.1 & 46.3 & 52.6 & 52.1 & 53.8 & 52.5 & 53.2 & 50.9 & 54.6 & 45.4 & 65.4 \\
 & \textbf{Subj.} & 3.56 & 3.75 & 3.56 & 3.58 & 3.55 & 3.71 & 3.53 & 3.60 & 2.74 & 3.54 & 3.62 & 3.60 & 3.62 & 3.56 & 3.69 & 3.60 & 3.57 & 3.84 \\
\bottomrule
\end{tabular}
}
\end{table*}

\subsection{Benchmarking Prompting Proficiency}

We design two MLLM prompting strategies to simulate "Novice" behavior via direct natural language prompting and "Skilled" behavior via structured reasoning, with specific prompts provided in Appendix \ref{app:mllm_prompts}. We employ GPT-5.4 as the core evaluator, selected for its superior human alignment in subjective evaluation and robustness in visual evaluation as verified above. Regarding metric calculation, we first aggregate modality-specific subjective and objective scores at the task instance level, which are subsequently analyzed or aggregated for specific comparisons. Case studies and detailed results are provided in Appendix \ref{app:evaluation_examples} and \ref{app:more_result}. Our key \textbf{\textit{Obs}}ervations are presented below.

\textbf{\textit{Obs.1} Effective decoupling of prompting proficiency.}
 As presented in Table \ref{tab:main_results}, across all T2I backends, prompter rankings remain highly consistent, despite localized rank inversions. $T_0$ MLLMs consistently outperform $T_2$ models on average across all backends, and we observe no significant \emph{homophily bias} (e.g., GPT models do not exhibit disproportionate gains on GI-1), validating that our metric captures a transferable, intrinsic capability. Moreover, the ranking of subjective scores largely aligns with objective scores, indicating that performance improvements are holistic rather than narrowly optimized. Notably, while $T_0$ MLLMs perform on par with skilled users, even $T_2$ MLLMs significantly outperform novice users, underscoring the potential of MLLMs in prompting. The gap is mainly reflected in objective metrics, while subjective quality is much closer.

{
\setlength{\intextsep}{5pt}
\setlength{\belowcaptionskip}{-5pt}
 \begin{figure}[H]
    \centering
    \includegraphics[width=0.92\columnwidth]{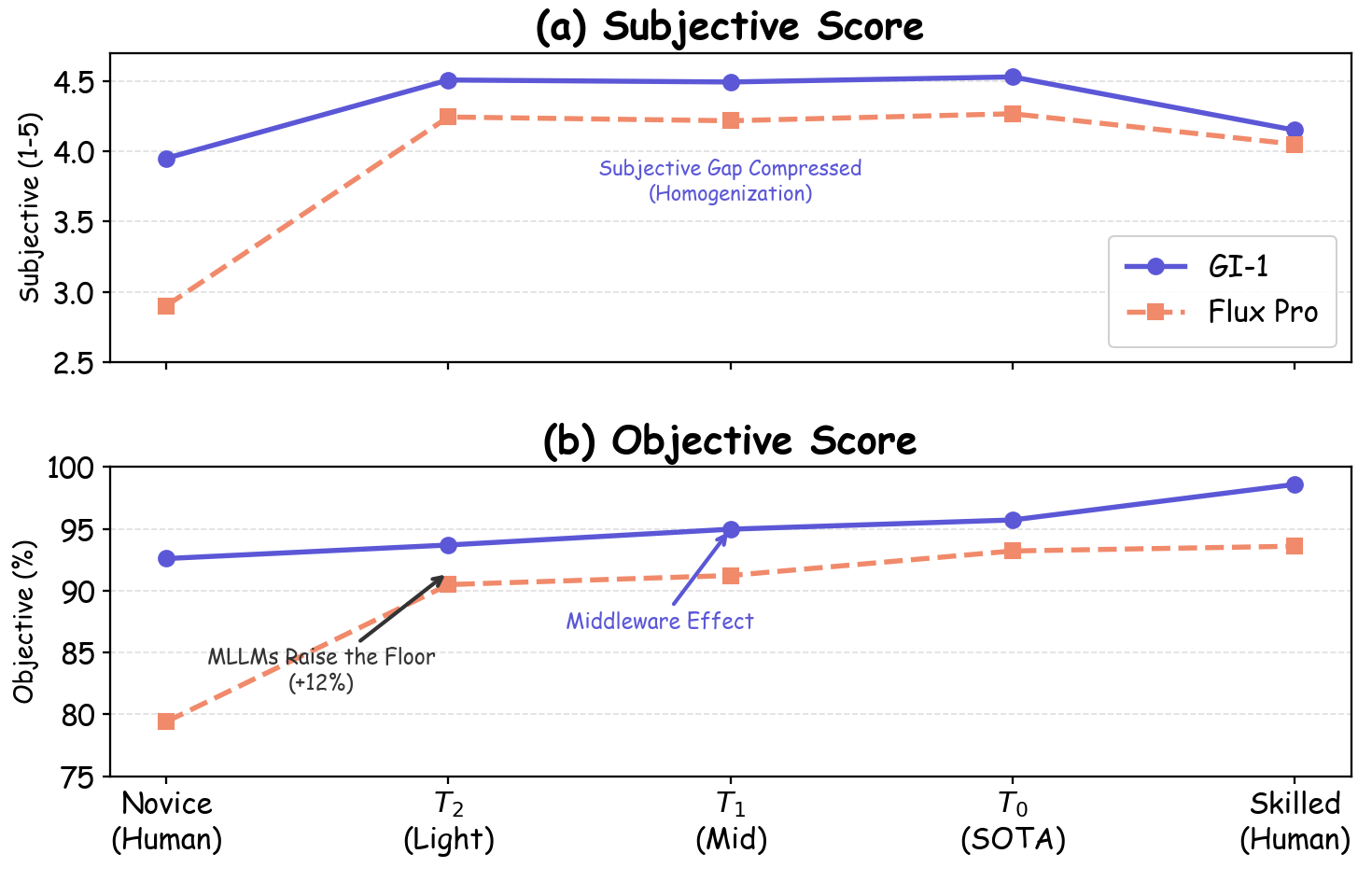}
    \caption{Performance on OE tasks across the proficiency spectrum. MLLM scores are averaged within tiers under novice prompting.}
    \label{fig:obs2}
\end{figure}
}

\textbf{\textit{Obs.2} Homogenization and over-structuring in creation.}
In OE tasks, the middleware plays a decisive role, acting as a performance baseline that significantly compresses variance among prompters. As shown in Figure \ref{fig:obs2}, while the gap between $T_2$ and $T_1$ remains observable, the marginal gains from $T_1$ to $T_0$ diminish due to \emph{homogenization}. This effect is also visible in the aggregated subjective scores over all tasks (Table \ref{tab:main_results}). The results also suggest that $T_0$ MLLMs outperform skilled users, possibly due to larger vocabularies and faster generation, primarily reflected in OE subjective image quality. Notably, the \emph{Skilled} prompting strategy also faces \emph{homogenization}, consistently yielding negligible or negative returns in generated images. This trend is consistent across backends and MLLMs, suggesting a systematic \emph{over-structuring} effect: under loosely constrained settings, rigid structures interfere with T2I models' own creativity.

\textbf{\textit{Obs.3} The paradox of constraints.}
A deeper decomposition reveals a counterintuitive phenomenon in CO tasks, highlighting the non-monotonic nature of performance gains. On T2I systems with strong middleware (e.g., GI-1), as presented in Table \ref{tab:obs34}, we observe severe \emph{logical interference}, as external MLLM reasoning reduces objective scores from a direct-input baseline of $69.6\%$ to $47.1\%$. We attribute this to conflicts between external reasoning and internal rewriting. In contrast, for models without MLLM middleware such as Flux Pro, external reasoning remains positive ($32.2\%\rightarrow37.5\%$). Crucially, skilled users are immune to this interference, achieving the highest accuracy of $81.5\%$. This demonstrates that high-level capability extends beyond reasoning alone and encompasses adaptability.

{
\setlength{\intextsep}{5pt}
\begin{table}[H]
\centering
\caption{Performance comparison on objective scores of CO and IM tasks. Direct denotes using the task descriptions as input to T2I models. G, F, and P denoting GI-1, Flux Pro, and Prompt.}
\label{tab:obs34}
\setlength{\tabcolsep}{3pt}
\renewcommand{\arraystretch}{1.1}
\resizebox{\columnwidth}{!}{
\begin{tabular}{l|c|ccc|c|ccc|c}
\toprule
& & \multicolumn{4}{c|}{\textbf{Novice Prompter}} & \multicolumn{4}{c}{\textbf{Skilled Prompter}} \\
\cmidrule(lr){3-6} \cmidrule(lr){7-10}
& \textbf{Direct} & \textbf{GPT-5.2} & \textbf{Gem-3} & \textbf{Cl-4.5} & \textbf{Human} & \textbf{GPT-5.2} & \textbf{Gem-3} & \textbf{Cl-4.5} & \textbf{Human} \\
\midrule
\textbf{CO G} & 69.6 & 47.2 & 45.5 & 48.6 & \textbf{62.8} & 48.1 & 46.3 & 46.7 & \textbf{81.5} \\
\textbf{CO F} & 32.2 & 38.6 & 39.4 & 34.4 & \textbf{50.1} & 38.5 & 39.2 & 38.1 & \textbf{72.4} \\
\textbf{CO P} & / & 37.5 & 36.2 & 27.8 & \textbf{48.2} & 35.6 & 37.4 & 33.2 & \textbf{78.1} \\
\midrule
\textbf{IM G} & / & \textbf{71.6} & 70.5 & 65.6 & 43.8 & 76.5 & \textbf{77.5} & 74.3 & 70.4 \\
\textbf{IM F} & / & 59.8 & \textbf{60.5} & 54.8 & 40.9 & \textbf{57.9} & 57.3 & 57.1 & 53.3 \\
\textbf{IM P} & / & \textbf{56.7} & 53.5 & 41.2 & 40.4 & 64.6 & 66.6 & 59.0 & \textbf{71.3} \\
\bottomrule
\end{tabular}
}
\end{table}
}

\textbf{\textit{Obs.4} Future of Image-Augmented Prompting.}
IM tasks exhibit distinct behaviors. Crucially, for instance, even when facing constraint complexity comparable to CO settings (up to 20 checkpoints per task), $T_0$ MLLMs with skilled prompts not only excel on the GI-1 backend ($76.1\%$) but demonstrate a clear advantage over human experts ($70.4\%$). We attribute it not merely to the MLLM's powerful visual encoder, whose precision in extracting and verbalizing image features surpasses human linguistic articulation, but fundamentally to the prevalence of \emph{mimicry over planning}. Under equally complex constraints, imitation-based prompting significantly outperforms pure planning. This aligns with the cognitive concept of \emph{recognition over recall}~\cite{srivastava2017simple}, where access to perceptual exemplars substantially reduces the burden of reconstructing complex structures. This finding also supports an emerging workflow that utilizes MLLMs to reverse-engineer prompts from reference images for subsequent intent injection, also validated by our small-scale qualitative studies. Consequently, we empirically suggest that visual exemplar guidance, namely \emph{Image-Augmented Prompting}, constitutes a promising paradigm.

\section{Limitations}
\label{sec:limitations}
We note several limitations of our work:

\begin{itemize}[
    leftmargin=*,     
    topsep=-0.5em,           
    itemsep=0pt       
]
    \item[\ding{182}] \textbf{Demographic Bias in Human Studies.}
    Participants in our study are demographically concentrated (Appendix~\ref{apps:user_demo}). While this distribution is intentionally aligned with the current active user base of T2I systems, human-related experimental results may carry sampling bias. AtelierJudge may inherit these biases.
    
    \item[\ding{183}] \textbf{Interaction Assumptions and Scope.}
    By restricting \textbf{\texttt{AtelierEval}} to a single-turn, pure text-to-image prompting setting, we construct a complete and minimal task partition under this interaction assumption. This abstraction enables controlled, consistent, and diagnostic analysis of prompting proficiency, but does not cover real-world workflows involving multi-turn interaction, iterative refinement with visual feedback, multimodal input, or search-based prompt optimization algorithms.

    \item[\ding{184}] \textbf{Task Difficulty Modeling.}
    In T2I prompting, task difficulty lacks a widely accepted objective metric. Existing work typically characterizes tasks by challenge types or counts, but cannot reliably distinguish which tasks are more difficult for humans or models. Accordingly, \textbf{\texttt{AtelierEval}} balances and reports challenge primitives without explicitly modeling task difficulty.
\end{itemize}

\section{Future Work}
\label{sec:future}

We identify four promising avenues for future research:

\begin{itemize}[
    leftmargin=*,     
    topsep=-0.5em,           
    itemsep=0pt       
]
    \item[\ding{182}] \textbf{Image-augmented and imitation-based prompting.}
    A central insight of this work is that IM tasks achieve higher objective constraint satisfaction than symbolic prompting in CO tasks, even with comparable numbers of constraints. This suggests a promising direction for future prompting agents that leverage image-augmented or retrieval-based mechanisms, analogous to RAG~\cite{li2025miv}. Instead of encoding complex constraints purely symbolically, future systems may retrieve high-quality visual exemplars aligned with user intent and generate prompts by adapting these references to task-specific requirements. This approach aligns with the cognitive ease of \emph{recognition over recall}, mitigating the limitations of purely linguistic reasoning in current text-to-image models.
    
    \item[\ding{183}] \textbf{Human--LLM collaboration in prompting.}
    While \textbf{\texttt{AtelierEval}} intentionally isolates prompters to measure intrinsic prompting proficiency, an important extension is to study collaborative workflows between humans and MLLMs~\cite{li2026prefix,luo2025hai,zou2026llmbasedhumanagentcollaborationinteraction}. Human intervention may partially mitigate constraint failures of MLLMs in constrained creation tasks, while MLLMs may in turn support humans as drafting agents with strong lexical coverage and imitation ability. Systematically evaluating such collaborative settings would require new protocols beyond single-agent prompting, which we leave for future work.

    \item[\ding{184}] \textbf{Beyond single-turn prompting.}
    This benchmark adopts a single-turn interaction paradigm to enable controlled and diagnostic evaluation. Extending it to interactive, multi-turn prompting with visual feedback, tool usage, or search-based optimization~\cite{du2026searchtransferamortizedagentic,zou2026userschangemindevaluating} remains an open challenge and may require new task abstractions and evaluation methodologies beyond isolated prompting policies.

    \item[\ding{185}] \textbf{Unified Multimodal Models.}
    An important future direction is to extend \textbf{\texttt{AtelierEval}} to unified multimodal models (UMMs) that can act both as prompters and as image generators \cite{deng2025bagel,chen2025januspro,xie2024showo}. Unlike conventional evaluations that separately assess language-side reasoning or image-generation quality, \textbf{\texttt{AtelierEval}} provides a unified protocol to disentangle intent-to-prompt translation from prompt-to-image execution within the same system. This makes it particularly suitable for studying self-prompting, prompt-model co-adaptation, and the interaction between internal reasoning and visual generation in unified models.
\end{itemize}

\section{Conclusion}

In this paper, we extend the evaluation paradigm of T2I systems beyond model-centric, prompt-fixed benchmarking to explicitly assess prompting proficiency. To operationalize this paradigm, we introduce \textbf{\texttt{AtelierEval}}, a unified benchmark that evaluates both humans and MLLMs as prompters. We further propose AtelierJudge to enable scalable and precise evaluation with \textbf{\texttt{AtelierEval}}. By formulating prompting proficiency as a cognitive competence, \textbf{\texttt{AtelierEval}} helps transform prompting from a heuristic artifact to a principled and measurable capability, supporting the democratization of generative AI.

\clearpage
\section*{Acknowledgements}
This work is supported in part by the NYUAD Center for Interdisciplinary Data Science \& AI (CIDSAI), funded by Tamkeen under the NYUAD Research Institute Award CG016. 

\section*{Impact Statement}
\label{sec:impact}

This work introduces \textbf{\texttt{AtelierEval}}, an evaluation framework for T2I prompting proficiency, together with its agentic evaluator, AtelierJudge. They provide infrastructure for measuring and analyzing how humans and MLLMs perform as T2I prompters across a wide range of tasks. Standardized benchmarking for T2I prompting significantly enhances the reproducibility and comparability of research on T2I prompting, reducing fragmented and repeat experimentation. By systematically evaluating the capability of MLLMs versus human prompters, our work serves as a diagnostic tool that democratizes access to high-quality creation. As open‑source tools equipped with intuitive interfaces, they also provide novices with a standardized environment for self-assessment and practice.

However, powerful prompting tools carry risks, as they may be exploited to optimize manipulative visual content or lower the threshold for generating unsafe material. Furthermore, our evaluators inevitably inherit biases, risking the reinforcement of aesthetic and representational inequalities. To mitigate these risks, we restrict tasks to non-sensitive scenarios and employ an integrated SafetyFilter. Crucially, we promote transparency by open-sourcing our toolkit with explicit documentation of scope and limitations, while cautioning against sole reliance on automated scoring in high-stakes contexts involving human creators. Detailed ethical protocols are provided in Appendix~\ref{app:ethics}.

\bibliography{refer}
\bibliographystyle{icml2026}
\newpage

\appendix
\onecolumn
\textbf{APPENDIX}
\addcontentsline{toc}{section}{Appendix}

\par
\par 
\par
\par

\noindent\hyperref[app:justify]{\textbf{A\hspace{1em}Formal Justification of Task Completeness}} \hfill \textbf{\pageref{app:justify}}\par
\hspace{2em}\noindent\hyperref[app:C1]{A.1\hspace{1em}Information Sources and Primitive Operations} \dotfill \pageref{app:C1}\par
\hspace{2em}\noindent\hyperref[app:C2]{A.2\hspace{1em}Text-only Tasks: OE and CO} \dotfill \pageref{app:C2}\par
\hspace{2em}\noindent\hyperref[app:C3]{A.3\hspace{1em}Image-conditioned Tasks: IM} \dotfill \pageref{app:C3}\par
\hspace{2em}\noindent\hyperref[app:C4]{A.4\hspace{1em}Coverage and Minimality} \dotfill \pageref{app:C4}\par
\hspace{2em}\noindent\hyperref[app:C5]{A.5\hspace{1em}Scope} \dotfill \pageref{app:C5}\par

\par

\noindent\hyperref[app:category]{\textbf{B\hspace{1em}Application Context Categorization}} \hfill \textbf{\pageref{app:category}}\par
\hspace{2em}\noindent\hyperref[app:4d_scheme]{B.1\hspace{1em}Tagging Scheme} \dotfill \pageref{app:4d_scheme}\par
\hspace{2em}\noindent\hyperref[app:4d_distribution]{B.2\hspace{1em}Distribution of Tags Across Task Categories} \dotfill \pageref{app:4d_distribution}\par

\par

\noindent\hyperref[app:data_collect]{\textbf{C\hspace{1em}Dataset Construction Details}} \hfill \textbf{\pageref{app:data_collect}}\par
\hspace{2em}\noindent\hyperref[apps:data_oe]{C.1\hspace{1em}OE Task Instantiation} \dotfill \pageref{apps:data_oe}\par
\hspace{2em}\noindent\hyperref[apps:data_co]{C.2\hspace{1em}CO Task Instantiation} \dotfill \pageref{apps:data_co}\par
\hspace{2em}\noindent\hyperref[apps:data_im]{C.3\hspace{1em}IM Task Instantiation} \dotfill \pageref{apps:data_im}\par

\par

\noindent\hyperref[app:im_seed]{\textbf{D\hspace{1em}Model/Seed-Agnostic Design of IM Tasks}} \hfill \textbf{\pageref{app:im_seed}}\par

\par

\noindent\hyperref[app:task_instances]{\textbf{E\hspace{1em}Examples of Task Instances}} \hfill \textbf{\pageref{app:task_instances}}\par
\hspace{2em}\noindent\hyperref[app:task_ex_oe]{E.1\hspace{1em}OE Task Example} \dotfill \pageref{app:task_ex_oe}\par
\hspace{2em}\noindent\hyperref[app:task_ex_co]{E.2\hspace{1em}CO Task Example} \dotfill \pageref{app:task_ex_co}\par
\hspace{2em}\noindent\hyperref[app:task_ex_im]{E.3\hspace{1em}IM Task Example} \dotfill \pageref{app:task_ex_im}\par

\par

\noindent\hyperref[app:subjective_dimensions]{\textbf{F\hspace{1em}Subjective Evaluation Dimensions}} \hfill \textbf{\pageref{app:subjective_dimensions}}\par
\hspace{2em}\noindent\hyperref[app:H1]{F.1\hspace{1em}Image-Level Dimensions} \dotfill \pageref{app:H1}\par
\hspace{2em}\noindent\hyperref[app:H2]{F.2\hspace{1em}Prompt-Level Dimensionse} \dotfill \pageref{app:H2}\par

\par

\noindent\hyperref[app:exemplar]{\textbf{G\hspace{1em}Evaluation Exemplar Memory Construction}} \hfill \textbf{\pageref{app:exemplar}}\par

\par

\noindent\hyperref[app:safety]{\textbf{H\hspace{1em}Details of Safety Filter Skill}} \hfill \textbf{\pageref{app:safety}}\par

\par

\noindent\hyperref[app:guideline]{\textbf{I\hspace{1em}Scoring Guidelines for Subjective Evaluation}} \hfill \textbf{\pageref{app:guideline}}\par

\par

\noindent\hyperref[app:judge_prompts]{\textbf{J\hspace{1em}Prompts for AtelierJudge}} \hfill \textbf{\pageref{app:judge_prompts}}\par
\hspace{2em}\noindent\hyperref[app:L1]{J.1\hspace{1em}Subjective Skill Prompts} \dotfill \pageref{app:L1}\par
\hspace{2em}\noindent\hyperref[app:L2]{J.2\hspace{1em}Objective Skill Prompts} \dotfill \pageref{app:L2}\par

\par

\noindent\hyperref[app:para]{\textbf{K\hspace{1em}Model Hyperparameters}} \hfill \textbf{\pageref{app:para}}\par

\par

\noindent\hyperref[app:mllm_prompts]{\textbf{L\hspace{1em}Prompts for MLLM Novice and Skilled Conditions}} \hfill \textbf{\pageref{app:mllm_prompts}}\par
\hspace{2em}\noindent\hyperref[app:N1]{L.1\hspace{1em}Novice MLLM Prompts} \dotfill \pageref{app:N1}\par
\hspace{2em}\noindent\hyperref[app:N2]{L.2\hspace{1em}Skilled MLLM Prompts} \dotfill \pageref{app:N2}\par

\par

\noindent\hyperref[app:demographic]{\textbf{M\hspace{1em}Participant Selection \& Statistics}} \hfill \textbf{\pageref{app:demographic}}\par
\hspace{2em}\noindent\hyperref[app:O1]{M.1\hspace{1em}Participant Selection Criteria} \dotfill \pageref{app:O1}\par
\hspace{2em}\noindent\hyperref[apps:user_demo]{M.2\hspace{1em}Participant Demographic Statistics} \dotfill \pageref{apps:user_demo}\par

\par

\noindent\hyperref[app:experts]{\textbf{N\hspace{1em}Human Expert Involvement}} \hfill \textbf{\pageref{app:experts}}\par

\par

\noindent\hyperref[app:evaluation_examples]{\textbf{O\hspace{1em}Subjective Evaluation Case Studies}} \hfill \textbf{\pageref{app:evaluation_examples}}\par
\hspace{2em}\noindent\hyperref[app:Q1]{O.1\hspace{1em}Subjective Prompt Evaluation (Open-Ended Task)} \dotfill \pageref{app:Q1}\par
\hspace{2em}\noindent\hyperref[app:Q2]{O.2\hspace{1em}Subjective Image Evaluation (Constrained Task)} \dotfill \pageref{app:Q2}\par

\par

\noindent\hyperref[app:valid_set]{\textbf{P\hspace{1em}Details of Validation Set}} \hfill \textbf{\pageref{app:valid_set}}\par

\par

\noindent\hyperref[app:ablation]{\textbf{Q\hspace{1em}Design Validation \& Ablation Study of AtelierJudge}} \hfill \textbf{\pageref{app:ablation}}\par

\par

\noindent\hyperref[app:more_result]{\textbf{R\hspace{1em}Detailed Benchmarking Results}} \hfill \textbf{\pageref{app:more_result}}\par

\par

\noindent\hyperref[app:scale]{\textbf{S\hspace{1em}Stability Analysis of Evaluation Scale}} \hfill \textbf{\pageref{app:scale}}\par

\par

\noindent\hyperref[apx:human_study_materials]{\textbf{T\hspace{1em}User Study Materials}} \hfill \textbf{\pageref{apx:human_study_materials}}\par
\hspace{2em}\noindent\hyperref[app:consent]{T.1\hspace{1em}Informed Consent Form} \dotfill \pageref{app:consent}\par
\hspace{2em}\noindent\hyperref[apx:pre-test]{T.2\hspace{1em}Pre-Test Questionnaire} \dotfill \pageref{apx:pre-test}\par

\par 

\noindent\hyperref[app:assessment_platform]{\textbf{U\hspace{1em}Participant Workflow and Interface Design}} \hfill \textbf{\pageref{app:assessment_platform}}\par
\hspace{2em}\noindent\hyperref[app:W1]{U.1\hspace{1em}Welcome and Authentication} \dotfill \pageref{app:W1}\par
\hspace{2em}\noindent\hyperref[app:W2]{U.2\hspace{1em}Assessment Structure and Randomization} \dotfill \pageref{app:W2}\par
\hspace{2em}\noindent\hyperref[app:W3]{U.3\hspace{1em}Task Type Interfaces} \dotfill \pageref{app:W3}\par
\hspace{2em}\noindent\hyperref[app:W4]{U.4\hspace{1em}Coverage and Minimality} \dotfill \pageref{app:W4}\par

\par

\noindent\hyperref[app:resource]{\textbf{V\hspace{1em}Computational Resource Consumption}} \hfill \textbf{\pageref{app:resource}}\par

\par

\noindent\hyperref[app:ethics]{\textbf{W\hspace{1em}Detailed Ethical Considerations and Procedures}} \hfill \textbf{\pageref{app:ethics}}\par
\hspace{2em}\noindent\hyperref[app:Y1]{W.1\hspace{1em}Ethics Approval and Consent} \dotfill \pageref{app:Y1}\par
\hspace{2em}\noindent\hyperref[apx:confidentiality]{W.2\hspace{1em}Participant Anonymity} \dotfill \pageref{apx:confidentiality}\par

\par
\par

\clearpage

\section{Formal Justification of Task Completeness}
\label{app:justify}

This appendix justifies the completeness claim in Section~\ref{sec:categories}.
Under a restricted interaction setting, we argue that the three task categories (OE, CO, IM) form a complete and minimal set of information-processing primitives for prompting.

\paragraph{Assumption 1 (Single-turn, Pure Text-to-Image).}
We work under the following setting:
\begin{quote}
\emph{
The prompter interacts with the text-to-image model in a single turn and produces exactly one textual prompt.
The generation model conditions solely on this text prompt, without access to any image-based inputs, intermediate feedback, or external control signals.
}
\end{quote}
This is exactly the interface instantiated by \textbf{\texttt{AtelierEval}}.

Under Assumption~1, prompting can be viewed as an information-processing problem at the model interface, in the sense of a source whose information must be transmitted through a constrained channel~\cite{shannon1948mathematical}.
Let $I$ denote the latent abstract intent, $O$ the observable task input, $p$ the textual prompt, $\pi$ the prompting policy, $\mathcal{M}$ the text-to-image model, and $S(\cdot,\cdot)$ an intent-consistency score.
The prompter selects $p = \pi(O)$ to maximize expected intent alignment:
\begin{equation}
    \pi^{*} 
    = \arg\max_{\pi} \; \mathbb{E}_{I}\, \mathbb{E}_{\mathcal{M}}
    \big[\, S(\mathcal{M}(\pi(O)), I) \,\big].
    \label{eq:prompt-objective}
\end{equation}

\subsection{Information Sources and Primitive Operations}
\label{app:C1}
We decompose the observable input as $O = (O_{\text{text}}, O_{\text{img}})$, where $O_{\text{text}}$ is the textual description and $O_{\text{img}}$ an optional target image.
Under Assumption~1, any prompting policy takes the form
\begin{equation}
    p = \pi(I, O_{\text{text}}, O_{\text{img}}),
\end{equation}
and the model $\mathcal{M}$ only observes $p$.
Thus all task-relevant information that can influence $\mathcal{M}$ must flow from the three sources $(I, O_{\text{text}}, O_{\text{img}})$ into the single textual channel.

We distinguish three primitive information-processing operations at this interface:
\begin{enumerate}
    \item \textbf{Textual convergence:} transforming, compressing, and reconciling constraints that are already present in $O_{\text{text}}$ into a single executable prompt (reformatting, reordering, aggregating, resolving conflicts).
    \item \textbf{Intent-driven expansion:} injecting additional structure, detail, and style into $p$ that is not specified in $O_{\text{text}}$ or $O_{\text{img}}$ but is consistent with the abstract intent $I$.
    \item \textbf{Visual-to-textual encoding:} describing aspects of $O_{\text{img}}$ (composition, layout, style, key objects) in text so that they can be realized by $\mathcal{M}$.
\end{enumerate}
By construction, every non-trivial contribution to $p$ must belong to exactly one of these three cases: it either reorganizes information already in $O_{\text{text}}$, introduces new information from $I$, or encodes information from $O_{\text{img}}$.
There is no additional side channel into $\mathcal{M}$.

\subsection{Text-only Tasks: OE and CO}
\label{app:C2}

When the input is purely textual, we write $O_{\text{text}}$ for clarity.
Let
\begin{equation}
    H(I \mid O_{\text{text}})
    \label{eq:conditional-entropy}
\end{equation}
denote the conditional entropy of the abstract intent $I$ given the textual input $O_{\text{text}}$, using standard information-theoretic notation~\cite{cover1999elements}.
We use $H(I \mid O_{\text{text}})$ qualitatively: it measures the residual uncertainty about $I$ after reading the task description, allowing us to distinguish regimes of relatively higher or lower uncertainty.

Real task descriptions often mix abstract, underspecified cues (themes, moods, high-level scenarios) with explicit, low-level constraints (counts, layouts, text rendering).
Accordingly, both convergence and expansion can appear in a single text-only task.
We therefore define OE and CO by their \emph{dominant} operation:
\begin{itemize}
    \item \textbf{Open-ended creation (OE):} $H(I \mid O_{\text{text}})$ is relatively high, and the dominant burden on the prompter is intent-driven expansion.
    Most of the non-trivial content in $p$ originates from $I$ rather than $O_{\text{text}}$.
    \item \textbf{Constrained creation (CO):} $H(I \mid O_{\text{text}})$ is relatively low, but $O_{\text{text}}$ contains many explicit constraints not yet compressed into a prompt.
    The dominant burden is textual convergence: most of the non-trivial content in $p$ reorganizes and integrates information already present in $O_{\text{text}}$.
\end{itemize}
Our OE and CO tasks are constructed so that one of these operations is clearly dominant while the other is relatively simple or trivial.

\subsection{Image-conditioned Tasks: IM}
\label{app:C3}

When $O_{\text{img}}$ is present, the prompter cannot pass it directly to $\mathcal{M}$ and must instead encode its content into text.
Let $I_{\text{img}}$ denote the semantic intent induced by the target image, and let $D(\cdot,\cdot)$ be a semantic distortion measure.
The core information-processing problem in IM tasks can be written as
\begin{equation}
    p^{*} = \arg\min_{p} \; 
    D\big(I_{\text{img}}, \mathcal{M}(p)\big),
    \label{eq:fidelity-objective}
\end{equation}
which is analogous to a fidelity-constrained encoding problem in rate–distortion theory~\cite{cover1999elements}.
Since the only path from $O_{\text{img}}$ to $\mathcal{M}$ is through $p$, the prompter must perform visual-to-textual encoding.
In our benchmark, IM tasks are imitation-oriented, so this encoding is explicitly fidelity-preserving.

More creative image-conditioned tasks (e.g., ``keep the composition but change the mood'') can be decomposed into the same primitives: encode the preserved aspects of $O_{\text{img}}$ (IM), expand $I$ with new elements or styles (OE), and converge all constraints into a single prompt (CO).

\subsection{Coverage and Minimality}
\label{app:C4}

\paragraph{Coverage.}
Under Assumption~1, any prompting policy $\pi$ can only draw information from $(I, O_{\text{text}}, O_{\text{img}})$.
Every non-trivial contribution to $p$ is therefore:
(i) convergence on $O_{\text{text}}$,
(ii) expansion from $I$, or
(iii) encoding from $O_{\text{img}}$.
Text-only tasks ($O_{\text{img}}$ absent) necessarily combine (i) and (ii), with OE and CO distinguished by which operation is dominant.
Image-conditioned tasks necessarily invoke (iii), because information about $O_{\text{img}}$ has no other path to the model.
Mixed cases are compositions of these operations.
Thus, OE, CO, and IM together cover all possible information-processing patterns available at the prompt interface.

\paragraph{Minimality.}
Each primitive is necessary.
There exist tasks whose successful solution requires:
\begin{itemize}
    \item non-trivial intent-driven expansion (OE-style underspecified scenarios);
    \item non-trivial textual convergence (CO-style densely constrained specifications);
    \item non-trivial visual-to-text encoding (IM-style imitation from a target image).
\end{itemize}
Removing any primitive would make the corresponding family of tasks inexpressible as a single prompt.

Conversely, any hypothetical new primitive must either:
(a) reorganize information already present in $O_{\text{text}}$ or $O_{\text{img}}$ (a special case of convergence or encoding), or
(b) inject information consistent with $I$ that is not specified in $O_{\text{text}}$ or $O_{\text{img}}$ (a special case of expansion).
It does not introduce a new information-flow direction, and can be expressed as a tactic within OE, CO, or IM, or as their composition.
In this sense, OE, CO, and IM form a complete and minimal task partition for prompting under Assumption~1, from an information-theoretic perspective.

\subsection{Scope}
\label{app:C5}
If Assumption~1 is relaxed (e.g., multi-round interaction, direct image conditioning at the model interface, external tools), additional information channels become available and further operations—such as iterative refinement or tool-augmented retrieval—may be required.
These settings fall outside the scope of \textbf{\texttt{AtelierEval}} and are left for future work.

\section{Application Context Categorization}
\label{app:category}

This section introduces our application context categorization and tagging scheme, which is used to characterize and summarize how our tasks map onto real-world T2I creative workflows, ensuring broad coverage during task construction and supporting qualitative analysis. This multi-dimensional, non-mutually exclusive design reflects that real-world T2I briefs often span multiple overlapping concerns (e.g., characters and objects both being central) \cite{lin2014microsoft}, and thus better captures how entities, structure, style, and thematic context jointly shape prompting requirements.

\subsection{Tagging Scheme}
\label{app:4d_scheme}
While task construction is driven by challenge primitives and cognitive categories, we explicitly considered common T2I application contexts throughout the design process. We treat the application context of each task as a multi-label annotation vector
\begin{equation}
\label{eq:4d_tagging}
C = \{E, S, V, T\},
\end{equation}

where $E$ denotes \emph{Entities}, $S$ denotes \emph{Structure}, $V$ denotes \emph{Visual Style}, and $T$ denotes \emph{Theme \& Context}. The tagging scheme is not intended as a strict ontology. Instead, it is a lightweight, non-exclusive label set designed to reflect how tasks map onto common T2I application scenarios. Concretely:
\begin{itemize}[leftmargin=*,
    topsep=-0.5em,
    partopsep=0pt,
    parsep=1pt,
    itemsep=0pt
    ]

\item[\ding{224}] \textbf{Entities ($E$)} describe what is depicted.

\item[\ding{224}] \textbf{Structure ($S$)} describes how content is arranged in space or sequence.

\item[\ding{224}] \textbf{Visual Style ($V$)} describes how the image is rendered.

\item[\ding{224}] \textbf{Theme \& Context ($T$)} describes the high-level narrative or aesthetic context.

\end{itemize}

Each task may carry multiple tags both across and within dimensions, e.g., a fantasy character sheet rendered in cel-shaded style with a knolling layout. Tags are only used for coverage analysis and qualitative discussion.

\subsection{Distribution of Tags Across Task Categories}
\label{app:4d_distribution}

We summarize the application-context coverage considered during task construction over all 360 tasks using the tagging scheme. Table~\ref{tab:4d_distribution} reports the occurrence counts of all used tags.

{
\setlength{\belowcaptionskip}{-5pt}
\begin{table}[H]
\centering
\small
\caption{Tags and their occurrence counts in the three task categories. The table is intended as a coverage summary rather than a balanced distribution, illustrating the broad coverage of real-world T2I application scenarios considered in \textbf{\texttt{AtelierEval}}.}
\label{tab:4d_distribution}

\begin{tabular}{@{}lccc|lccc@{}}
\toprule
\multicolumn{4}{c|}{\textbf{}} & \multicolumn{4}{c}{\textbf{}} \\
\textbf{Tag} & \textbf{OE} & \textbf{CO} & \textbf{IM} &
\textbf{Tag} & \textbf{OE} & \textbf{CO} & \textbf{IM} \\
\midrule

\multicolumn{4}{l|}{\textit{Entities}} &
\multicolumn{4}{l}{\textit{Visual Style}} \\
\midrule

\#Object        & 56 & 96 & 39 & \#Photorealistic     & 43 & 37 & 48 \\
\#Character     & 49 & 37 & 58 & \#Traditional\_Media & 50 & 12 & 6  \\
\#Environment   & 45 & 34 & 54 & \#Vector\_Flat       & 7  & 52 & 18 \\
\#Typography    & 6  & 76 & 18 & \#3D\_Render         & 15 & 26 & 27 \\
\#Data\_Element & 17 & 32 & 14 & \#Cel\_Shaded        & 5  & 8  & 19 \\
\midrule

\multicolumn{4}{l|}{\textit{Structure}} &
\multicolumn{4}{l}{\textit{Theme \& Context}} \\
\midrule

\#Portrait\_CloseUp  & 27 & 25 & 22 & \#Corporate\_Clean      & 21 & 56 & 35 \\
\#Full\_Body\_Shot   & 18 & 21 & 20 & \#SciFi\_Cyberpunk      & 24 & 14 & 17 \\
\#Schematic\_Diagram & 18 & 25 & 26 & \#Fantasy\_Mythic       & 26 & 9  & 13 \\
\#Knolling\_Layout   & 10 & 15 & 6  & \#Abstract\_Conceptual  & 25 & 2  & 11 \\
\#Sequential\_Panel  & 7  & 6  & 15 & \#Cute\_Pop             & 15 & 27 & 13 \\
\#UI\_Interface      & 8  & 6  & 12 & \#Retro\_Vintage        & 27 & 15 & 4  \\
\#Isometric\_View    & 0  & 6  & 4  & \#Horror\_Dark          & 16 & 3  & 1  \\

\bottomrule
\end{tabular}
\end{table}
}

\section{Dataset Construction Details}
\label{app:data_collect}

{
\setlength{\belowcaptionskip}{-5pt}
\begin{table}[H]
  \centering
  \caption{Structural challenge load statistics aggregated across the 360 tasks. For each task category, we report summary statistics of the number of semantic and constraint challenge primitive instances per task.}
  \label{tab:challenge-load}
  \begin{tabular}{lcccccccc}
    \toprule
    & \multicolumn{4}{c}{\# semantic primitive instances / task}
    & \multicolumn{4}{c}{\# constraint primitive instances / task} \\
    \cmidrule(lr){2-5}\cmidrule(lr){6-9}
    Category & Min & Max & Mean & Std & Min & Max & Mean & Std \\
    \midrule
    OE & 6 & 10 & 8.2 & 0.9 & 0 & 0 & 0.0 & 0.0 \\
    CO & 3 & 4 & 3.7 & 0.4 & 10 & 18 & 12.7 & 3.1 \\
    IM & 3 & 7 & 4.3 & 1.0 & 7 & 16 & 12.4 & 2.8 \\
    \bottomrule
  \end{tabular}
\end{table}
}

This section supplements Section \ref{sec:dataset} by detailing the task instantiation procedures. We organize the discussion by the three task categories and describe the corresponding instantiation procedures for each. Illustrative examples are provided in Appendix \ref{app:task_instances}. The summarized statistics of challenge primitives are provided in Table \ref{tab:challenge-load}.

\subsection{OE Task Instantiation}
\label{apps:data_oe}

Open-ended Creation (OE) tasks are designed to evaluate a prompter's ability to translate abstract, fuzzy, and unstructured requests into executable prompts. They focus on upstream semantic analysis and creative translation. OE tasks are presented as full natural-language paragraphs that deliberately introduce \emph{narrative noise} to simulate real creative briefs. The description contains a substantial amount of contextually meaningful information that is common in real-world workflows but has no direct visual counterpart, and prompters must filter such narrative context, identify visually relevant intent, and translate it into an executable specification. OE tasks only use semantic challenge primitives $\{S_i\}$ and exclude explicit $\{C_j\}$. They are instantiated across diverse T2I application contexts to avoid tying the evaluation to any particular subject matter or style. Formally, we write an OE task as:

 {
\setlength{\abovedisplayskip}{3pt} 
\setlength{\belowdisplayskip}{3pt}
\begin{equation}
\label{eq:oe_compose}
t_{\mathrm{OE}} \;=\; \bigl(N_{text},\; n_{1} S_1 + n_{2} S_2 + n_{3} S_3 + n_{4} S_4,\; A_t\bigr),\quad
6 \;\le\; \sum_{i=1}^4 n_{i} \;\le\; 10,
\end{equation}
}

where $N_{text}$ is the high-noise natural-language description, $A_t$ is the multi-label application-context tag vector defined in Appendix~\ref{app:category}, and $n_{i} \in \mathbb{Z}_{\ge 0}$ is the number of instances of semantic primitive $S_i$ embedded in the task.

The challenge load of an OE task is thus determined jointly by the total number of embedded semantic cues $\sum_i n_{i}$ and the translation difficulty of their concrete instances. Even within the same primitive type, instances can vary in how hard they are to render into visual terms (e.g., a simple emotional descriptor versus a highly abstract aesthetic intent, or a broad audience description versus a highly specific cultural group). During task design, we explicitly control both factors: every OE task contains between 6 and 10 key semantic cues that require translation, and we avoid stacking multiple high-difficulty instances within a single task, thereby constraining the challenge load to a reasonable and comparable range across OE tasks.

Before inclusion in the benchmark, all OE tasks undergo manual validation to ensure that tasks are understandable and executable by two independent domain experts. Each expert independently translates the task into prompts; a task is accepted only if both agree that the intent is clear and unambiguous, and that all four text-to-image models used in our experiments can produce images within multiple generations that broadly satisfy the intended requirements.

\subsection{CO Task Instantiation}
\label{apps:data_co}

Constrained Creation (CO) tasks are designed to evaluate a prompter’s ability to integrate multiple explicit constraints into an executable prompt under clearly specified requirements. In task construction, each CO task starts with a concise natural-language description that provides basic context and semantic direction. This description explicitly contains a small number of semantic challenge primitives $\{S_i\}$, specifying affect, audience, or stylistic intent with relatively low ambiguity. For each task, $3$--$4$ types of semantic primitives from $S_1$--$S_4$ are involved, and each type typically appears only once. In contrast, constraint challenge primitives $\{C_j\}$ are provided through structured fields, explicitly separating semantic background from executable constraints. Each CO task includes $3$--$5$ different types of constraint primitives, and each type may contain multiple concrete constraint instances. They are also instantiated across diverse T2I application contexts to avoid tying the evaluation to any particular subject matter or style. Formally, we write a CO task as

 {
\setlength{\abovedisplayskip}{3pt} 
\setlength{\belowdisplayskip}{3pt}
\begin{equation}
\label{eq:co_compose}
t_{\mathrm{CO}} \;=\;
\bigl(
D_{\text{text}},\;
m_{1} S_1 + m_{2} S_2 + m_{3} S_3 + m_{4} S_4,\;
n_{1} C_1 + n_{2} C_2 + n_{3} C_3 + n_{4} C_4 + n_{5} C_5,\;
A_t
\bigr),
\end{equation}
}
with the following constraints:
 {
\setlength{\abovedisplayskip}{3pt} 
\setlength{\belowdisplayskip}{3pt}
\begin{equation}
\label{eq:co_bounds}
3 \le \sum_{i=1}^4 m_i \le 4,\quad
3 \le \sum_{j=1}^5 \mathbb{I}(n_j > 0) \le 5,
\end{equation}
}

where $D_{\text{text}}$ denotes the clear natural-language task description, $m_i \in \{0,1\}$ indicates whether semantic primitive $S_i$ is present, and $n_j \in \mathbb{Z}_{\ge 0}$ denotes the number of concrete constraint instances associated with primitive $C_j$.

The challenge load of a CO task is determined jointly by the number of active constraint primitives and the total number of executable constraint instances. During task design, we explicitly control these factors by constraining the total number of key constraint checks to the range above, while avoiding stacking too many high-strictness constraint instances within a single task.

Before inclusion in the benchmark, all CO tasks are manually validated by two independent domain experts. The experts independently write prompts for each task and examine whether, using the four text-to-image models employed in our experiments, the generated images are logically interpretable. A task is included only if the experts agree that the task description is unambiguous, the structured constraints contain no systematic conflicts, and that across repeated generations or different prompt attempts, each category of specified constraints admits at least one satisfiable generation instance.

\subsection{IM Task Instantiation}
\label{apps:data_im}

Imitation (IM) tasks are designed to evaluate a prompter’s ability to translate visual observations into executable prompts~\cite{yuan2022optical}. Each IM task is instantiated from a target image, which serves as the ground truth specification and is firstly obtained through mixed sources, including real photographs sourced from CC0 or CC0-equivalent open-license repositories, AI-generated images, and manual drafts. Then we manually refine all collected images to ensure full compliance with a custom checklist for evaluation purposes and prevent potential data leakage. Such refinement includes adding or removing text or objects, adjusting color, scaling or cropping, cloning or removing stamps, and AI-based editing.

IM tasks instantiate visual counterparts of challenge primitives. We restrict IM to primitives with sufficient visual observability. Specifically, IM tasks exclude $S_2$ (Audience Intent), $S_4$ (Semantic Negation), and $C_5$ (Hard Constraint). Audience intent reflects creator-side goals that are not uniquely encoded in visual appearance; semantic negation specifies what should not be generated, which is underdetermined from a single image; and global hard constraints cannot be inferred from an image without additional contextual assumptions. Including these primitives would introduce systematic ambiguity. Formally, we write an IM task as
\begin{equation}
\label{eq:im_compose_corrected}
t_{\mathrm{IM}} \;=\;
\bigl(
I_{\mathrm{target}},\;
k_{1} \mathrm{VS}_1 + k_{2} \mathrm{VS_3} + n_{1} \mathrm{VC}_1 + n_{2} \mathrm{VC}_2 + n_{3} \mathrm{VC}_3 + n_{4} \mathrm{VC}_4,\;
A_t
\bigr),
\end{equation}
with the following bound on checklist size:
\begin{equation}
\label{eq:im_bounds_corrected}
10 \;\le\; \sum_{m} k_{m} + \sum_{j} n_{j} \;\le\; 20.
\end{equation}

Here $I_{\mathrm{target}}$ denotes the finalized target image, $\mathrm{VS}_1$ and $\mathrm{VS}_3$ are the visual counterparts of $S_1$ and $S_3$, and $\mathrm{VC}_1$–$\mathrm{VC}_4$ are the visual counterparts of constraint primitives $C_1$–$C_4$.

The challenge load of an IM task is determined by the number and diversity of checklist items extracted from the target image. During task design, we explicitly control this load by constraining the total number of checklist items to a moderate range, as shown in Equation \ref{eq:im_bounds_corrected}, and by avoiding excessive stacking of highly ambiguous or fine-grained visual properties.

Before inclusion in the benchmark, all IM tasks undergo manual validation by two independent domain experts. 
Each expert independently attempts to reproduce the target image by writing prompts and generating images; a task is included only if both experts can, through multiple attempts, produce at least one image that fully satisfies all checklist items. 
This validation ensures that IM tasks are practically reproducible and that evaluation reflects the prompter’s ability to encode visual semantics, rather than artifacts of target construction.

Ultimately, we note that fixed-seed reconstruction, i.e., evaluation strategies that fix the random seed of a text-to-image model and assess a prompter’s ability to reproduce a target image under the same model configuration, is a deliberately excluded design choice for IM tasks. A detailed discussion of this design decision is provided in Appendix~\ref{app:im_seed}.

\section{Model/Seed-Agnostic Design of IM Tasks}
\label{app:im_seed}

This appendix clarifies the motivation behind the model- and seed-agnostic design of IM tasks. We argue that such a formulation is necessary to ensure that IM evaluation reflects prompting proficiency itself.

\textbf{Conventional Fixed-Seed Design.}
A natural and widely adopted evaluation method for imitation ability is to fix both the generation model and random seed, and evaluate a prompter by reproducing a target image under identical settings with a reconstruction-based score. This setup offers strong determinism: when the prompt exactly matches the original, the target image can in principle be reproduced without ambiguity. While such strategies have been explored in prior work \cite{dong2023prompt,mahajan2024prompting}, they do not meet our evaluation requirements in several important respects:

\begin{itemize}[leftmargin=*,
    topsep=-0.5em,
    partopsep=0pt,
    parsep=0pt,
    itemsep=0pt
    ]

\item[\ding{224}] \textbf{Model Coverage.}
Most commercial text-to-image models, including widely used systems such as the DALL·E and NanoBanana series, do not expose random seed control, and restricting evaluation to seed-controllable models would therefore substantially narrow benchmark coverage. Relying exclusively on open-source models does not resolve this issue, as current open-source systems lack sufficient robustness to support the full range of complex IM tasks, limiting prompter evaluation.

\item[\ding{224}] \textbf{Evaluation Stability.}
Although the random seed controls stochastic initialization, text-to-image models remain highly sensitive to prompt-level variations through their conditional representations. Even minor linguistic changes that preserve semantic equivalence can therefore induce substantial differences in composition and object configuration under the same seed. This sensitivity introduces large variance into reconstruction-based scores, making evaluation outcomes depend more on randomness than on a prompter’s underlying ability.

\item[\ding{224}] \textbf{Evaluation Objective.}
Fixed-model, fixed-seed reconstruction evaluates whether a prompter can reproduce a specific realization of a generative model, effectively measuring prompt inversion or model-specific overfitting. In contrast, IM tasks aim to assess a prompter’s ability to analyze visual content and encode its underlying semantics in a transferable, model-agnostic form. Consequently, success under reconstruction-based scores does not necessarily reflect stronger prompting ability.

\end{itemize}

\textbf{Our Design.}
We deliberately avoid reconstruction-based evaluation under fixed model–seed configurations for IM tasks. Instead, target images are treated as semantic specifications and are collected from multiple sources, as detailed in Appendix~\ref{apps:data_im}. This design supports multiple text-to-image backends without privileging any particular model, and aligns the evaluation objective with prompting proficiency rather than reconstruction fidelity. Empirically, this design choice is supported by our experimental results. As shown in Table \ref{tab:im}, prompter performance on IM tasks exhibits consistent rankings across all evaluated T2I backends, indicating that the evaluation captures the model-agnostic notion of prompting proficiency.

\section{Examples of Task Instances}
\label{app:task_instances}

To make the design principles and construction procedures more concrete, we also present one representative task from each category. For each instance, we (i) show the original task as presented to prompters and (ii) analyze how its construction arises from composing the challenge primitives in Table~\ref{tab:challenge_primitives}. We additionally visualize the full prompting pipeline for each task instance (Figures~\ref{fig:example_oe}, Figures~\ref{fig:example_co} and Figures~\ref{fig:example_im}). Notably, these examples are selected to illustrate how variations in challenge load naturally emerge from different combinations of primitives, without relying on ad-hoc complexity labels.

\subsection{OE Task Example}
\label{app:task_ex_oe}

\textbf{Task Description.}
The OE track probes divergent production under high narrative noise and without explicit constraint primitives $\{C_j\}$. Task \texttt{oe\_29} below is instantiated in a commercial illustration context and targets the \emph{Environment} and \emph{Photorealistic} application categories.

\begin{tcolorbox}[breakable,colback=gray!10,colframe=gray!50,boxrule=0.5pt]
\textbf{Task ID:} \texttt{oe\_29}\quad
\textbf{Title:} Sunset Mountain Landscape\\[0.4em]
\textbf{Task:}
The Aether Hotel chain is commissioning art prints for their new Alpine location, and our studio is bidding for the project. The theme is Majesty of Nature. The art consultants brief asks for a landscape photography-style scene of mountains at sunset. Key elements are warm golden light, mist-filled valleys, and dramatic depth. This is for their high-end guests, so it must look premium and peaceful. It must look hyper-realistic, like a high-end photo. It cannot be a simple drawing or painting, and it absolutely must not contain any people, roads, or buildings.
\end{tcolorbox}

\textbf{Challenge Primitive Decomposition.}
This instance deliberately concentrates semantic challenge primitives while avoiding explicit structural constraints:
\begin{itemize}[leftmargin=*]
    \item \textbf{Abstract Intent ($S_1$).} The theme ``Majesty of Nature'' and the requirement that the scene feel ``premium and peaceful'' encode affective goals that must be visualized rather than directly executed.
    \item \textbf{Audience Intent ($S_2$).} The mention of ``high-end guests'' and the hotel commissioning context implicitly targets a luxury audience, steering style and polish without prescribing concrete attributes.
    \item \textbf{Implicit Style ($S_3$).} Phrases such as ``landscape photography-style'' and ``hyper-realistic, like a high-end photo'' specify a photographic look and camera-like realism without enumerating technical parameters (lens, focal length, etc.), requiring the prompter to complete these details.
    \item \textbf{Semantic Negation ($S_4$).} The brief bans ``simple drawing or painting'' and ``any people, roads, or buildings.'' These exclusions operate at the semantic level (which concepts must be absent), not as hard procedural constraints $C_5$.
\end{itemize}
No explicit constraint primitives $C_1$--$C_4$ are introduced: there is no fixed object count, layout template, or textual content to render. The task thus isolates the semantic interpretation and translation component of prompting proficiency.

To operationalize this decomposition in our objective evaluation, we summarize the image-level and prompt-level criteria in Table~\ref{tab:checklist_oe29}.

\begin{table}[H]
\centering
\small
\caption{Objective image- and prompt-based checklists for OE task \texttt{oe\_29}.}
\label{tab:checklist_oe29}

\begin{tabular}{@{}p{0.43\linewidth}|p{0.53\linewidth}@{}}
\toprule
\multicolumn{1}{c|}{\textbf{Image-based checklist}} &
\multicolumn{1}{c@{}}{\textbf{Prompt-based checklist}} \\
\midrule

I1.\ Mountains are visible in the image. &
P1.\ The prompt mentions mountains or related terms. \\
I2.\ The scene is set at sunset. &
P2.\ The prompt mentions sunset or related terms. \\
I3.\ Mist or fog is visible in valleys. &
P3.\ The prompt mentions mist, fog, or related terms. \\
I4.\ No human figures are visible in the image. &
P4.\ The prompt excludes people, human figures, or related terms. \\
I5.\ No roads or buildings are visible in the image. &
P5.\ The prompt excludes roads, buildings, or man-made structures. \\
I6.\ The visual style is photorealistic &
P6.\ The prompt specifies photographic or photorealistic as the style. \\
\bottomrule
\end{tabular}

\end{table}

\textbf{Design Rationale.}
This task's challenge load lies primarily in filtering narrative noise and aggregating a small, coherent set of semantic requirements. A competent prompter can translate the brief into a concise prompt (e.g., by foregrounding the Alpine setting, golden-hour lighting, and absence of man-made elements) without managing complex combinatorial constraints. Typical failure modes include (i) under-specifying the atmosphere, yielding generic mountain photos that miss the ``Majesty of Nature'' mood ($S_1$), or (ii) neglecting semantic negation, leading to cabins, hiking trails, or human figures in the generations ($S_4$). These errors directly diagnose weaknesses in semantic decoding rather than structural prompt construction.

\begin{figure}[H]
    \centering
    \includegraphics[width=0.98\linewidth]{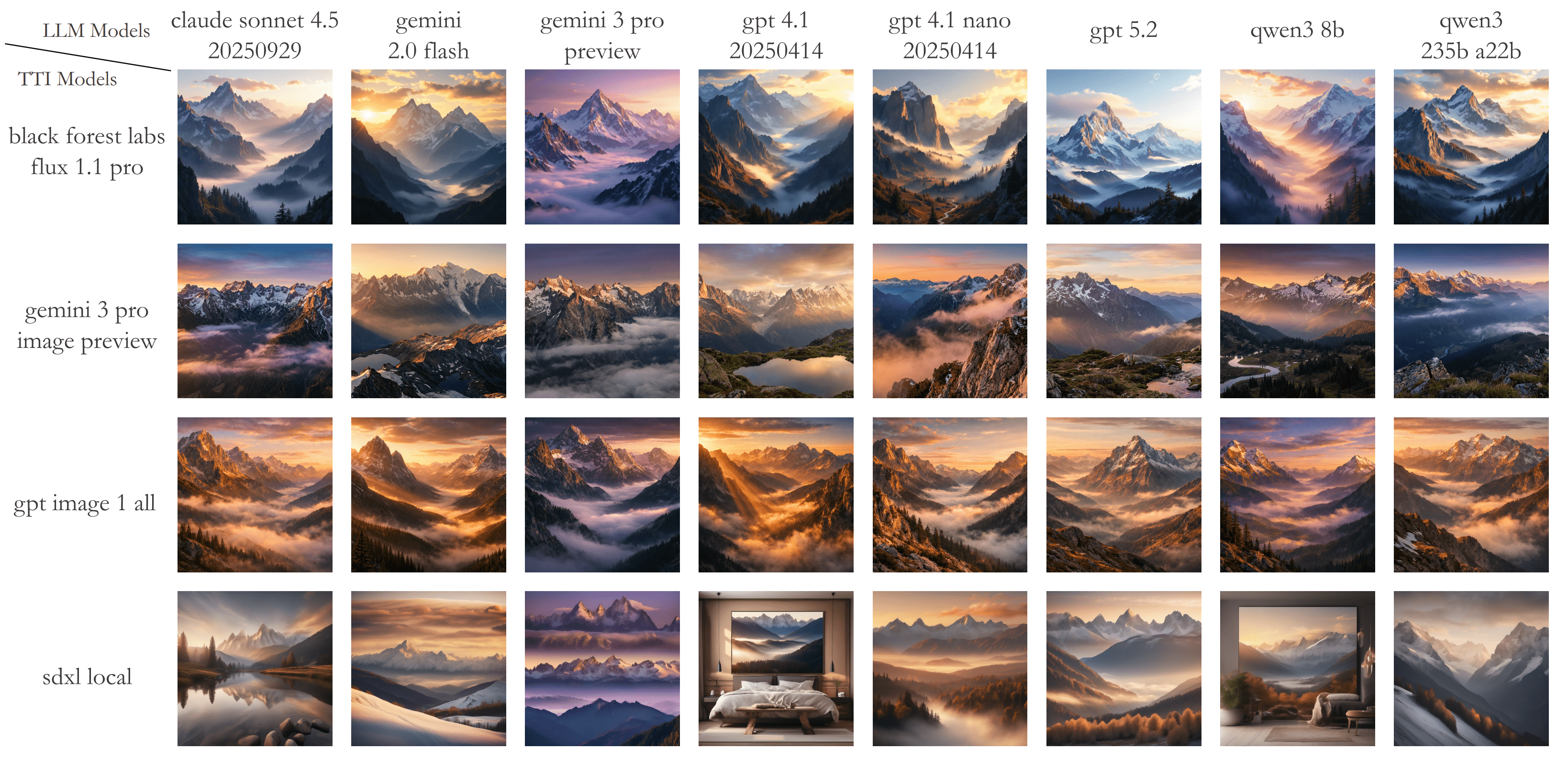}
  \caption{Qualitative generations for OE task \texttt{oe\_29} from the skilled condition.}
    \label{fig:example_oe}
\end{figure}

\textbf{Qualitative generations across models.}
To see how they manifest in practice across our model zoo, we also show qualitative generations for this instance in Figure~\ref{fig:example_oe}. A notable failure mode arises from the hotel-related narrative context: several prompters overemphasize the commissioning background, leading SDXL to generate interior hotel scenes rather than outdoor mountain landscapes.

\subsection{CO Task Example}
\label{app:task_ex_co}

\textbf{Task Description.}
CO tasks emphasize convergent production under low narrative noise, concentrating multiple explicit constraints $\{C_j\}$ that must be jointly realized in a single prompt. Task \texttt{co\_106} below is instantiated in a sequential-panel comic design setting.

\begin{tcolorbox}[breakable,colback=gray!10,colframe=gray!50,boxrule=0.5pt]
\textbf{Task ID:} \texttt{co\_106}\quad
\textbf{Title:} Morning Routine Four-Panel Comic\\[0.4em]
\textbf{Brief:}
A lifestyle blog needs a simple comic strip about daily routines for their wellness section. The comic must be relatable and clean, for general readers. This is a slice-of-life comic strip. The comic must not include any speech bubbles inside panels.\\[0.4em]
\textbf{Character and Object Attributes:}
The main character must have black hair in all four panels. The character must wear a blue shirt in all panels. The alarm clock must be red.\\[0.4em]
\textbf{Layout:}
Standard horizontal four-panel layout. Panel~1: character asleep with alarm ringing. Panel~2: character groggily sitting up. Panel~3: character stretching. Panel~4: character smiling while holding coffee mug.\\[0.4em]
\textbf{Quantity:}
Exactly four panels of equal size arranged horizontally. The same character must appear in all four panels.\\[0.4em]
\textbf{Text Rendering:}
Below each panel, captions must read: ``Panel 1: 6:00 AM'', ``Panel 2: Wake up'', ``Panel 3: Stretch'', ``Panel 4: Ready for the day''.\\[0.4em]
\textbf{Global Constraints:}
The comic must use \emph{only} black (hair and outlines), blue (shirt), red (alarm clock), and white background. All panels must have identical dimensions. Art style must be simple line art with flat colors only.
\end{tcolorbox}

\textbf{Challenge Primitive Decomposition.}
Compared to OE, this instance carries a much denser mix of constraint primitives, while semantic primitives are expressed explicitly with little ambiguity:
\begin{itemize}[leftmargin=*]
    \item \textbf{Semantic primitives.} The slice-of-life setting and requirement that the strip be ``relatable and clean'' instantiate $S_1$ and $S_2$, but here they function mainly as context that bounds aesthetics rather than as implicit cues.
    \item \textbf{Attribute Binding ($C_1$).} Hair color, shirt color, and alarm clock color must be consistently bound to the correct entities across all panels (e.g., no panel where the shirt changes color).
    \item \textbf{Spatial Relation ($C_2$).} The standard horizontal four-panel layout, panel-wise narrative progression, and the positioning of captions beneath each panel all impose explicit spatial structure.
    \item \textbf{Quantity ($C_3$).} The requirement of ``exactly four panels of equal size arranged horizontally'' and ``the same character must appear in all four panels'' imposes strict cardinality constraints on both panels and character instances.
    \item \textbf{Text ($C_4$).} The four captions require precise text rendering and spelling, with panel indices and phrases that must appear verbatim under the corresponding panels.
    \item \textbf{Hard Constraint ($C_5$).} The restricted color palette and prohibition of speech bubbles act as global, non-relaxable constraints that any valid solution must respect.
\end{itemize}
The resulting challenge load is thus dominated by $\{C_1,\dots,C_5\}$, with semantics largely transparent. This matches the CO design goal of making the main difficulty the integration of many interacting constraints into one executable prompt.


We summarize the image-level and prompt-level criteria in 
Table~\ref{tab:checklist_co106}.

\begin{table}[H]
\centering
\small
\caption{Objective image- and prompt-based checklists for CO task \texttt{co\_106}.}
\label{tab:checklist_co106}

\begin{tabular}{@{}p{0.48\linewidth}|p{0.48\linewidth}@{}}
\toprule
\multicolumn{1}{c|}{\textbf{Image-based checklist}} &
\multicolumn{1}{c@{}}{\textbf{Prompt-based checklist}} \\
\midrule

I1.\ Four panels are present in the image. &
P1.\ The prompt mentions four panels or panel sequence. \\
I2.\ A character is present in all panels. &
P2.\ The prompt mentions a character, person, or related terms. \\
I3.\ An alarm clock is present in the image. &
P3.\ The prompt mentions an alarm clock or related terms. \\
I4.\ A coffee mug is present in the image. &
P4.\ The prompt mentions a coffee mug, mug, or related terms. \\
I5.\ The character has black hair in all four panels. &
P5.\ The prompt describes character with black hair in all panels. \\
I6.\ The character wears a blue shirt in all panels. &
P6.\ The prompt describes character wearing blue shirt in all panels. \\
I7.\ The alarm clock is red. &
P7.\ The prompt specifies alarm clock as red. \\
I8.\ The layout is standard horizontal four-panel. &
P8.\ The prompt describes horizontal four-panel layout or four panels in a row. \\
I9.\ Panel 1 shows character asleep with alarm ringing. &
P9.\ The prompt describes panel 1 with character asleep and alarm ringing. \\
I10.\ Panel 2 shows character groggily sitting up. &
P10.\ The prompt describes panel 2 with character groggily sitting up. \\
I11.\ Panel 3 shows character stretching. &
P11.\ The prompt describes panel 3 with character stretching. \\
I12.\ Panel 4 shows character smiling while holding coffee mug. &
P12.\ The prompt describes panel 4 with character smiling holding coffee mug. \\
I13.\ Exactly four panels of equal size are arranged horizontally (not more, not fewer). &
P13.\ The prompt specifies exactly four panels of equal size arranged horizontally. \\
I14.\ Text below panel 1 reads `Panel 1: 6:00 AM' or caption reads `6:00 AM'. &
P14.\ The prompt specifies caption `Panel 1: 6:00 AM' or `6:00 AM' below panel 1. \\
I15.\ Text below panel 2 reads `Panel 2: Wake up' or caption reads `Wake up'. &
P15.\ The prompt specifies caption `Panel 2: Wake up' or `Wake up' below panel 2. \\
I16.\ Text below panel 3 reads `Panel 3: Stretch' or caption reads `Stretch'. &
P16.\ The prompt specifies caption `Panel 3: Stretch' or `Stretch' below panel 3. \\
I17.\ Text below panel 4 reads `Panel 4: Ready for the day' or caption reads `Ready for the day'. &
P17.\ The prompt specifies caption `Panel 4: Ready for the day' or `Ready for the day' below panel 4. \\
I18.\ No speech bubbles are visible inside panels. &
P18.\ The prompt excludes speech bubbles inside panels or mentions no speech bubbles/captions only below. \\
I19.\ The comic uses only four colors: black (hair and outlines), blue (shirt), red (alarm clock), and white background. &
P19.\ The prompt specifies four colors: black (hair/outlines), blue (shirt), red (alarm clock), white background. \\
I20.\ The art style is simple line art with flat colors only. &
P20.\ The prompt describes simple line art with flat colors or minimal style. \\
I21.\ The visual style is flat vector or simple comic strip illustration. &
P21.\ The prompt specifies flat vector, comic strip, slice-of-life comic, or related flat comic style terms. \\
\bottomrule
\end{tabular}

\end{table}

\begin{figure}[H]
    \centering
    \includegraphics[width=0.98\linewidth]{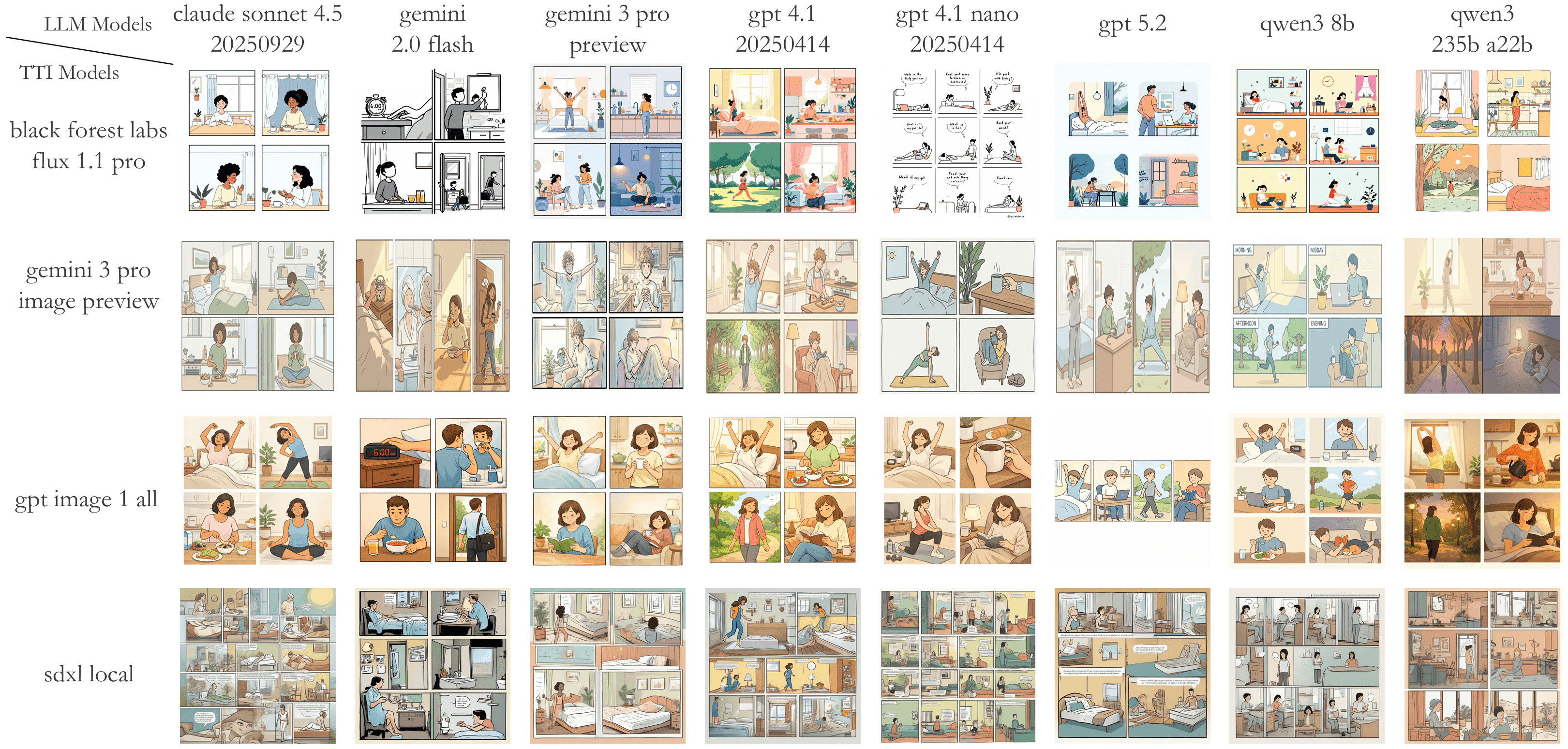}
  \caption{Qualitative generations for CO task \texttt{co\_106} from the skilled condition.}
    \label{fig:example_co}
\end{figure}

\textbf{Design Rationale.}
This task's challenge lies in many ways. Prompters must simultaneously (i) encode a multi-panel narrative, (ii) maintain consistent character identity and attributes across panels, (iii) control color usage globally, and (iv) specify exact captions. Typical failure modes include attribute leakage (e.g., shirt or alarm clock colors drifting in one panel; violating $C_1$), incorrect panel count or arrangement ($C_2$--$C_3$), missing or misspelled captions ($C_4$), or the model introducing extra colors or speech bubbles (breaching $C_5$). These errors directly reveal whether the prompter can systematically translate a structured, low-noise spec into a compact yet sufficiently over-specified prompt.

\textbf{Qualitative generations across models.}
Beyond the objective checklist in Table~\ref{tab:checklist_co106}, it is also informative to inspect how different prompter–backend combinations actually instantiate this four-panel comic; Figure~\ref{fig:example_co} visualizes these generations for task \texttt{co\_106}.

\subsection{IM Task Example}
\label{app:task_ex_im}

\begin{figure}[H]
    \centering
    \includegraphics[width=0.4\linewidth]{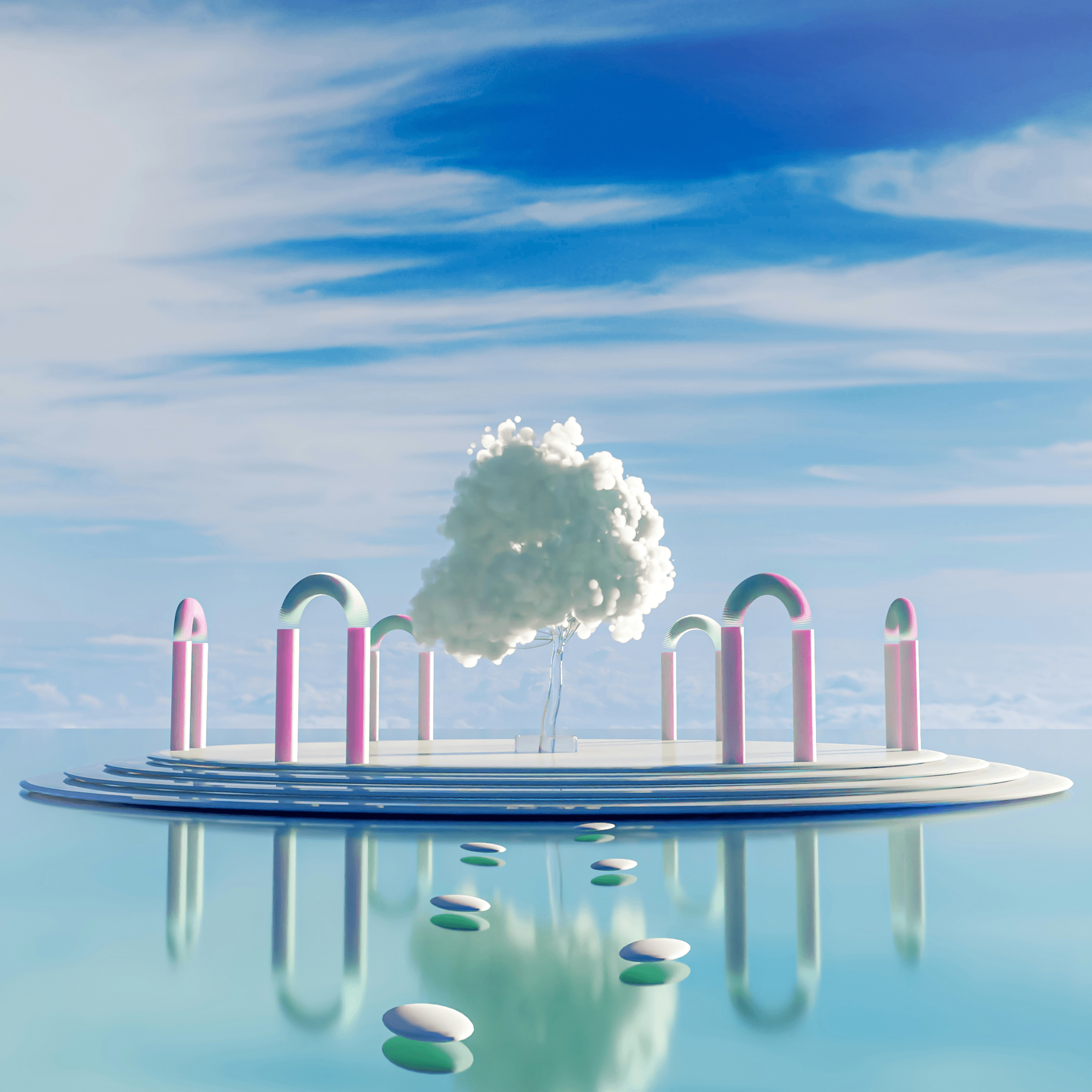}
    \caption{Target image for IM task \texttt{im\_43}. Prompters only see the image, not the internal seed prompt.}
    \label{fig:im_43}
\end{figure}

\textbf{Task Description.}
IM tasks assess cognition-oriented prompting: given only a target image, the prompter must encode perceptual information into text so that a T2I model can reproduce it. Figure~\ref{fig:im_43} shows the target image for task \texttt{im\_43}, which belongs to the \emph{Environment}, \emph{Object}, \emph{3D Render}, and \emph{Abstract Conceptual} categories.

\begin{tcolorbox}[breakable,colback=gray!10,colframe=gray!50,boxrule=0.5pt]
\textbf{Task ID:} \texttt{im\_43}\quad
\textbf{Instruction to prompters:}\\[0.3em]
\emph{You are given the target image in Figure~\ref{fig:im_43}. Write a single English prompt that would enable a text-to-image model to reproduce an image as close as possible to this target, including its composition, geometry, materials, lighting, and overall mood.}
\end{tcolorbox}

\begin{table}[H]
\centering
\small
\caption{Objective image- and prompt-based checklists for IM task \texttt{im\_43}.}
\label{tab:checklist_im43}

\begin{tabular}{@{}p{0.48\linewidth}|p{0.48\linewidth}@{}}
\toprule
\multicolumn{1}{c|}{\textbf{Image-based checklist}} &
\multicolumn{1}{c@{}}{\textbf{Prompt-based checklist}} \\
\midrule

I1.\ A floating platform or circular stage is present in the center of the image &
P1.\ The prompt mentions a floating platform or circular stage or related terms \\

I2.\ The platform has multiple vertical cylindrical pillars or columns standing on it &
P2.\ The prompt describes multiple vertical cylindrical pillars or columns or related terms \\

I3.\ The pillars have pink or magenta colored sections &
P3.\ The prompt specifies pink or magenta colored sections on pillars or related terms \\

I4.\ The pillars have teal or turquoise colored sections &
P4.\ The prompt mentions teal or turquoise colored sections on pillars or related terms \\

I5.\ Some pillars have curved or arched tops connecting two vertical sections &
P5.\ The prompt describes curved or arched tops connecting pillars or arch structures or related terms \\

I6.\ A large white cloud formation is positioned above the platform in the center &
P6.\ The prompt mentions a large white cloud above the platform or in the center or related terms \\

I7.\ A thin vertical line or string appears to connect the cloud to the platform &
P7.\ The prompt describes a thin line or string or connection between cloud and platform or related terms \\

I8.\ The platform appears to be floating above water or a reflective surface &
P8.\ The prompt specifies floating above water or reflective surface or related terms \\

I9.\ The platform has multiple horizontal circular tiers or layers &
P9.\ The prompt mentions multiple horizontal circular tiers or layers on platform or related terms \\

I10.\ The tiers are colored in blue and white tones &
P10.\ The prompt describes blue and white colored tiers or layers or related terms \\

I11.\ Reflections of the pillars are visible in the water below &
P11.\ The prompt mentions reflections in water or mirrored elements below or related terms \\

I12.\ Multiple small disc-shaped objects are floating in the water beneath the platform &
P12.\ The prompt describes small disc-shaped objects or floating discs in water or related terms \\

I13.\ The visual style is 3D rendered or surreal digital art &
P13.\ The prompt mentions 3D render or surreal digital art or related terms \\

I14.\ The color palette includes pastel tones of pink, blue, teal, and white &
P14.\ The prompt describes pastel color palette or pink blue teal white colors or related terms \\

I15.\ The composition is symmetrical with the platform centered &
P15.\ The prompt mentions symmetrical composition or centered platform or related terms \\

\bottomrule
\end{tabular}

\end{table}

\textbf{Challenge Primitive Decomposition.}
Unlike OE and CO, the challenge primitives here are instantiated visually rather than textually. Consistent with the IM design, audience intent $S_2$, semantic negation $S_4$, and hard constraints $C_5$ are absent.
\begin{itemize}[leftmargin=*]
    \item \textbf{Abstract Intent ($S_1$).} The image conveys a calm, surreal dreamscape with a ``cloud tree'' floating above a mirror-like sea. Prompters must infer and verbalize this atmosphere (e.g., serenity, dreaminess, futurism) from purely visual cues.
    \item \textbf{Implicit Style ($S_3$).} The pastel palette, smooth gradients, soft focus, and 3D-rendered look imply a minimalistic surreal style and high-resolution rendering, which are not named explicitly but must be captured (e.g., ``dreamy surrealism, pastel 3D render'').
    \item \textbf{Attribute Binding ($C_1$).} Several distinctive object--attribute bindings must be preserved: the cloud shaped like a tree sitting on a transparent glass pedestal, pastel pink and aqua arches arranged around a white platform, and smooth stepping stones leading toward the center.
    \item \textbf{Spatial Relation ($C_2$).} The composition is strongly center-focused and symmetrical, with circular, tiered platforms and clear foreground--midground--background separation, plus horizontal reflections in the water. Capturing these relationships is crucial for reproducing the scene.
    \item \textbf{Quantity ($C_3$).} While exact counts (e.g., number of arches or stepping stones) are not emphasized, the prompt must encode the presence of multiple repeated elements in a ring-like arrangement and a path of stepping stones, which serve as soft quantity cues.
\end{itemize}
These primitives together require the prompter to perform fine-grained visual analysis and encode it into text without any scaffolding from a textual brief.

We summarize the image-level and prompt-level criteria in 
Table~\ref{tab:checklist_im43}.

\textbf{Design Rationale and Difficulty.}
This task combines a relatively simple, symmetric global layout with several non-trivial perceptual details. Novice prompters often under-specify either the abstract mood (omitting the surreal, dreamlike quality; weakening $S_1$ and $S_3$) or the structural composition (forgetting the glass pedestal, circular tiers, or stepping-stone path; harming $C_1$--$C_2$). Skilled prompters, in contrast, tend to systematically decompose the image into objects, materials, spatial relations, and rendering style, then recombine them into a compact prompt that preserves both aesthetics and geometry. This behavior directly reflects the cognition-oriented encoding ability that IM tasks are designed to probe.

\begin{figure}[H]
    \centering
    \includegraphics[width=0.98\linewidth]{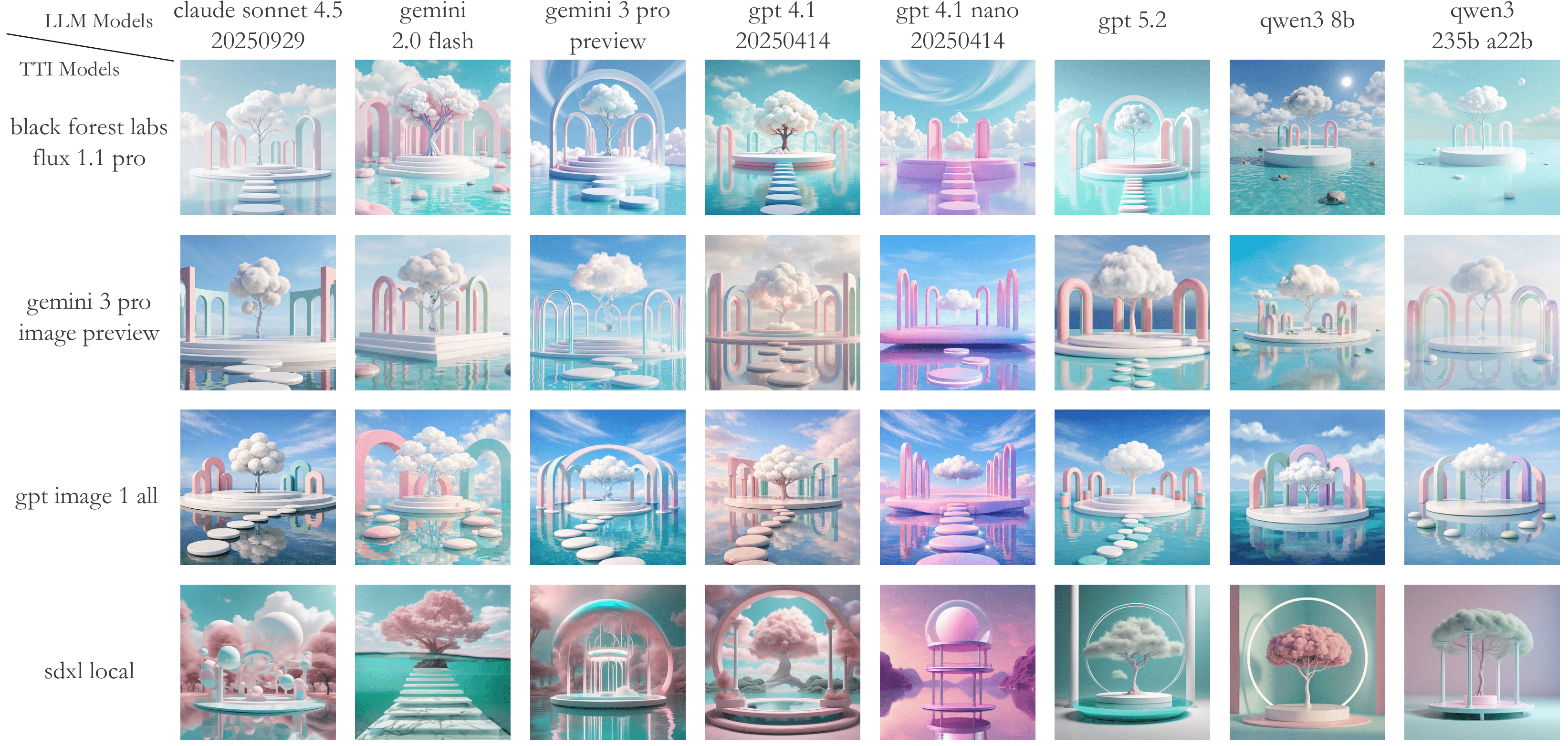}
  \caption{Qualitative generations for IM task \texttt{im\_43} from the novice condition.}
    \label{fig:example_im}
\end{figure}

\textbf{Qualitative generations across models.}
Finally, to complement the analysis of the IM instance, Figure~\ref{fig:example_im} shows how our full set of prompter–backend combinations attempt to recreate the target scene from Figure~\ref{fig:im_43}.

\section{Subjective Evaluation Dimensions}
\label{app:subjective_dimensions}

This section details the subjective evaluation dimensions used by \textsc{AtelierJudge} to assess the quality of prompt--image pairs, 4 dimensions for prompts and 4 dimensions for images. These dimensions are designed to capture perceptual, aesthetic, and semantic qualities that are difficult to verify through binary constraints, while remaining grounded in observable and model-agnostic criteria. All 8 dimensions are rated on a 1--5 Likert scale. Higher scores indicate stronger and more consistent satisfaction of the listed criteria, rather than merely increased verbosity or detail. The same scale and criteria are used for both human experts and AtelierJudge. Full scoring rubrics are provided in Appendix \ref{app:guideline}.

\subsection{Image-Level Dimensions}
\label{app:H1}

\textbf{Mood \& Atmosphere.}
This dimension evaluates whether the image conveys an emotional tone and overall atmosphere that aligns with the intent expressed or implied in the prompt. It focuses on the coherence between affective intent and visual realization, including whether the emotional quality (e.g., joyful, melancholic, tense, serene) is appropriate in strength and consistency. Visual elements such as color palette, lighting, composition, and texture are considered insofar as they jointly support the intended mood and communicate both explicit and implicit emotional cues.

\textbf{Visual Composition.}
This dimension assesses the structural quality of the image and the unity among its visual elements. Key considerations include the presence of a clear visual focus, effective layering and depth, and harmonious relationships between subject and background. The evaluator examines whether spatial arrangement, balance, and visual flow guide attention naturally, and whether multiple elements are integrated into a coherent whole through compositional principles such as alignment, repetition, contrast, and spatial relationships. In design-oriented tasks (e.g., logos or posters), emphasis is placed on meaningful integration rather than simple aggregation.

\textbf{Color \& Lighting.}
This dimension evaluates the harmony and appropriateness of color usage and lighting. Assessment includes whether the color scheme supports the intended atmosphere, whether saturation and brightness are well balanced, and whether lighting direction and intensity are visually and physically plausible. Shadows and highlights are examined for consistency and their contribution to depth, form, and mood, as well as the overall coordination between color and lighting.

\textbf{Technical Flawlessness.}
This dimension focuses on identifying common technical artifacts and visual defects characteristic of generative models. Evaluators look for issues such as anatomical distortions (especially of hands and faces), implausible physical structures, incorrect perspective, unnatural textures or repetitive patterns, blurred or fused object boundaries, and other noticeable generation artifacts. The absence of such flaws indicates higher technical soundness, independent of stylistic or aesthetic preference.

\subsection{Prompt-Level Dimensions}
\label{app:H2}

\textbf{Instructional Clarity.}
This dimension evaluates the linguistic clarity and executability of the prompt. It considers whether the prompt is grammatically complete, logically coherent, and free from ambiguity that could lead to misinterpretation. Clear identification of the main subject, stable descriptions without internal contradictions, and well-structured sentences that convey intent unambiguously are key factors. The goal is to assess whether the prompt can be reliably understood and acted upon by a text-to-image system without requiring guesswork.

\textbf{Creative Elaboration.}
This dimension assesses the richness and specificity of visual information provided in the prompt. Evaluation focuses on whether the prompt goes beyond minimal descriptions to include concrete details about subjects, environments, materials, lighting, atmosphere, or composition. Prompts that demonstrate imagination through distinctive, sensory-rich descriptions are rated higher than those that remain generic or underspecified.

\textbf{Terminology Proficiency.}
This dimension evaluates the prompter's ability to appropriately use domain-relevant visual terminology in a model-agnostic manner. This includes accurate and consistent use of concepts from photography, visual arts, and rendering (e.g., depth of field, impressionist style, volumetric lighting), without internal contradictions. Terminology should be precise, non-conflicting, and well matched to the task intent. The evaluation explicitly discourages reliance on model-specific syntax or engine-dependent keywords, emphasizing transferable visual literacy over system-specific tricks.

\textbf{Intent Formalization.}
This dimension assesses the prompter's ability to translate abstract, implicit, or high-level intent into concrete visual specifications. Evaluators examine whether abstract emotions, concepts, audience requirements, or implicit stylistic expectations are effectively grounded in describable visual elements, styles, or compositional choices. High-quality prompts avoid passing vague abstractions directly to the model and instead operationalize them into visual cues that faithfully serve the original intent.

\section{Evaluation Exemplar Memory Construction}
\label{app:exemplar}

\begin{figure}[H]
    \centering
    \includegraphics[width=0.9\linewidth]{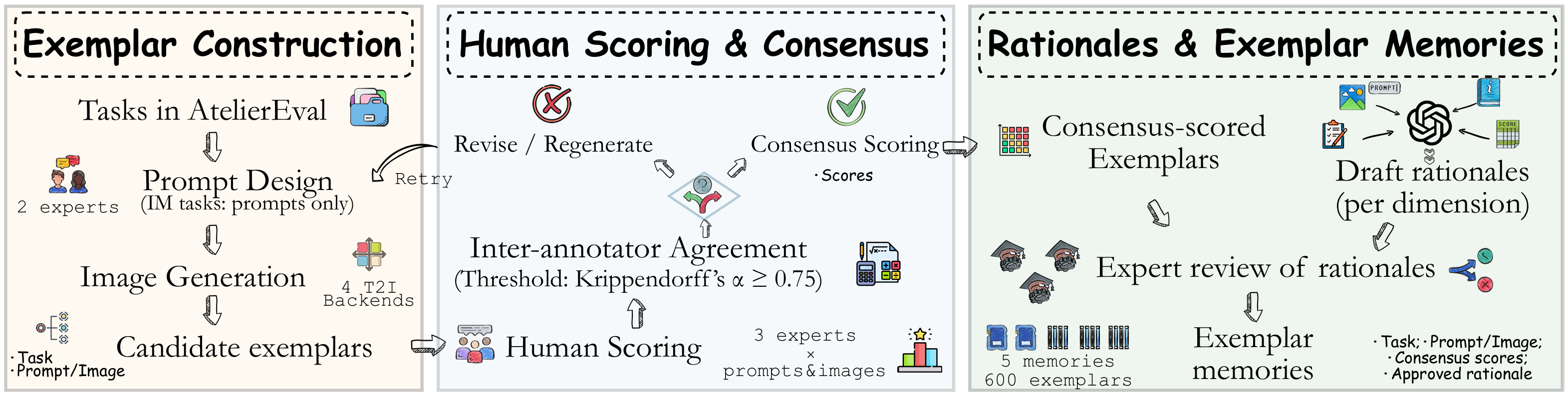}
  \caption{Overview of the memory construction pipeline.}
    \label{fig:memory}
\end{figure}

This appendix details the construction and quality control of the exemplar memories used by the subjective evaluation skills of AtelierJudge. All exemplars are constructed and fixed prior to the main experiments to ensure consistency and reproducibility. An overview of the exemplar memory construction and quality-control pipeline is shown in Figure~\ref{fig:memory}.

For each task in \textbf{\texttt{AtelierEval}}, we construct two evaluation exemplars, one for the prompt and one for the image (except for IM tasks). Each exemplar consists of 4 components: \ding{182} the task directly from \textbf{\texttt{AtelierEval}}, \ding{183} the prompt or generated image used for evaluation, \ding{184} human-annotated scores across subjective dimensions, and \ding{185} brief rationales explaining the assigned scores. Prompts are collaboratively authored by two experts to ensure broad coverage of different subjective quality patterns and score ranges. In total, 360 prompts are created. For each prompt, one of the four evaluated T2I backends is used to generate a corresponding image, yielding 360 images in total. For IM tasks, only prompts are retained, while generated images are used solely to verify prompt feasibility.

After image generation, we construct subjective human-labeled scores for each exemplar. Three experts independently assign scores to each prompt and generated image as two mutually independent evaluation targets, following the same subjective scoring guidelines as AtelierJudge (Appendix~\ref{app:guideline}). To ensure balanced coverage, the exemplar set is curated such that, for each task category, every subjective dimension contains representative exemplars at all score levels~\cite{qin2025enhancing}.

We assess inter-annotator agreement (IAA) using Krippendorff’s $\alpha$ \cite{krippendorff2018content}, which is suitable for ordinal data. Only exemplars meeting the predefined agreement threshold ($\alpha \geq 0.75$) are retained. This threshold typically allows one expert’s score to differ from the other two by at most one scale point. Exemplars failing the criterion are discarded, and the corresponding prompts are rewritten or images regenerated. For retained exemplars, the three experts further discuss them to reach a final consensus score. Exemplars for which consensus cannot be reached are also discarded. Final scores stored in the exemplar memories are determined by expert consensus rather than numerical averaging. 

After the consensus scores are finalized, we attach rationales to each exemplar, with one rationale corresponding to each subjective dimension. For scalability, rationales are generated post-hoc by GPT-5.2. For each dimension, the model is provided with a structured chain-of-thought prompt containing (i) the task, (ii) the prompt or generated image, (iii) the scoring guideline for the dimension, and (iv) the finalized expert score. The model is instructed to produce a concise, criterion-grounded explanation that justifies the given score (“knowing the answer and explaining why”). Although the rationale drafter is implemented using GPT-5.2 and may introduce stylistic self‑consistency, all rationales are generated post‑hoc given fixed expert consensus scores. To further mitigate such bias, before being admitted to the exemplar memory, each generated rationale is reviewed and approved by three human experts through a joint deliberative process. During this review, experts verify that the rationale faithfully reflects the scoring rubric, accurately explains the assigned score, and does not introduce extraneous or misleading information. Only rationales that pass expert review are stored together with the finalized scores and subsequently used for memory-augmented evaluation. Through this process, we obtain 600 high-quality exemplars across five memories to support the subjective evaluation skills of AtelierJudge.

\section{Details of Safety Filter Skill}
\label{app:safety}

AtelierJudge includes a safety filtering design, which serves as non-scoring gates to ensure ethical compliance. Safety filtering consists of two skills operating on different modalities: a QA-based prompt safety filter skill and a VQA-based image safety filter skill, which are executed sequentially in the skill routing process but remain logically decoupled.

\begin{itemize}[leftmargin=*,
    topsep=-0.5em,
    partopsep=0pt,
    parsep=0pt,
    itemsep=0pt
    ]

\item[\ding{224}] \textbf{Prompt Safety Filter Skill.}
Before image generation, all prompts are checked by this skill, which is implemented using GPT-5.2. This skill operates solely on the prompt and detects standard categories of unsafe content (e.g., violence, sexual content, hate, child safety violations, and illegal activities). Across all experiments, only 2 prompts out of 7,200 total prompts were initially flagged and later manually confirmed to be safe.

\item[\ding{224}] \textbf{Image safety filter skill.}
As stated in Section \ref{sec:exp_set}, we evaluate four T2I backends, and three of them are accessed via provider APIs that already include safety mechanisms. Among all submissions, only Flux-Pro refused image generation for 5 prompts out of the 7,200 prompts. All five prompts were subsequently confirmed to be safe by human experts. To enforce a unified standard across backends, an image safety filter skill—also implemented via GPT-5.2—is applied to all generated images, regardless of the backend. This skill uses the same set of safety categories as the prompt safety filter skill. No generated image was flagged as unsafe by the unified image safety filter skill.

\end{itemize}

Submissions that cannot produce an image are excluded from all aggregate metrics. The extremely low trigger rate of the safety filter skills indicates that the expert-designed tasks are well aligned with content safety requirements and that these skills function as non-intrusive gating components rather than confounding factors in our experiments. For reproducibility, the full prompts and code used to implement the safety filter skills are open-sourced in our repository.

\section{Scoring Guidelines for Subjective Evaluation}
\label{app:guideline}

This appendix presents the detailed scoring rubrics used to assess subjective quality in \textbf{\texttt{AtelierEval}}. To ensure a unified evaluation standard, these rubrics are employed identically by both human experts during the construction of the exemplar memory, and AtelierJudge during the automated agentic evaluation process.

\begin{tcolorbox}[breakable,colback=white,colframe=black,boxrule=0.5pt,
title=\textbf{Prompt Evaluation Guidelines}]

\textbf{1. Instructional Clarity} (Grammar, logic, lack of ambiguity, structure)
\begin{itemize}[leftmargin=*, noitemsep, topsep=0pt]
  \item \textbf{1 (Failure):} Incoherent, contradictory, or grammatically broken. Impossible to execute reliably.
  \item \textbf{2 (Poor):} Significant confusion or conflicting instructions. Logic is hard to follow.
  \item \textbf{3 (Acceptable):} Understandable but contains minor ambiguities or loose sentence structure. Requires some model guesswork.
  \item \textbf{4 (Good):} Generally clear and well-structured. Only very minor phrasing issues.
  \item \textbf{5 (Excellent):} Perfectly structured, logical in flow, and unambiguous. Explicitly identifies subjects and relations. Zero chance of misinterpretation.
\end{itemize}

\textbf{2. Creative Elaboration} (Richness, detail, sensory specificity)
\begin{itemize}[leftmargin=*, noitemsep, topsep=0pt]
  \item \textbf{1 (Empty):} Bare minimum description. Lacks any detail beyond the core subject.
  \item \textbf{2 (Generic):} Uses clichéd descriptions. Details are vague (e.g., "nice background").
  \item \textbf{3 (Basic):} Provides standard details (color, size) but lacks imagination or sensory depth.
  \item \textbf{4 (Detailed):} Good use of adjectives and specific descriptions. Sets a clear scene.
  \item \textbf{5 (Rich):} Highly evocative. Describes textures, atmosphere, materials, and specific nuances. Demonstrates strong imagination.
\end{itemize}

\textbf{3. Terminology Proficiency} (Use of visual/artistic vocabulary, model-agnosticism)
\begin{itemize}[leftmargin=*, noitemsep, topsep=0pt]
  \item \textbf{1 (Poor):} Uses wrong terms, relies on ``magic words'' (e.g., ``4k'', ``trending''), or non-visual text.
  \item \textbf{2 (Naive):} Uses artistic terms incorrectly or relies heavily on engine-specific syntax (e.g., --v 6.0) inside the text.
  \item \textbf{3 (Average):} Uses basic visual terms correctly (e.g., ``oil painting'', ``sunset'').
  \item \textbf{4 (Advanced):} Uses more complex terms (e.g., ``macro lens'', ``impasto'') correctly.
  \item \textbf{5 (Expert):} Precise use of domain-specific vocabulary (e.g., ``volumetric lighting'', ``depth of field'', ``chiaroscuro'') correctly and effectively.
\end{itemize}

\textbf{4. Intent Formalization} (Translating abstract goals into concrete visual specs)
\begin{itemize}[leftmargin=*, noitemsep, topsep=0pt]
  \item \textbf{1 (Abstract):} Pastes abstract concepts (e.g., ``a sad vibe'') directly without visual translation.
  \item \textbf{2 (Mostly Abstract):} Slight attempt at visual description, but mostly relies on the model to interpret feelings.
  \item \textbf{3 (Mixed):} Partially translates intent but relies on some abstract descriptions.
  \item \textbf{4 (Mostly Concrete):} Most abstract concepts are converted to visual cues, with minor gaps.
  \item \textbf{5 (Concrete):} Fully operationalizes abstract intent into observable visual elements (e.g., translates ``sadness'' into ``muted blue tones, rain-streaked windows, slumped posture'').
\end{itemize}

\end{tcolorbox}

\begin{tcolorbox}[breakable,colback=white,colframe=black,boxrule=0.5pt,
title=\textbf{Image Evaluation Guidelines}]

\textbf{1. Mood \& Atmosphere} (Emotional tone, consistency with intent)
\begin{itemize}[leftmargin=*, noitemsep, topsep=0pt]
  \item \textbf{1 (Mismatch):} The image conveys the completely wrong emotion or has no discernible atmosphere.
  \item \textbf{2 (Weak):} The mood is barely present or confusing.
  \item \textbf{3 (Generic):} The mood is somewhat aligned but weak or inconsistent. Lacks strong emotional impact.
  \item \textbf{4 (Strong):} The atmosphere is clear and mostly consistent with the intent.
  \item \textbf{5 (Evocative):} The atmosphere is palpable, consistent, and perfectly matches the intended emotional tone (e.g., tension, serenity).
\end{itemize}

\textbf{2. Visual Composition} (Structure, balance, focus, depth)
\begin{itemize}[leftmargin=*, noitemsep, topsep=0pt]
  \item \textbf{1 (Chaotic):} Cluttered, lacks a focal point, or poor spatial arrangement. Hard to parse.
  \item \textbf{2 (Unbalanced):} Elements feel randomly placed. Poor use of space.
  \item \textbf{3 (Standard):} Functional composition. Center-focused or basic rule-of-thirds, but lacks depth or dynamic flow.
  \item \textbf{4 (Good):} Clear focal point and good balance. Uses space well.
  \item \textbf{5 (Masterful):} Excellent use of depth, layering, and guiding lines. Visual elements are integrated harmoniously; the eye is led naturally.
\end{itemize}

\textbf{3. Color \& Lighting} (Harmony, direction, saturation, physics)
\begin{itemize}[leftmargin=*, noitemsep, topsep=0pt]
  \item \textbf{1 (Bad):} Clashing colors, flat lighting, or physically impossible shadows. Looks washed out or oversaturated.
  \item \textbf{2 (Dull):} Colors are muddy or lighting makes the subject hard to see.
  \item \textbf{3 (Passable):} Lighting is logical but flat. Colors are acceptable but not distinct or strictly harmonized.
  \item \textbf{4 (Cohesive):} Good color harmony and clear light source.
  \item \textbf{5 (Cinematic):} Lighting creates volume and mood. Color palette is sophisticated and cohesive. Shadows and highlights are physically accurate and aesthetic.
\end{itemize}

\textbf{4. Technical Flawlessness} (Artifacts, distortions, anatomy, rendering)
\begin{itemize}[leftmargin=*, noitemsep, topsep=0pt]
  \item \textbf{1 (Broken):} Severe artifacts (mangled hands, extra limbs), blurred boundaries, or distinct digital noise. Unusable.
  \item \textbf{2 (Obvious Flaws):} Distracting distortions or mutations are immediately visible.
  \item \textbf{3 (Minor Flaws):} Generally good, but contains noticeable small artifacts, slight perspective issues, or unnatural textures on close inspection.
  \item \textbf{4 (Clean):} High quality rendering with only negligible, hard-to-spot imperfections.
  \item \textbf{5 (Flawless):} Clean, crisp rendering. Anatomy, perspective, and textures appear natural and intentional. No visible generative artifacts.
\end{itemize}

\end{tcolorbox}
\section{Prompts for AtelierJudge}
\label{app:judge_prompts}
\newcommand{\placeholder}[1]{\texttt{\color{green!50!black}<<#1>>}}

\subsection{Subjective Skill Prompts}
\label{app:L1}

\begin{tcolorbox}[breakable,colback=white,colframe=black,boxrule=0.5pt,
title=\textbf{Prompt Subjective Skill Template}]

{\color{blue}
\textbf{System Prompt}

You are an expert Text-to-Image Prompt Engineer and Evaluator. Your goal is to assess the quality of a text prompt based on how effectively it translates a user's intent into an executable description.
You MUST follow these rules:
\begin{itemize}[leftmargin=*, topsep=0pt]
  \item Act as a calibrated judge using the provided Reference Exemplars.
  \item Evaluate strictly based on the provided 1-5 scale definitions.
  \item Output ONLY a valid JSON object containing scores and rationales.
\end{itemize}
}

\textbf{User Prompt:}

\rule{\textwidth}{0.4pt}
\textit{[Start of Exemplar 1]}

\textbf{Task:} \placeholder{EXEMPLAR\_1\_TASK}

\textbf{Prompt to Evaluate:} \placeholder{EXEMPLAR\_1\_PROMPT}

\placeholder{PROMPT\_EVALUATION\_GUIDELINES}

Step-by-Step Reasoning Instructions:
\begin{itemize}[leftmargin=*, noitemsep, topsep=0pt]
  \item Analyze the Task Description to understand the goal.
  \item Read the Candidate Prompt.
  \item For each dimension, assign a score (integer 1-5) and write a brief rationale explaining why.
\end{itemize}

\textbf{Output JSON:} \placeholder{EXEMPLAR\_1\_JSON}

\textit{[End of Exemplar 1]}

...

\textit{[Start of Exemplar k]}

\textbf{Task:} \placeholder{EXEMPLAR\_k\_TASK}

\textbf{Prompt to Evaluate:} \placeholder{EXEMPLAR\_k\_PROMPT}

\textbf{Evaluation Guidelines:}

Evaluate the Candidate Prompt on the following 1-5 scale. Read the definitions carefully.

\placeholder{PROMPT\_EVALUATION\_GUIDELINES}

Step-by-Step Reasoning Instructions:
\begin{itemize}[leftmargin=*, noitemsep, topsep=0pt]
  \item Analyze the Task Description to understand the goal.
  \item Read the Candidate Prompt.
  \item For each dimension, assign a score (integer 1-5) and write a brief rationale explaining why.
\end{itemize}

\textbf{Output JSON:} \placeholder{EXEMPLAR\_k\_JSON}

\textit{[End of Exemplar k]}

\rule{\textwidth}{0.4pt}

\textbf{Task:}
\placeholder{TASK}

\textbf{Prompt to Evaluate:}
\placeholder{PROMPT}

\textbf{Evaluation Guidelines:}

Evaluate the Candidate Prompt on the following 1-5 scale. Read the definitions carefully.

\placeholder{PROMPT\_EVALUATION\_GUIDELINES}

Step-by-Step Reasoning Instructions:
\begin{itemize}[leftmargin=*, noitemsep, topsep=0pt]
  \item Analyze the Task to understand the goal.
  \item Read the Prompt.
  \item For each dimension, assign a score (integer 1-5) and write a brief rationale explaining why.
\end{itemize}

\textbf{Output JSON:}
\end{tcolorbox}

\begin{tcolorbox}[breakable,colback=white,colframe=black,boxrule=0.5pt,
title=\textbf{Image Subjective Skill Template}]

{\color{blue}
\textbf{System Prompt}

You are an expert Art Director and Visual Critic. Your goal is to assess the quality of a generated image based on its aesthetic value, technical execution, and alignment with visual intent.
You MUST follow these rules:
\begin{itemize}[leftmargin=*, topsep=0pt]
 \item Act as a calibrated judge using the provided Reference Exemplars.
 \item Evaluate strictly based on the provided 1-5 scale definitions.
 \item Output ONLY a valid JSON object containing scores and rationales.
\end{itemize}
}

\textbf{User Prompt:}

\rule{\textwidth}{0.4pt}
\textit{[Start of Exemplar 1]}

\textbf{Task:} \placeholder{EXEMPLAR\_1\_TASK}

\textbf{Image to Evaluate:} \placeholder{EXEMPLAR\_1\_IMAGE\_INPUT}

\textbf{Evaluation Guidelines:}

Evaluate the Candidate Image on the following 1-5 scale. Read the definitions carefully.

\placeholder{IMAGE\_EVALUATION\_GUIDELINES}

Step-by-Step Reasoning Instructions:
\begin{itemize}[leftmargin=*, noitemsep, topsep=0pt]
\item Analyze the Task Description to understand the goal.
\item Observe the Candidate Image.
\item For each dimension, assign a score (integer 1-5) and write a brief rationale explaining why.
\end{itemize}

\textbf{Output JSON:} \placeholder{EXEMPLAR\_1\_JSON}

\textit{[End of Exemplar 1]}

...

\textit{[Start of Exemplar k]}

\textbf{Task:} \placeholder{EXEMPLAR\_k\_TASK}

\textbf{Image to Evaluate:} \placeholder{EXEMPLAR\_k\_IMAGE\_INPUT}

\textbf{Evaluation Guidelines:}

Evaluate the Candidate Image on the following 1-5 scale. Read the definitions carefully.

\placeholder{IMAGE\_EVALUATION\_GUIDELINES}

Step-by-Step Reasoning Instructions:
\begin{itemize}[leftmargin=*, noitemsep, topsep=0pt]
 \item Analyze the Task to understand the goal.
 \item Observe the Image.
 \item For each dimension, assign a score (integer 1-5) and write a brief rationale explaining why.
\end{itemize}

\textbf{Output JSON:} \placeholder{EXEMPLAR\_k\_JSON}

\textit{[End of Exemplar k]}

\rule{\textwidth}{0.4pt}

\textbf{Task:}
\placeholder{TASK}

\textbf{Image to Evaluate:}
\placeholder{IMAGE\_INPUT}

\textbf{Evaluation Guidelines:}

Evaluate the Candidate Image on the following 1-5 scale. Read the definitions carefully.

\placeholder{IMAGE\_EVALUATION\_GUIDELINES}

Step-by-Step Reasoning Instructions:
\begin{itemize}[leftmargin=*, noitemsep, topsep=0pt]
 \item Analyze the Task Description to understand the goal.
 \item Observe the Candidate Image.
 \item For each dimension, assign a score (integer 1-5) and write a brief rationale explaining why.
\end{itemize}

\textbf{Output JSON:}
\end{tcolorbox}

\subsection{Objective Skill Prompts}
\label{app:L2}
\begin{tcolorbox}[breakable,colback=white,colframe=black,boxrule=0.5pt,
title=\textbf{Prompt Objective Skill Template}]

{\color{blue}
\textbf{System Prompt:}

You are a strict prompt checklist evaluator.

You MUST follow these rules:
\begin{itemize}[leftmargin=*]
  \item You MUST evaluate EACH checklist item independently.
  \item You MUST base your judgment ONLY on the given prompt text.
  \item You MUST output ONLY a valid JSON object.
  \item JSON keys MUST EXACTLY match the checklist text (character-for-character).
  \item JSON values MUST be ONLY 0 or 1.
  \item You MUST NOT include explanations, comments, or any text outside the JSON.
  \item If a requirement is NOT clearly and explicitly specified in the prompt, you MUST output 0.
\end{itemize}
}

\textbf{User Prompt:}

You are given a text prompt for image generation and a checklist of requirements.
Your task is to determine whether the prompt clearly specifies EVERY requirement.

\medskip
\textbf{Prompt:}

\placeholder{PROMPT}

\medskip
\textbf{Checklist (evaluate each item one by one):}

\placeholder{CHECKLIST}

\medskip
Instructions:
\begin{enumerate}[leftmargin=*]
  \item Go through EVERY checklist item. Do not skip any item.
  \item For each item, output 1 if the requirement is clearly and explicitly specified in the prompt; otherwise output 0.
  \item Your final answer MUST be ONLY a single valid JSON object.
  \item Do NOT add, remove, or modify any checklist item text.
\end{enumerate}

\medskip
\textbf{Required Output Format (example):}

\{
  ``Checklist item text 1'': 1,\\
  ``Checklist item text 2'': 0
\}

\medskip
Now read the prompt carefully and output ONLY the JSON object.
\end{tcolorbox}

\begin{tcolorbox}[breakable,colback=white,colframe=black,boxrule=0.5pt,
title=\textbf{Image Objective Skill Template}]

\textbf{System Prompt:}
{\color{blue}
You are a strict visual checklist evaluator.

You MUST follow these rules:
\begin{itemize}[leftmargin=*]
  \item You MUST evaluate EACH checklist item independently.
  \item You MUST base your judgment ONLY on the given image.
  \item You MUST output ONLY a valid JSON object.
  \item JSON keys MUST EXACTLY match the checklist text (character-for-character).
  \item JSON values MUST be ONLY 0 or 1.
  \item You MUST NOT include explanations, comments, or any text outside the JSON.
  \item If a requirement is NOT clearly and fully satisfied, you MUST output 0.
\end{itemize}
}

\textbf{User Prompt:}

You are given an image generated by a text-to-image model and a checklist of visual requirements.
Your task is to determine whether the image satisfies EVERY checklist item.

\medskip
\textbf{Image:}

\placeholder{IMAGE}

\medskip
\textbf{Checklist (evaluate each item one by one):}

\placeholder{CHECKLIST}

\medskip
Instructions:
\begin{enumerate}[leftmargin=*]
  \item Go through EVERY checklist item. Do not skip any item.
  \item For each item, output 1 if the requirement is clearly satisfied by the image; otherwise output 0.
  \item Your final answer MUST be ONLY a single valid JSON object.
  \item Do NOT add, remove, or modify any checklist item text.
\end{enumerate}

\medskip
\textbf{Required Output Format (example):}

\{
  ``Checklist item text 1'': 1,\\
  ``Checklist item text 2'': 0
\}

\medskip
Now examine the image and output ONLY the JSON object.
\end{tcolorbox}

\section{Model Hyperparameters}
\label{app:para}

\textbf{MLLMs.}
The following hyperparameters are shared by all MLLMs used in our experiments, both when acting as prompters and as the evaluator in \textbf{\texttt{AtelierJudge}}. All models are called by API, utilizing the newest version on Dec. 29, 2025.
\begin{tcolorbox}[
    breakable,
    colback=white,
    colframe=black,
    boxrule=0.5pt,
    title=\textbf{Multimodal LLM Hyperparameters (Prompter \& Evaluator)}
]
\begin{itemize}[leftmargin=*,topsep=2pt,itemsep=1pt,parsep=0pt]
    \item Context window cap: 16{,}384 tokens;
    \item Temperature: 0.0;
    \item Top-$p$: 0.7;
    \item Maximum output tokens: 4{,}096.
\end{itemize}
\end{tcolorbox}

\textbf{T2I Models.}
We provide the key configuration details for the locally deployed SDXL. Notably, due to the CLIP-based text encoder, SDXL operates with a maximum context length of 77 tokens, which may result in truncation for some prompts. For the remaining T2I models which are all accessed through DALLE-style APIs, no hyperparameters are exposed to the user except the image resolution (1024$\times$1024).
\begin{tcolorbox}[
    breakable,
    colback=white,
    colframe=black,
    boxrule=0.5pt,
    title=\textbf{Text-to-Image Hyperparameters (SDXL)}
]
\begin{itemize}[leftmargin=*,topsep=2pt,itemsep=1pt,parsep=0pt]
    \item Model: SDXL-base-1.0;
    \item Image size: 1024$\times$1024;
    \item Sampler / scheduler: Euler;
    \item Inference steps: 30;
    \item Guidance scale (CFG): 7.5;
    \item Precision: FP16;
    \item Refiner: None;
    \item Batch Size: 4;
    \item Random seed: independently sampled per image.
\end{itemize}
\end{tcolorbox}


\section{Prompts for MLLM Novice and Skilled Conditions}
\label{app:mllm_prompts}

This appendix documents the prompts used to instantiate the novice and skilled MLLM prompter conditions in our experiments.  
The novice condition uses minimal task instructions, whereas the skilled condition employs structured system prompts that explicitly encode high-level cognitive strategies, chain-of-thought \cite{wei2022chain,kojima2022large} style reasoning and self-verification steps. We present the exact prompts below for reproducibility.

\subsection{Novice MLLM Prompts}
\label{app:N1}

\begin{tcolorbox}[breakable,colback=white,colframe=black,boxrule=0.5pt,
title=\textbf{Novice Group Prompt Template (OE / CO)}]
Convert the following request into an image generation prompt. Use natural language only.

\textbf{Request:}  
\placeholder{TASK}
\end{tcolorbox}

\begin{tcolorbox}[breakable,colback=white,colframe=black,boxrule=0.5pt,
title=\textbf{Novice Group Prompt Template (IM)}]
Write an image generation prompt that would produce an image similar to the one shown. Use natural language only.

\textbf{Image:}  
\placeholder{IMAGE}
\end{tcolorbox}

\subsection{Skilled MLLM Prompts}
\label{app:N2}
\begin{tcolorbox}[breakable,colback=white,colframe=black,boxrule=0.5pt,
title=\textbf{Skilled Group Prompt Template (OE)}]

\textbf{System Prompt:}

{\color{blue}
You are an expert at translating creative concepts into vivid, detailed prompts for text-to-image generation systems.
}

\textbf{User Prompt:}

\textbf{Your Task}

Convert the user's request into an evocative, well-structured image generation prompt using natural language only. Structured formats such as lists are allowed.

\medskip
\textbf{Professional Approach}

\begin{enumerate}[leftmargin=*,topsep=2pt,itemsep=1pt,parsep=0pt]

\item \textbf{Extract Core Intent from Narrative}
\begin{itemize}[leftmargin=*,topsep=1pt,itemsep=0pt,parsep=0pt]
    \item The request may contain contextual noise (e.g., ``for next season,'' ``the boss mentioned'').
    \item Identify what truly matters for visual output vs.\ background context.
    \item Look for implicit requirements: target audience, emotional tone, abstract concepts.
\end{itemize}

\item \textbf{Translate Abstract to Concrete}
\begin{itemize}[leftmargin=*,topsep=1pt,itemsep=0pt,parsep=0pt]
    \item Abstract emotions (e.g., ``loneliness,'' ``eco-friendly'') $\rightarrow$ specific visual metaphors.
    \item Audience-based descriptions (e.g., ``for veterans,'' ``professional'') $\rightarrow$ appropriate style choices.
    \item Implicit styles (e.g., ``logo,'' ``poster'') $\rightarrow$ concrete artistic techniques.
    \item Negative constraints (e.g., ``no modern elements'') $\rightarrow$ explicit exclusions.
\end{itemize}

\item \textbf{Build Rich Visual Descriptions}
\begin{itemize}[leftmargin=*,topsep=1pt,itemsep=0pt,parsep=0pt]
    \item Describe the scene or subject with sensory detail.
    \item Establish clear focal points and visual hierarchy.
    \item Specify artistic style, medium, or aesthetic approach.
    \item Include atmospheric elements: lighting, color palette, mood.
    \item Add quality and technical specifications as appropriate.
\end{itemize}

\item \textbf{Self-Verification}
\begin{itemize}[leftmargin=*,topsep=1pt,itemsep=0pt,parsep=0pt]
    \item Check: Have I captured all implicit requirements?
    \item Check: Does this prompt create a compelling, coherent vision?
    \item Check: Is there enough detail to guide generation?
\end{itemize}

\end{enumerate}

\medskip
\textbf{Output Format}

Provide ONLY the image generation prompt. Do not include explanations or meta-commentary.

\medskip
Now, convert the following request into an image generation prompt:

\placeholder{TASK}

\end{tcolorbox}

\begin{tcolorbox}[breakable,colback=white,colframe=black,boxrule=0.5pt,
title=\textbf{Skilled Group Prompt Template (CO)}]

\textbf{System Prompt:}

{\color{blue}
You are an expert at translating precise technical requirements into accurate, executable prompts for text-to-image generation systems.
}

\textbf{User Prompt:}

\textbf{Your Task}

Convert the user's request into a clear, detailed image generation prompt using natural language only. Structured formats such as lists are allowed.

\medskip
\textbf{Professional Approach}

\begin{enumerate}[leftmargin=*,topsep=2pt,itemsep=1pt,parsep=0pt]

\item \textbf{Analyze Requirements Thoroughly}
\begin{itemize}[leftmargin=*,topsep=1pt,itemsep=0pt,parsep=0pt]
    \item Identify the core subject and purpose.
    \item Note ALL specified constraints and requirements.
    \item Recognize any negative constraints (what must NOT appear).
    \item Understand the visual style or context implied.
\end{itemize}

\item \textbf{Prioritize Precision and Completeness}
\begin{itemize}[leftmargin=*,topsep=1pt,itemsep=0pt,parsep=0pt]
    \item Every stated constraint is a REQUIREMENT, not a suggestion.
    \item Pay special attention to exact colors, positions, and quantities.
    \item Ensure specific text content and placement are correctly specified.
    \item Maintain correct attribute bindings and spatial relationships.
    \item Explicitly state exclusions (what should NOT be included).
\end{itemize}

\item \textbf{Structure Your Prompt Logically}
\begin{itemize}[leftmargin=*,topsep=1pt,itemsep=0pt,parsep=0pt]
    \item Start with the main subject and setting.
    \item Describe the visual style and medium.
    \item Detail spatial arrangements and compositions.
    \item Specify exact colors, text, quantities, and attributes.
    \item Explicitly state any exclusions or negative constraints.
    \item Include quality and technical specifications.
\end{itemize}

\item \textbf{Self-Verification Checklist}
\begin{itemize}[leftmargin=*,topsep=1pt,itemsep=0pt,parsep=0pt]
    \item Check: Have I addressed EVERY specified constraint?
    \item Check: Are colors, numbers, and positions exact?
    \item Check: Is all required text content included correctly?
    \item Check: Have I stated what should NOT appear?
    \item Check: Is the language clear and unambiguous?
\end{itemize}

\end{enumerate}

\medskip
\textbf{Output Format}

Provide ONLY the image generation prompt. Do not include explanations or meta-commentary.

\medskip
Now, convert the following request into an image generation prompt:

\placeholder{TASK}

\end{tcolorbox}

\begin{tcolorbox}[breakable,colback=white,colframe=black,boxrule=0.5pt,
title=\textbf{Skilled Group Prompt Template (IM)}]

\textbf{System Prompt:}

{\color{blue}
You are an expert at analyzing images and generating precise descriptive prompts for text-to-image generation systems.
}

\textbf{User Prompt:}

\textbf{Your Task}

Carefully analyze the provided image and convert it into a detailed, accurate image generation prompt using natural language only. Structured formats such as lists are allowed.

\medskip
\textbf{Professional Approach}

\begin{enumerate}[leftmargin=*,topsep=2pt,itemsep=1pt,parsep=0pt]

\item \textbf{Systematic Visual Analysis}
\begin{itemize}[leftmargin=*,topsep=1pt,itemsep=0pt,parsep=0pt]
    \item Overall composition: What is the main subject? How is the scene organized?
    \item Style and medium: What artistic style, rendering technique, or photographic approach?
    \item Atmosphere and mood: What emotional tone or aesthetic does it convey?
    \item Technical aspects: Lighting, perspective, color palette, depth of field.
\end{itemize}

\item \textbf{Precise Element Identification}
\begin{itemize}[leftmargin=*,topsep=1pt,itemsep=0pt,parsep=0pt]
    \item Count carefully: How many of each object? (Avoid counting errors.)
    \item Attributes matter: Which object has which color/property? (Avoid attribute confusion.)
    \item Spatial relationships: What is where? (Left/right, foreground/background, on/under.)
    \item Text content: If text appears, what does it say exactly? (OCR carefully.)
    \item Fine details: Materials, textures, expressions, accessories.
\end{itemize}

\item \textbf{Describe Systematically}
\begin{itemize}[leftmargin=*,topsep=1pt,itemsep=0pt,parsep=0pt]
    \item Start with subject and main action (if any).
    \item Describe artistic style, medium, and rendering approach.
    \item Detail specific elements with their attributes (colors, materials, states).
    \item Specify spatial arrangement and composition.
    \item Capture atmosphere, lighting, and mood.
    \item Note any text, symbols, or graphic elements.
    \item Include quality and technical characteristics.
\end{itemize}

\item \textbf{Self-Verification Against Image}
\begin{itemize}[leftmargin=*,topsep=1pt,itemsep=0pt,parsep=0pt]
    \item Check: Does my description match what I actually see?
    \item Check: Have I counted objects correctly?
    \item Check: Are attributes bound to the right objects?
    \item Check: Have I captured the style and mood accurately?
    \item Check: Is any text transcribed correctly?
\end{itemize}

\end{enumerate}

\medskip
\textbf{Output Format}

Provide ONLY the image generation prompt. Do not include explanations or meta-commentary.

\medskip
Now, analyze the provided image and convert it into an image generation prompt:

\placeholder{IMAGE}

\end{tcolorbox}

\section{Participant Selection \& Statistics}
\label{app:demographic}

This section describes how human participants were selected for the study, and reports summary statistics of the final participant pool. We detail the recruitment channels, inclusion criteria, group assignment procedure, and aggregated demographic characteristics of the participants. The detailed informed consent form and screening questions for participant selection are provided in Appendix~\ref{apx:human_study_materials}.

\subsection{Participant Selection Criteria}
\label{app:O1}
Human participants were recruited on a voluntary basis through multiple channels, including personal contacts, online advertisements on social media, and announcements on university forums. To ensure competency with experimental tools and minimize learning effects related to tool unfamiliarity, all participants were required to meet the following eligibility criteria:

\begin{tcolorbox}[breakable,colback=white,colframe=black,boxrule=0.5pt,title=\textbf{Eligibility Criteria}]
\begin{itemize}[leftmargin=*]
    \item At least 18 years of age;
    \item Proficiency in reading and writing English sufficient to understand task instructions;
    \item Ability to complete the study using a desktop or laptop environment with stable internet access.
\end{itemize}
\end{tcolorbox}

From the initial pool of applicants, we selected 24 participants for the novice group and 24 participants for the skilled group. We applied group-specific screening criteria to ensure a clear separation between novice and skilled prompters. These criteria are intended to operationalize working definitions of novice and skilled prompters solely for the purpose of this benchmark. Both groups can reliably complete the tasks without additional training, while differing in the extent of their prompting experience and systematic control strategies.

\begin{tcolorbox}[breakable,colback=white,colframe=black,boxrule=0.5pt,title=\textbf{Novice Group Screening Criteria}]
\begin{itemize}[leftmargin=*]
    \item Limited prior exposure to text-to-image generation tools, typically involving exploratory use;
    \item Ability to perform basic prompt-based image generation (e.g., entering a textual description and triggering generation), without systematic experience in prompt engineering;
    \item No regular use of advanced controls, model parameters, or structured prompting workflows.
\end{itemize}
\end{tcolorbox}

\begin{tcolorbox}[breakable,colback=white,colframe=black,boxrule=0.5pt,title=\textbf{Skilled Group Screening Criteria}]
\begin{itemize}[leftmargin=*]
    \item Sustained and regular use of text-to-image generation tools over an extended period;
    \item Familiarity with common prompting strategies and techniques for controlling visual attributes (e.g., composition, lighting, style, or subject binding);
    \item Demonstrated awareness of technical or artistic control dimensions in text-to-image generation, as assessed by the pre-test questionnaire.
\end{itemize}
\end{tcolorbox}

Candidates who applied for the skilled group were required to provide evidence of prior work to support their self-reported expertise. This typically consisted of example prompts and corresponding generated images or links to publicly available portfolios voluntarily shared by the candidate. In some borderline cases, additional clarification was obtained through a brief online discussion with cameras turned off to better understand the candidate’s experience. Only candidates who satisfied the corresponding group-specific screening criteria were included in the final sample. Candidates who did not meet the requirements for the skilled group were either considered for the novice group (if consistent with the novice criteria) or not enrolled in the study. At present, T2I prompting lacks standardized skill metrics or objective proficiency tests. Skilled user identification thus relies on behavioral and experiential signals. Indeed, this lack of objective criteria directly motivates the development of \textbf{\texttt{AtelierEval}}, which aims to quantify prompting proficiency beyond coarse self-reported categories.

\subsection{Participant Demographic Statistics}
\label{apps:user_demo}

\begin{figure}[H]
    \centering
    \includegraphics[width=0.98\linewidth]{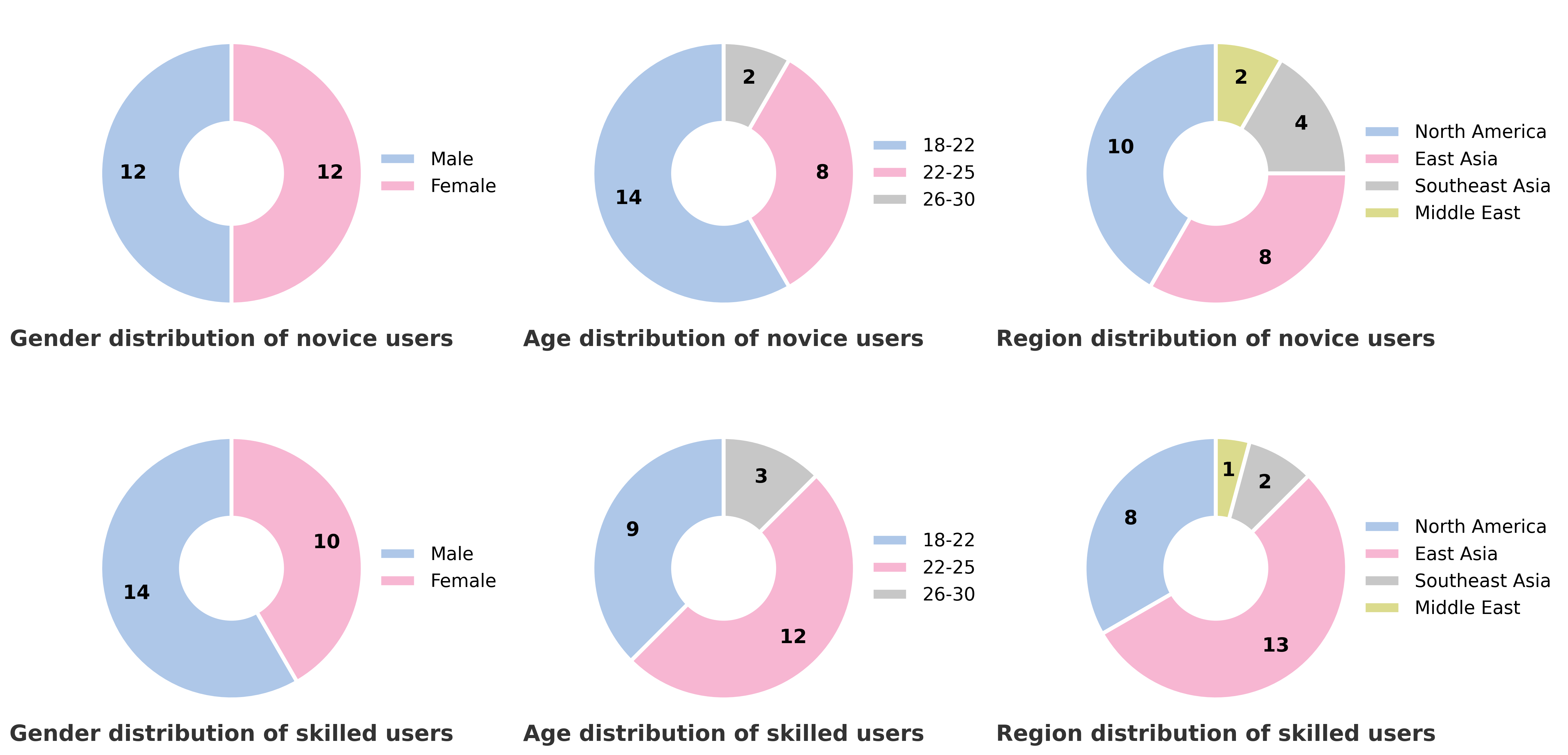}
  \caption{Gender, age range, and region of residence distributions are shown separately for the novice and skilled groups.}
    \label{fig:userdata}
\end{figure}

We report aggregated demographic statistics of the final participant pool in Figure \ref{fig:userdata}. Participants in our study are mainly young adults with regions of residence primarily concentrated in North America and East Asia. This demographic profile reflects a substantial subset of the current active user population of T2I systems \cite{gao2023exploring,demandsage2026_midjourney}. However, these attributes are reported to provide contextual information about the study population rather than to serve as explanatory variables for performance differences. Following prior work that cautions against collecting and interpreting race or ethnicity without a clearly articulated analytical purpose \cite{biega2020operationalizing,chen2023and}, we report region of residence as a less sensitive more appropriate descriptor of participant background given the research scope.

\section{Human Expert Involvement}
\label{app:experts}

We recruited a total of six human experts to facilitate the development and validation of our framework. Each expert possesses at least three years of research experience specifically within the T2I domain. These experts were organized into task-specific groups to handle dataset construction, exemplar memory construction, and evaluator verification. To prevent data contamination and ensure rigorous validation, the three experts assigned to exemplar memory construction were strictly distinct from the three experts responsible for evaluator verification. Dataset construction was performed by a subset of three experts selected from this total pool of six. Specific methodologies regarding dataset design and manual validation are detailed in Appendix~\ref{app:data_collect}. The construction of the exemplar memories used by AtelierJudge is described in Appendix~\ref{app:exemplar}, while the meta-evaluation protocol and human expert baseline are detailed in Section~\ref{meta_evaluation}.

\section{Subjective Evaluation Case Studies}
\label{app:evaluation_examples}

To demonstrate the rigor of our assessment logic and the necessity of decoupling subjective quality from objective constraints, we provide two qualitative case studies from the \textbf{\texttt{AtelierEval}} dataset below.

\subsection{Subjective Prompt Evaluation (Open-Ended Task)}
\label{app:Q1}

\textbf{Task (OE\_45):}
"I'm the art director for Metropolis Weekly. Our next cover story is a long-form investigative piece on The Air We Breathe. The deadline is tight. We need a cover image that shows the impact of air pollution on city life. The mood must be oppressive but also realistic. Our readers are educated urban-dwellers, so it needs to be subtle but powerful. It must not look like a futuristic, Blade Runner city. It needs to look like a real, modern city. And no cartoons."

\medskip

\textbf{Prompt (From Qwen-3-VL-235B-A22B, Expert):}
"A realistic, modern cityscape at dusk, with dense smog hanging low over the skyline, casting a hazy, gray pall over the streets. The buildings are contemporary but not futuristic—glass towers and concrete structures with subtle signs of wear. People move through the streets, some wearing masks, others coughing or looking away from the polluted air. The atmosphere feels heavy and oppressive, with muted colors and soft shadows emphasizing the weight of the pollution. The image is grounded in reality, avoiding any sci-fi or cartoonish elements, and conveys a quiet, powerful message about the unseen dangers of air quality in urban life."

\medskip

\textbf{AtelierJudge Evaluation:}

\begin{itemize}[leftmargin=*,
    topsep=-0.5em,
    partopsep=0pt,
    parsep=0pt,
    itemsep=0pt
    ]

\item[\ding{224}] \textbf{Instructional Clarity (Score: 5):} The prompt is syntactically flawless and logically structured. It explicitly addresses negative constraints ("not futuristic") and defines the subject clearly, leaving no ambiguity for the model.

\item[\ding{224}] \textbf{Creative Elaboration (Score: 3):} The imagery relies on standard, first-order associations with pollution (masks, coughing, gray smog). While descriptive, it lacks the unique, subtle narrative framing requested ("subtle but powerful").

\item[\ding{224}] \textbf{Terminology Proficiency (Score: 2):} The prompt relies entirely on layperson's adjectives ("heavy", "muted", "soft shadows") rather than professional photographic or rendering terminology (e.g., "volumetric fog", "desaturated grading", "telephoto compression").

\item[\ding{224}] \textbf{Intent Formalization (Score: 3):} A partial failure in translation. While some mood is visualized, the prompt resorts to pasting the abstract goal directly ("conveys a quiet, powerful message...") instead of operationalizing that message into observable visual cues.

\end{itemize}

\subsection{Subjective Image Evaluation (Constrained Task)}
\label{app:Q2}

This example illustrates a case of \textit{high-quality hallucination}, where the model ignores explicit constraints (Objective Failure) but produces an aesthetically coherent image. This highlights the necessity of decoupling subjective quality from objective constraint verification.

\textbf{Task (CO\_65):}
"A cafe needs a promotional banner for its website. The design should feel nostalgic and warm, for a general audience. The style is \textbf{retro 1950s advertisement}. The banner must not contain any modern devices... The coffee cup must be red... \textbf{ONLY red and cream colors}. Text 'Fresh Coffee' must be in the center."

\textbf{Prompt (From Gemini-3-Pro-Preview and Flux.1 Pro, Novice):}

\begin{figure}[H]
    \centering
    \includegraphics[width=0.4\linewidth]{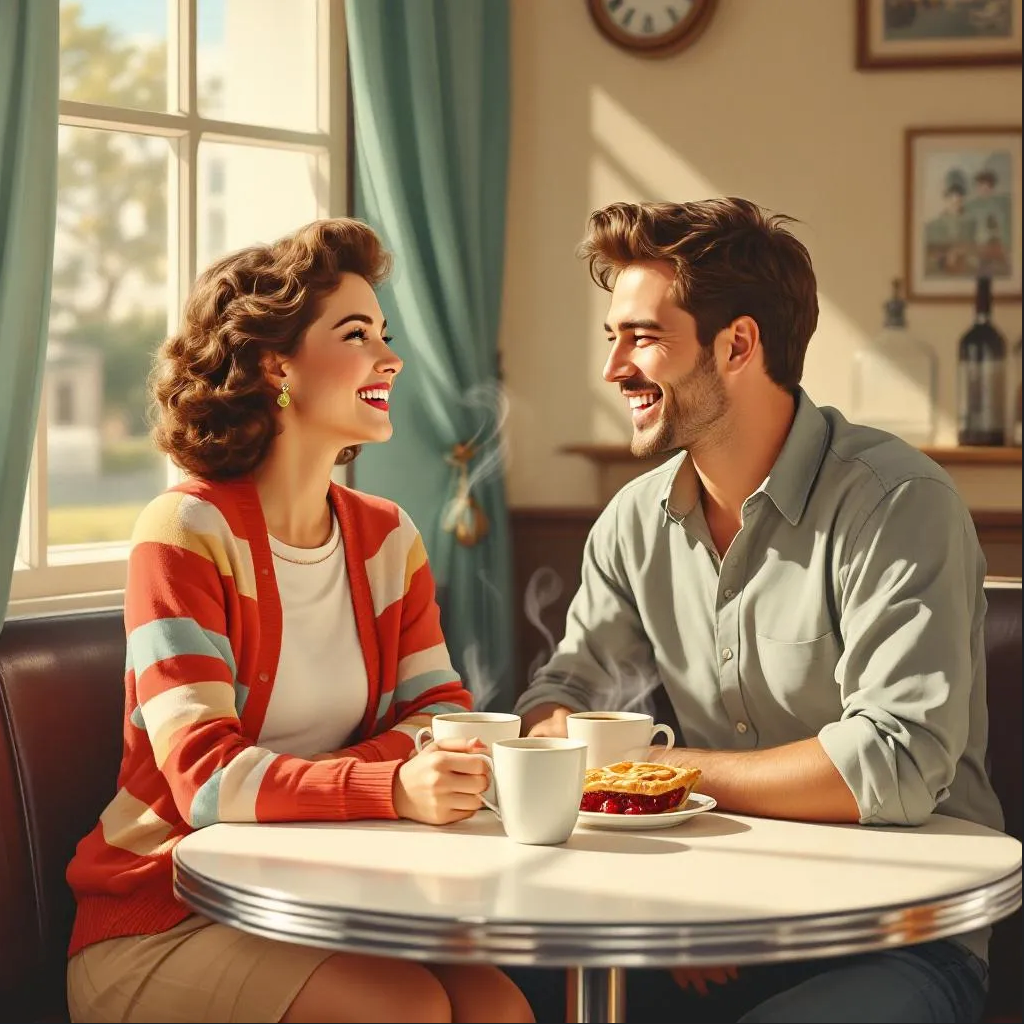}
    \caption{The generated image evaluated by AtelierJudge.}
    \label{fig:example_subj_image}
\end{figure}

\textbf{AtelierJudge Evaluation:}

\begin{itemize}[leftmargin=*,
    topsep=-0.5em,
    partopsep=0pt,
    parsep=0pt,
    itemsep=0pt
    ]

\item[\ding{224}] \textbf{Mood \& Atmosphere (Score: 3):} While the image captures the "warm and nostalgic" keyword, it completely misses the "Retro Advertisement" genre, presenting a cinematic still instead of a graphical banner. The mood is pleasant but fails to convey the intended commercial tone.

\item[\ding{224}] \textbf{Visual Composition (Score: 4):} \textit{(Note: Specific layout constraints are penalized in the Objective Check.)} As a standalone image, the composition is well-balanced and strictly framed. The visual weight is distributed effectively between the two subjects, creating a functional, albeit standard, photographic composition.

\item[\ding{224}] \textbf{Color \& Lighting (Score: 4):} \textit{(Note: Palette restrictions are penalized in the Objective Check.)} Subjectively, the warm lighting and retro color grading are harmonious and aesthetically pleasing, despite fulfilling the wrong color palette constraints (full color vs. red/cream).

\item[\ding{224}] \textbf{Technical Flawlessness (Score: 3):} The image exhibits distinct "AI artifacts": the skin textures appear overly smooth and plastic ("waxy"), and the steam from the cups has an unnatural, illustrative look that clashes with the realistic style. These recognizable generative traces prevent a higher score.

\end{itemize}
\section{Details of Validation Set}
\label{app:valid_set}

This appendix details the construction of the expert-annotated validation set used to meta-evaluate AtelierJudge. This set consists of 360 prompt--image pairs, one per task in \textbf{\texttt{AtelierEval}}, and is used to assess both the subjective scoring skills and the objective checklist-based skills of AtelierJudge.

\textbf{Sampling and composition.}
For each of the 360 tasks, we select exactly one prompt--image pair from the full pool of collected submissions. Each pair contains the original prompt provided to a T2I backend and the corresponding generated image, so that the two modalities always describe the same submission. Pairs are drawn via stratified random sampling over task type, so that the marginal distributions of tasks, systems, and difficulty closely match those of the overall benchmark. Each pair is associated with the task-specific checklist used for objective constraint verification.

\textbf{Annotation.}
For objective validation, one domain expert first labels whether the constraint is explicitly specified in the prompt and whether it is satisfied in the image. A second expert then independently reviews all labels and corrects any mistakes; residual disagreements are resolved through discussion, and the agreed labels are treated as the final objective annotations. For subjective validation, three independent experts rate each prompt and each image separately on all subjective dimensions. We take the arithmetic mean of the three scores as the subjective ground truth.

\section{Design Validation \& Ablation Study of AtelierJudge}
\label{app:ablation}

In this appendix, we validate the design choices of the subjective skills in AtelierJudge. We investigate the impact of retrieval strategies~\cite{li2025towards,wu2026gamhierarchicalgraphbasedagentic}, the number of retrieved exemplars ($K$), and the choice of embedding models. All experiments are conducted using GPT-5.2. We report the MAE, W1-A, and $\rho$ against human expert ratings on the validation set (Appendix \ref{app:valid_set}).

\begin{table}[H]
\centering
\caption{Ablation study on memory retrieval strategies.}
\label{tab:ablation_strategy}
\renewcommand{\arraystretch}{1.1}
\setlength{\tabcolsep}{8pt}
\resizebox{0.45\linewidth}{!}{
\begin{tabular}{l|ccc}
\toprule
\textbf{Strategy} & \textbf{MAE} $\downarrow$ & \textbf{W1-A} $\uparrow$ & \boldmath$\rho$ $\uparrow$ \\
\midrule
Zero-shot & 0.72 & 0.64 & 0.56 \\
Fixed Few-shot & 0.55 & 0.81 & 0.68 \\
Random Retrieval & 0.61 & 0.75 & 0.62 \\
\rowcolor{gray!10} \textbf{Similarity Retrieval (Ours)} & \textbf{0.34} & \textbf{0.93} & \textbf{0.79} \\
\bottomrule
\end{tabular}
}
\end{table}

\textbf{Impact of Retrieval Strategy.}
To verify the necessity of the similarity-based retrieval mechanism, we compare our method against three baselines, including the zero-shot baseline, a fixed few-shot baseline that utilizes a fixed set of 3 diverse exemplars (one per task category and covering 1-5 scores) for all queries, and a random baseline that randomly selects 3 exemplars from the corresponding memory for each query. As shown in Table \ref{tab:ablation_strategy}, while providing fixed or random examples improves performance over the zero-shot baseline by establishing a general scoring scale, AtelierJudge significantly outperforms all other strategies. This confirms that retrieving semantically relevant exemplars is crucial for aligning the evaluator with the specific nuances of the evaluated task.

\begin{table}[H]
\centering
\caption{Impact of the number of retrieved exemplars ($K$).}
\label{tab:ablation_k}
\renewcommand{\arraystretch}{1.1}
\setlength{\tabcolsep}{10pt}
\resizebox{0.45\linewidth}{!}{
\begin{tabular}{l|ccc}
\toprule
\textbf{Setting} & \textbf{MAE} $\downarrow$ & \textbf{W1-A} $\uparrow$ & \boldmath$\rho$ $\uparrow$ \\
\midrule
Baseline (No Memory) & 0.72 & 0.64 & 0.56 \\
\midrule
$K=1$ & 0.56 & 0.83 & 0.63 \\
$K=2$ & 0.39 & 0.89 & 0.76 \\
\rowcolor{gray!10} \textbf{$K=3$ (Ours)} & \textbf{0.34} & \textbf{0.93} & \textbf{0.79} \\
$K=4$ & 0.35 & 0.91 & 0.78 \\
\bottomrule
\end{tabular}
}
\end{table}

\textbf{Sensitivity to Exemplar Count ($K$).}
We investigate the optimal number of retrieved exemplars $K$ by varying it from 1 to 4. As presented in Table \ref{tab:ablation_k}, the performance of AtelierJudge achieves peak when $K=3$, as multiple exemplars provide a more robust triangulation of the scoring criteria. However, increasing $K$ to 4 yields diminished performance. We attribute it to the context window with less relevant information, especially when processing images. Consequently, we adopt $K=3$ as the optimal value.

\begin{table}[H]
\centering
\caption{Comparison of different embedding models for retrieval.}
\label{tab:ablation_models}
\renewcommand{\arraystretch}{1.1}
\setlength{\tabcolsep}{6pt}
\resizebox{0.6\columnwidth}{!}{
\begin{tabular}{ll|ccc}
\toprule
\textbf{Text Encoder} & \textbf{Image Encoder} & \textbf{MAE} $\downarrow$ & \textbf{W1-A} $\uparrow$ & \boldmath$\rho$ $\uparrow$ \\
\midrule
\multicolumn{2}{l|}{\textit{Baseline (No Retrieval)}} & 0.72 & 0.64 & 0.56 \\
\midrule
NV-Embed-v2 (7B) & DINO-V2-Giant & 0.41 & 0.88 & 0.74 \\
Nomic-Embed-Text & DINO-V3 & 0.35 & 0.92 & 0.78 \\
\rowcolor{gray!10} \textbf{Nomic-Embed-Text (Ours)} & \textbf{DINO-V2-Giant} & \textbf{0.34} & \textbf{0.93} & \textbf{0.79} \\
\bottomrule
\end{tabular}
}
\end{table}

\textbf{Impact of Embedding Models.}
We justify our choice of embedding models by comparing them against large-scale models. Following the selection strategies of recent studies \cite{luo2025agentauditor,jose2025dinov2}, we select our embedding models as Nomic-Embed-Text-V1.5 (dim=512) for text and DINO-V2-Giant for images. We replace the text encoder with NV-Embed-v2 \cite{lee2024nv} and the image encoder with DINO-V3 \cite{simeoni2025dinov3}, two widely-recognized 7B models for comparison. The results in Table \ref{tab:ablation_models} reveal two insights. First, scaling up the image encoder to DINO-V3 brings no observable improvement, suggesting that DINO-V2 already captures sufficient perceptual features for this task. Second, surprisingly, NV-Embed-v2 leads to a performance degradation. We hypothesize that while NV-Embed-v2 excels in general benchmarks, its high-dimensional embedding space (dim=4096) may over-emphasize semantic minutiae rather than the instructional alignment features required for our specific retrieval task. These findings validate our choice.

\section{Detailed Benchmarking Results}
\label{app:more_result}

\begin{table}[H]
\centering
\caption{Detailed experimental results for novice and skilled prompters across different T2I backends on CO tasks. Subjective (Subj.) scores are reported on a 5-point scale, and Objective (Obj.) scores are reported as percentages}
\label{tab:co}
\setlength{\tabcolsep}{8pt}
\renewcommand{\arraystretch}{1.1}
\resizebox{0.9\textwidth}{!}{%
\begin{tabular}{ll|cc|cc|cc|cc|cc|cc}
\toprule
\multicolumn{2}{c|}{\multirow{2}{*}{Model}} & \multicolumn{2}{c|}{Prompt} & \multicolumn{2}{c|}{Image Avg.} & \multicolumn{2}{c|}{nBanana} & \multicolumn{2}{c|}{GI-1} & \multicolumn{2}{c|}{Flux Pro} & \multicolumn{2}{c}{SDXL} \\
\cmidrule{3-14}
 & & Subj. & Obj. & Subj. & Obj. & Subj. & Obj. & Subj. & Obj. & Subj. & Obj. & Subj. & Obj. \\
\midrule
\multirow{2}{*}{GPT-5.2} & no. & 2.99 & 37.5 & 3.65 & 41.8 & 3.83 & 47.3 & 3.88 & 47.2 & 3.60 & 38.6 & 3.27 & 34.2 \\
 & sk. & 2.93 & 35.6 & 3.60 & 41.9 & 3.86 & 46.8 & 3.90 & 48.1 & 3.48 & 38.5 & 3.16 & 34.1 \\
\hline
\multirow{2}{*}{Gem-3} & no. & 2.81 & 36.2 & 3.68 & 41.6 & 3.76 & 45.5 & 3.88 & 45.5 & 3.63 & 39.4 & 3.46 & 35.8 \\
 & sk. & 2.81 & 37.4 & 3.65 & 41.5 & 3.81 & 46.0 & 3.91 & 46.3 & 3.61 & 39.2 & 3.29 & 34.4 \\
\hline
\multirow{2}{*}{Cl-4.5} & no. & 2.57 & 27.8 & 3.50 & 40.7 & 3.65 & 45.0 & 3.77 & 48.6 & 3.43 & 34.4 & 3.17 & 34.9 \\
 & sk. & 2.69 & 33.2 & 3.54 & 41.4 & 3.73 & 45.7 & 3.83 & 46.7 & 3.43 & 38.1 & 3.19 & 34.9 \\
\hline
\multirow{2}{*}{GPT-4.1} & no. & 2.66 & 32.9 & 3.57 & 42.4 & 3.67 & 47.9 & 3.86 & 46.2 & 3.51 & 40.4 & 3.26 & 35.0 \\
 & sk. & 2.74 & 36.1 & 3.65 & 42.0 & 3.86 & 47.5 & 3.94 & 46.5 & 3.56 & 41.0 & 3.23 & 33.1 \\
\hline
\multirow{2}{*}{Gem-2} & no. & 2.27 & 22.8 & 3.41 & 38.8 & 3.54 & 44.9 & 3.77 & 45.7 & 3.24 & 32.7 & 3.10 & 31.9 \\
 & sk. & 2.32 & 27.1 & 3.45 & 39.4 & 3.71 & 46.1 & 3.77 & 45.8 & 3.28 & 34.5 & 3.04 & 31.3 \\
\hline
\multirow{2}{*}{Qwen-L} & no. & 2.73 & 31.9 & 3.64 & 41.3 & 3.79 & 43.9 & 3.88 & 45.5 & 3.54 & 40.1 & 3.36 & 35.7 \\
 & sk. & 2.82 & 35.3 & 3.58 & 40.0 & 3.75 & 44.0 & 3.79 & 44.6 & 3.46 & 38.2 & 3.30 & 33.4 \\
\hline
\multirow{2}{*}{GPT-4n} & no. & 2.50 & 28.4 & 3.49 & 42.1 & 3.73 & 45.7 & 3.82 & 47.4 & 3.32 & 38.3 & 3.07 & 37.0 \\
 & sk. & 2.61 & 32.4 & 3.52 & 43.2 & 3.75 & 46.6 & 3.73 & 48.7 & 3.42 & 38.4 & 3.19 & 39.3 \\
\hline
\multirow{2}{*}{Qwen-S} & no. & 2.49 & 26.5 & 3.50 & 40.1 & 3.71 & 46.5 & 3.78 & 46.7 & 3.36 & 34.8 & 3.16 & 32.3 \\
 & sk. & 2.47 & 26.9 & 3.43 & 39.2 & 3.59 & 46.1 & 3.71 & 45.8 & 3.32 & 34.6 & 3.10 & 30.5 \\
\hline
\multirow{2}{*}{Human} & no. & 3.10 & 48.2 & 2.77 & 51.4 & 2.98 & 55.5 & 2.54 & 62.8 & 2.98 & 50.1 & 2.57 & 37.2 \\
 & sk. & 4.05 & 78.1 & 3.80 & 75.7 & 3.92 & 82.8 & 3.87 & 81.5 & 3.82 & 72.4 & 3.61 & 66.0 \\
\bottomrule
\end{tabular}%
}
\end{table}

\begin{table}[H]
\centering
\caption{Detailed experimental results for novice and skilled prompters across different T2I backends on IM tasks. Subjective (Subj.) scores are reported on a 5-point scale, and Objective (Obj.) scores are reported as percentages}
\label{tab:im}
\setlength{\tabcolsep}{8pt}
\renewcommand{\arraystretch}{1.1}
\resizebox{0.9\textwidth}{!}{%
\begin{tabular}{ll|cc|cc|cc|cc|cc|cc}
\toprule
\multicolumn{2}{c|}{\multirow{2}{*}{Model}} & \multicolumn{2}{c|}{Prompt} & \multicolumn{2}{c|}{Image Avg.} & \multicolumn{2}{c|}{nBanana} & \multicolumn{2}{c|}{GI-1} & \multicolumn{2}{c|}{Flux Pro} & \multicolumn{2}{c}{SDXL} \\
\cmidrule{3-14}
 & & Subj. & Obj. & Subj. & Obj. & Subj. & Obj. & Subj. & Obj. & Subj. & Obj. & Subj. & Obj. \\
\midrule
\multirow{2}{*}{GPT-5.2} & no. & 4.46 & 56.7 & / & 60.9 & / & 74.8 & / & 71.6 & / & 59.8 & / & 37.5 \\
 & sk. & 4.66 & 64.6 & / & 62.9 & / & 79.2 & / & 76.5 & / & 57.9 & / & 37.9 \\
\hline
\multirow{2}{*}{Gem-3} & no. & 4.25 & 53.5 & / & 60.4 & / & 72.4 & / & 70.5 & / & 60.5 & / & 38.4 \\
 & sk. & 4.62 & 66.6 & / & 63.1 & / & 81.9 & / & 77.5 & / & 57.3 & / & 35.7 \\
\hline
\multirow{2}{*}{Cl-4.5} & no. & 3.87 & 41.2 & / & 57.1 & / & 68.4 & / & 65.6 & / & 54.8 & / & 39.5 \\
 & sk. & 4.32 & 59.0 & / & 61.7 & / & 77.3 & / & 74.3 & / & 57.1 & / & 38.3 \\
\hline
\multirow{2}{*}{GPT-4.1} & no. & 3.63 & 32.9 & / & 55.4 & / & 62.1 & / & 62.0 & / & 55.7 & / & 41.7 \\
 & sk. & 4.07 & 45.1 & / & 58.4 & / & 69.4 & / & 68.1 & / & 57.5 & / & 38.7 \\
\hline
\multirow{2}{*}{Gem-2} & no. & 3.23 & 24.8 & / & 52.6 & / & 60.1 & / & 58.9 & / & 50.0 & / & 41.2 \\
 & sk. & 3.50 & 34.4 & / & 57.4 & / & 67.0 & / & 66.8 & / & 55.2 & / & 40.4 \\
\hline
\multirow{2}{*}{Qwen-L} & no. & 3.79 & 38.5 & / & 48.9 & / & 60.5 & / & 58.4 & / & 46.0 & / & 30.7 \\
 & sk. & 4.39 & 52.8 & / & 57.9 & / & 74.1 & / & 70.9 & / & 53.5 & / & 32.9 \\
\hline
\multirow{2}{*}{GPT-4n} & no. & 3.55 & 30.6 & / & 49.9 & / & 55.6 & / & 56.4 & / & 49.7 & / & 37.7 \\
 & sk. & 3.84 & 43.4 & / & 54.5 & / & 64.7 & / & 62.3 & / & 53.0 & / & 38.0 \\
\hline
\multirow{2}{*}{Qwen-S} & no. & 3.26 & 42.6 & / & 41.7 & / & 53.7 & / & 54.0 & / & 37.2 & / & 22.7 \\
 & sk. & 3.64 & 46.3 & / & 40.9 & / & 56.4 & / & 50.8 & / & 36.3 & / & 20.5 \\
\hline
\multirow{2}{*}{Human} & no. & 3.02 & 40.4 & / & 40.0 & / & 37.2 & / & 43.8 & / & 40.9 & / & 38.0 \\
 & sk. & 3.94 & 71.3 & / & 58.9 & / & 73.0 & / & 70.4 & / & 53.3 & / & 38.9 \\
\bottomrule
\end{tabular}%
}
\end{table}

\begin{table}[H]
\centering
\caption{Detailed experimental results for novice and skilled prompters across different T2I backends on OE tasks. Subjective (Subj.) scores are reported on a 5-point scale, and Objective (Obj.) scores are reported as percentages}
\label{tab:oe}
\setlength{\tabcolsep}{8pt}
\renewcommand{\arraystretch}{1.1}
\resizebox{0.9\textwidth}{!}{%
\begin{tabular}{ll|cc|cc|cc|cc|cc|cc}
\toprule
\multicolumn{2}{c|}{\multirow{2}{*}{Model}} & \multicolumn{2}{c|}{Prompt} & \multicolumn{2}{c|}{Image Avg.} & \multicolumn{2}{c|}{nBanana} & \multicolumn{2}{c|}{GI-1} & \multicolumn{2}{c|}{Flux Pro} & \multicolumn{2}{c}{SDXL} \\
\cmidrule{3-14}
 & & Subj. & Obj. & Subj. & Obj. & Subj. & Obj. & Subj. & Obj. & Subj. & Obj. & Subj. & Obj. \\
\midrule
\multirow{2}{*}{GPT-5.2} & no. & 4.70 & 95.8 & 4.27 & 94.3 & 4.45 & 97.5 & 4.53 & 96.9 & 4.25 & 94.0 & 3.85 & 89.0 \\
 & sk. & 4.72 & 90.8 & 4.32 & 91.4 & 4.48 & 94.4 & 4.55 & 95.5 & 4.32 & 90.1 & 3.93 & 85.7 \\
\hline
\multirow{2}{*}{Gem-3} & no. & 4.37 & 87.8 & 4.34 & 92.0 & 4.44 & 93.8 & 4.56 & 93.4 & 4.33 & 91.4 & 4.03 & 89.4 \\
 & sk. & 4.60 & 92.5 & 4.35 & 91.7 & 4.49 & 93.7 & 4.57 & 93.8 & 4.39 & 93.2 & 3.94 & 86.2 \\
\hline
\multirow{2}{*}{Cl-4.5} & no. & 3.90 & 94.4 & 4.26 & 94.4 & 4.39 & 96.7 & 4.50 & 96.8 & 4.22 & 94.2 & 3.95 & 90.0 \\
 & sk. & 4.30 & 88.2 & 4.30 & 92.2 & 4.44 & 94.6 & 4.51 & 94.9 & 4.25 & 91.0 & 4.01 & 88.2 \\
\hline
\multirow{2}{*}{GPT-4.1} & no. & 4.10 & 86.3 & 4.28 & 91.3 & 4.47 & 92.7 & 4.51 & 94.7 & 4.26 & 90.8 & 3.89 & 87.2 \\
 & sk. & 4.34 & 87.0 & 4.33 & 91.4 & 4.47 & 93.9 & 4.55 & 93.7 & 4.31 & 92.4 & 4.00 & 85.6 \\
\hline
\multirow{2}{*}{Gem-2} & no. & 3.38 & 82.8 & 4.23 & 91.4 & 4.36 & 92.9 & 4.46 & 94.1 & 4.11 & 90.5 & 4.00 & 88.4 \\
 & sk. & 3.88 & 83.1 & 4.32 & 91.8 & 4.41 & 93.1 & 4.49 & 93.8 & 4.29 & 92.5 & 4.09 & 87.8 \\
\hline
\multirow{2}{*}{Qwen-L} & no. & 4.43 & 92.4 & 4.33 & 93.6 & 4.47 & 96.8 & 4.51 & 96.2 & 4.29 & 92.5 & 4.07 & 88.9 \\
 & sk. & 4.56 & 88.4 & 4.39 & 91.3 & 4.49 & 94.6 & 4.59 & 94.0 & 4.41 & 89.9 & 4.07 & 86.5 \\
\hline
\multirow{2}{*}{GPT-4n} & no. & 3.76 & 84.3 & 4.29 & 90.5 & 4.44 & 92.7 & 4.52 & 93.3 & 4.23 & 89.8 & 3.99 & 86.1 \\
 & sk. & 4.00 & 83.8 & 4.31 & 90.7 & 4.42 & 91.7 & 4.51 & 93.7 & 4.28 & 90.7 & 4.02 & 86.5 \\
\hline
\multirow{2}{*}{Qwen-S} & no. & 3.71 & 82.8 & 4.30 & 91.7 & 4.40 & 92.2 & 4.50 & 94.1 & 4.26 & 91.2 & 4.04 & 89.2 \\
 & sk. & 4.02 & 84.9 & 4.31 & 91.5 & 4.43 & 93.6 & 4.49 & 96.1 & 4.32 & 91.2 & 4.03 & 85.3 \\
\hline
\multirow{2}{*}{Human} & no. & 2.58 & 80.9 & 3.25 & 79.4 & 3.24 & 81.9 & 3.95 & 92.6 & 2.90 & 79.4 & 2.91 & 63.7 \\
 & sk. & 3.65 & 92.3 & 4.14 & 95.5 & 4.30 & 98.9 & 4.15 & 98.6 & 4.05 & 93.6 & 4.07 & 91.2 \\
\bottomrule
\end{tabular}%
}
\end{table}

\section{Stability Analysis of Evaluation Scale}
\label{app:scale}

In the main experiments summarized in Table~\ref{tab:main_results}, each prompter--task pair is evaluated under a fixed sampling scheme: for every task, the prompter produces a single natural-language prompt, each prompt is executed by every T2I backend to generate four images, and AtelierJudge retains the highest-scored image (\emph{top-1}) to compute all metrics. This setting involves two potential sources of sampling variability: (i) the MLLM prompter when generating prompts and (ii) the T2I backend when generating images. Although all MLLMs are decoded with temperature $0.0$ and shared hyperparameters (Appendix M), making prompt sampling nearly deterministic, it might be questioned whether using only one prompt and four images per task is sufficient for stable evaluation. This appendix therefore studies the sampling stability of our evaluation scale and examines how scores change when increasing the number of prompts and images.

\paragraph{Experimental setup.}
We conduct a controlled stability study under a single representative configuration. The prompter is GPT-5.2 and the T2I backend is GI-1, a strong pair that is also used extensively in the main results. We evaluate on a stratified subset of 120 tasks from \textbf{\texttt{AtelierEval}}, balanced across the three task categories (OE, CO, IM) introduced in Section~\ref{sec:categories}. Decoding and evaluator hyperparameters follow Appendix~\ref{app:para}: GPT-5.2 runs with temperature $0.0$, and all prompts and images are scored by AtelierJudge as described in Section 4 using the subjective dimensions and memories from Appendix H and Appendix~\ref{app:judge_prompts}. For all conditions we reuse the full evaluation pipeline of Section~\ref{sec:protocol} and Section~\ref{sec:exp_set} and vary only the numbers of prompts and images.

For the \emph{prompt-scale} analysis we vary the number of independently sampled prompts per task,$N_{\text{prompt}} \in \{1, 2, 3, 4\}$, while keeping the image sampling scheme fixed to four images per prompt. For each $N_{\text{prompt}}$ we run the complete pipeline and report two aggregated metrics over tasks: the mean subjective prompt score (1--5) and the mean prompt-side checklist satisfaction rate in percent, referred to as \emph{prompt accuracy}. Even though decoding uses temperature $0.0$, we still treat repeated generations as independent runs and check whether these aggregate metrics drift as $N_{\text{prompt}}$ increases.

For the \emph{image-scale} analysis we fix a single prompt per task and vary the number of images sampled from GI-1, $N_{\text{img}} \in \{1, 2, 4, 8, 16\}$. For each task and each $N_{\text{img}}$, GI-1 generates $N_{\text{img}}$ images under the same configuration as in the main experiments. AtelierJudge scores all images independently, and we retain the highest-scored image (\emph{top-1}) when aggregating metrics, exactly as in the main experiments. For each $N_{\text{img}}$ we again report two metrics averaged over tasks: the mean subjective score of the top-1 image and the mean image-side checklist satisfaction rate in percent (\emph{top-1 image accuracy}). In both analyses, checklist satisfaction is computed following the objective evaluation protocol of Section~\ref{sec: S2}.

\begin{figure}[H]
  \centering
  \includegraphics[width=0.98\linewidth]{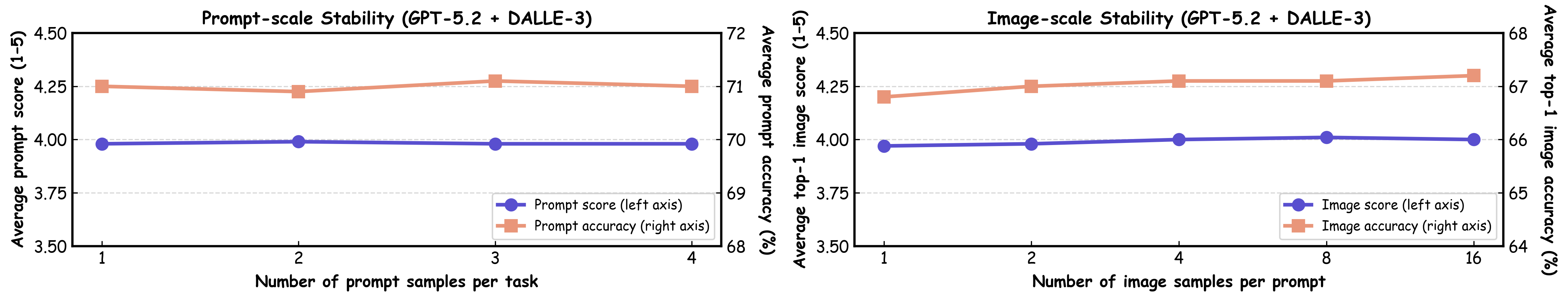}
  \caption{Prompt-scale and image-scale stability for GPT-5.2 as prompter and GI-1 as the T2I backend. Left: effect of the number of prompt samples per task on mean prompt score (left axis) and prompt accuracy (right axis). Right: effect of the number of image samples per prompt on mean top-1 image score (left axis) and top-1 image accuracy (right axis). Accuracy is the checklist-based constraint satisfaction rate defined in Section~\ref{sec: S2}.}
  \label{fig:app-scale}
\end{figure}

\paragraph{Results and discussion.}
Figure~\ref{fig:app-scale} shows the stability results. Across all values of $N_{\text{prompt}}$ and $N_{\text{img}}$, the curves for subjective scores and objective checklist-based accuracies remain nearly flat, and the small fluctuations are within random variation, indicating that neither metric is materially affected by the number of sampled prompts or images. Taken together, these observations justify the default design used throughout the main experiments: evaluating each prompter--task pair with one prompt and four images per T2I backend, retaining the top-1 image for scoring. This configuration lies on the plateau of both stability curves and therefore provides a computationally efficient yet converged operating point. The small variations observed across different $N_{\text{prompt}}$ and $N_{\text{img}}$ further indicate that our evaluation scale is robust to these sampling hyperparameters for both subjective scores and objective accuracy, and that the main experimental conclusions are stable.

\section{User Study Materials}
\label{apx:human_study_materials}
This section presents all materials provided to the user study participants, including the informed consent form and the pre-test screening questionnaire. To preserve anonymity during the review process, certain experiment-irrelevant details have been omitted. Detailed ethical considerations about our human study are provided in Appendix \ref{app:ethics}.

\subsection{Informed Consent Form}
\label{app:consent}
\textit{The following is a static version of the informed consent form for inclusion. In the actual study, participants were required to check a box and sign, indicating they had read, understood, and voluntarily agreed to participate.}

\textbf{Research Project Title:} AtelierEval: Agentic Evaluation of Text-to-Image Prompters \\
\textbf{Principal Investigator:} xxxxxxx \\
\textbf{Institution:} xxxxxxxxxxxxxx \\
\textbf{Contact Information:} xxxxxxxxxxxxxx@xxx.xxx

\subsubsection*{1. Introduction}
You are invited to participate in an academic study aimed at establishing a standardized benchmark for evaluating \textbf{Prompting Proficiency} in the era of Generative AI. Your participation will provide critical data to understand how humans translate visual intent into textual instructions compared to AI. This study is conducted entirely in English and requires proficiency in reading complex instructions and writing descriptive text. To ensure the ethical conduct of this research and the protection of all participants, this study has been reviewed and approved by the Institutional Review Board (IRB). If you have any questions, please contact the Principal Investigator by email.

\subsubsection*{2. Procedures}
If you agree to participate in this study, you will be asked to complete the following steps:

\begin{itemize}[leftmargin=*]
    \item \textbf{Pre-Test Screening:} You will first complete a brief questionnaire (approx. 5--10 minutes) regarding your experience with Text-to-Image tools and background. This ensures you meet the study's criteria and allows us to categorize participants into ``Novice'' or ``Skilled'' groups.

    \item \textbf{Mandatory Technical Setup:} 
    \begin{itemize}
        \item \textbf{Screen Recording:} To ensure integrity and verify that no unauthorized AI tools are used, you are \textbf{required} to record your entire screen during the session. Before starting the recording, please close all personal messaging apps and irrelevant browser tabs to protect your privacy. You must record the full screen. To ensure manageable file sizes and clarity, please restrict the recording resolution from 1920$\times$1200 to 1280$\times$720. Please note that failure to provide a valid screen recording, evidence of prohibited tool usage, or obvious lack of effort (e.g., irrelevant inputs) will result in disqualification and forfeiture of all compensation.
        \item \textbf{Network Access:} You are kindly required to access HuggingFace and DeepL in the test. If you experience difficulties accessing them due to network restrictions, we will provide a secure VPN service for the duration of the study.
    \end{itemize}

    \item \textbf{Prompting Tasks:} If selected, you will access a web-based interface (Gradio-based, similar to Stable Diffusion WebUI) to write text prompts to generate images based on specific requirements. You will complete a total of 10 prompting tasks for each of the below three task types:
    \begin{itemize}
        \item \textbf{Open-ended Creation:} Writing prompts based on abstract requests to test creativity.
        \item \textbf{Constrained Creation:} Writing prompts based on requests that strictly adhere to logical or technical constraints.
        \item \textbf{Imitation:} Writing prompts to reverse-engineer and reproduce a reference image.
    To encourage careful thinking and prevent low-quality submissions, a mandatory minimum timer for 1 min is required for each task. The submission button will remain disabled until the timer expires. You are required to complete 5 tasks each  round, 6 rounds in total. Between rounds, you may take a short break to rest (typically up to 10 minutes). Please note that if there is no activity for more than 10 minutes, the session may be automatically ended, and you may need to restart the study.
    \end{itemize}

    \item \textbf{Tool Usage Policy:}
    \begin{itemize}
        \item \textbf{Allowed:} You are permitted to use standard mode of DeepL Translator strictly for language translation.
        \item \textbf{Prohibited:} You may NOT use AI chatbots (e.g., ChatGPT, Claude, Gemini). You may NOT use any other functions of DeepL, such as DeepL Write for AI Polish/Rewrite. The content of the prompt must originate from you.
    \end{itemize}

    \item \textbf{Time Commitment:}
    \begin{itemize}
        \item \textbf{Novice Group:} Estimated total time is approximately 45 to 60 minutes.
        \item \textbf{Skilled Group:} Estimated total time is approximately 1.5 to 3 hours, reflecting the expectation of high-fidelity, professional-grade inputs.
    \end{itemize}

    \item \textbf{Data Collection:} The system will record your submitted text prompts, timestamps, generated images, and operational logs. Upon completion, you must manually upload the screen recording file to the provided secure Google Drive or Baidu Netdisk link.
    
\end{itemize}

\subsubsection*{3. Risks}
\begin{itemize}[leftmargin=*]
    \item \textbf{Risks:} As approved by the IRB, the risks associated with this study are minimal. You may experience some stress or frustration if the generated images do not meet your expectations, which is a common occurrence in generative AI. All provided images and task descriptions are screened to eliminate any harmful contents. You may contact us if you find anything that makes you uncomfortable. All your personal identity data will be strictly anonymized.
\end{itemize}

\subsubsection*{4. Compensation}
Upon completion of all 30 tasks and the questionnaires, novice participants will receive a compensation of 12 USD (or the equivalent in local currency), while skilled participants will receive 50 USD. Payments can be processed via Amazon Gift Card, Zelle, Alipay, or WeChat Pay, with the specific method to be coordinated with each participant following the study's conclusion.

\subsubsection*{5. Confidentiality}
We will take strict measures to protect your privacy.
\begin{itemize}[leftmargin=*]
    \item \textbf{Anonymous Access:} To ensure your anonymity, you will not use your personal HuggingFace account. You will be provided with a uniformly assigned, anonymous HuggingFace account to access HuggingFace Space for the Gradio-based user interface for the tasks. The credentials for this account will be sent to your registered email address.
    \item \textbf{Data Usage:} The personal information you provide will be used for specific, distinct purposes. Your contact information, typically email, will be used strictly for study-related communication and compensation. Your demographic and background information will be used for anonymized statistical analysis in our research. Your screen recording is strictly for verification. It will be accessible only to the research team and will be permanently deleted after the verification process is complete.
    \item \textbf{Publication:} All published research findings will use fully anonymized, aggregated data. No information that could personally identify you will be disclosed.
\end{itemize}

\subsubsection*{6. Voluntary Participation}
Your participation is voluntary. You may withdraw at any time without penalty.

\subsection{Pre-Test Questionnaire}
\label{apx:pre-test}
\textit{The following questionnaire was administered to screen and assign participants.}
This questionnaire is designed to understand your background to ensure you meet the participation criteria for this study and to assign you to the most suitable task group. The information you provide will be kept strictly confidential. Please ensure that all information you provide is truthful. We reserve the right to withhold compensation if any of the information is found to be false. We will try our best to assign you to the group you applied for, but we do not guarantee it.

\subsubsection*{Part 1: Basic Information}
\begin{enumerate}
\item \textbf{Name:}
\item \textbf{Email Address:}
\item \textbf{Which group are you applying for? (Select one)}
\begin{itemize}
\item Skilled
\item Novice
\end{itemize}
\end{enumerate}

\subsubsection*{Part 2: Demographic \& Educational Background}
\begin{enumerate} \setcounter{enumi}{3}
    \item \textbf{Age range:}
    \begin{itemize}
        \item 18--22
        \item 22-25
        \item 26--30
        \item 30--40
        \item 40+
        \item Prefer not to say
    \end{itemize}

    \item \textbf{Gender:}
    \begin{itemize}
        \item Male
        \item Female
        \item Non-binary
        \item Prefer not to say
    \end{itemize}

    \item \textbf{Current primary region of residence or work (for the past 3+ years):}
    \begin{itemize}
        \item East Asia (e.g., Mainland China, Hong Kong SAR, Taiwan, Japan, Korea)
        \item Southeast Asia (e.g., Singapore, Malaysia, Thailand)
        \item South Asia
        \item Middle East
        \item North America
        \item Europe
        \item Latin America
        \item Sub-Saharan Africa
        \item Oceania
        \item Other
        \item Prefer not to say
    \end{itemize}

    \item \textbf{What is your current or highest level of education?}
    \begin{itemize}
        \item Secondary Education or Technical College
        \item Undergraduate
        \item Master
        \item PhD
        \item Prefer not to say
    \end{itemize}

    \item \textbf{Major(s)} (Use ``None'' if Secondary Education):
\end{enumerate}

\subsubsection*{Part 3: English Proficiency}
\begin{enumerate} \setcounter{enumi}{8}
\item \textbf{Is English your native language?} (Yes / No)
\item \textbf{(If No) Standardized English test scores (if applicable):}
\begin{itemize}
\item TOEFL
\item IELTS
\item Duolingo
\item CET-4
\item CET-6
\item Other
\item I have not taken any
\end{itemize}
\end{enumerate}

\subsubsection*{Part 4: Text-to-Image Experience \& Proficiency}
\textit{This section determines your group. Please answer truthfully.}
\begin{enumerate} \setcounter{enumi}{10}
\item \textbf{How long have you been using T2I tools?}
\begin{itemize}
\item I have never used them.
\item Less than 1 month.
\item 1--6 months.
\item 6 months -- 1 year.
\item More than 1 year.
\end{itemize}
\item \textbf{Which types of tools do you use regularly? (Select all that apply)}
\begin{itemize}
\item \textbf{Conversational Interfaces:} Interfaces where you describe the image in natural language, and an AI assistant rewrites/handles the prompt for you (e.g., the image generation function of ChatGPT, Bing Image Creator, Doubao, Gemini).
\item \textbf{Professional/Direct Web Tools:} Web interfaces where your exact text is sent to the model without hidden rewriting (e.g., Google AI Studio, Midjourney, Ideogram).
\item \textbf{Local Interfaces:} Advanced local interfaces allowing model switching, ControlNet, or node-based editing (e.g., Stable Diffusion WebUI, ComfyUI).
\item \textbf{Design Software Integration:} Image generation tools integrated in professional design software (e.g., Photoshop Generative Fill).
\end{itemize}
\item \textbf{Knowledge Check: Which of the following concepts can you confidently explain or use? (Check all that apply)}
\textit{This helps us gauge your technical and artistic depth.}
\begin{itemize}
\item \textbf{Technical:} Seed / Randomness
\item \textbf{Technical:} CFG Scale (Classifier-Free Guidance)
\item \textbf{Technical:} Checkpoints / Base Models / LoRA / Embeddings / Textual Inversion
\item \textbf{Artistic:} Composition Rules (e.g., Rule of Thirds, Golden Ratio)
\item \textbf{Artistic:} Lighting Styles (e.g., Volumetric, Chiaroscuro, Rim Light)
\item \textbf{Artistic:} Camera Angles/Lens (e.g., Isometric, Macro, Wide-angle)
\item \textbf{None of the above}
\end{itemize}
\item \textbf{Self-Assessment: How do you typically construct a prompt?}
\begin{itemize}
\item \textbf{Natural Language:} ``A cat sitting on a bench, sunny day.''
\item \textbf{Keyword Stacking:} ``Cat, bench, park, sun, 4k, high quality, masterpiece.''
\item \textbf{Structured Engineering:} I systematically organize subject, medium, style, artist references, and technical parameters.
\end{itemize}
\end{enumerate}

\subsubsection*{Part 5: Portfolio Verification (Required for Skilled Users)}
\textit{To verify your expertise level, please share 1--3 recent examples of your work. Both the \textbf{prompt text} and the \textbf{generated image} are required. Alternatively, if you have a portfolio on a social media or AI art platform (e.g., Civitai, ArtStation), please provide the link.}
\begin{enumerate} \setcounter{enumi}{14}
\item \textbf{Portfolio Link or Upload Description:}
\end{enumerate}

\section{Participant Workflow and Interface Design}
\label{app:assessment_platform}

This section provides a comprehensive walkthrough of the \textbf{\texttt{AtelierEval}} assessment platform, illustrating both the participant experience and the underlying workflow design~\cite{wang2024interactive,qin2025interpretable}. We present the complete procedural flow that participants encounter during evaluation, showcasing the user interface design and highlighting the framework's ecological validity. This walkthrough serves dual purposes: (1) demonstrating the practical implementation of our benchmark, and (2) providing transparency for reproducibility. The entire protocol is designed as a self-contained experience, with all tasks performed within a user-friendly, Gradio-based web interface hosted on Hugging Face Spaces.

\subsection{Welcome and Authentication}
\label{app:W1}

\begin{figure}[h]
    \centering
    \includegraphics[width=0.85\textwidth]{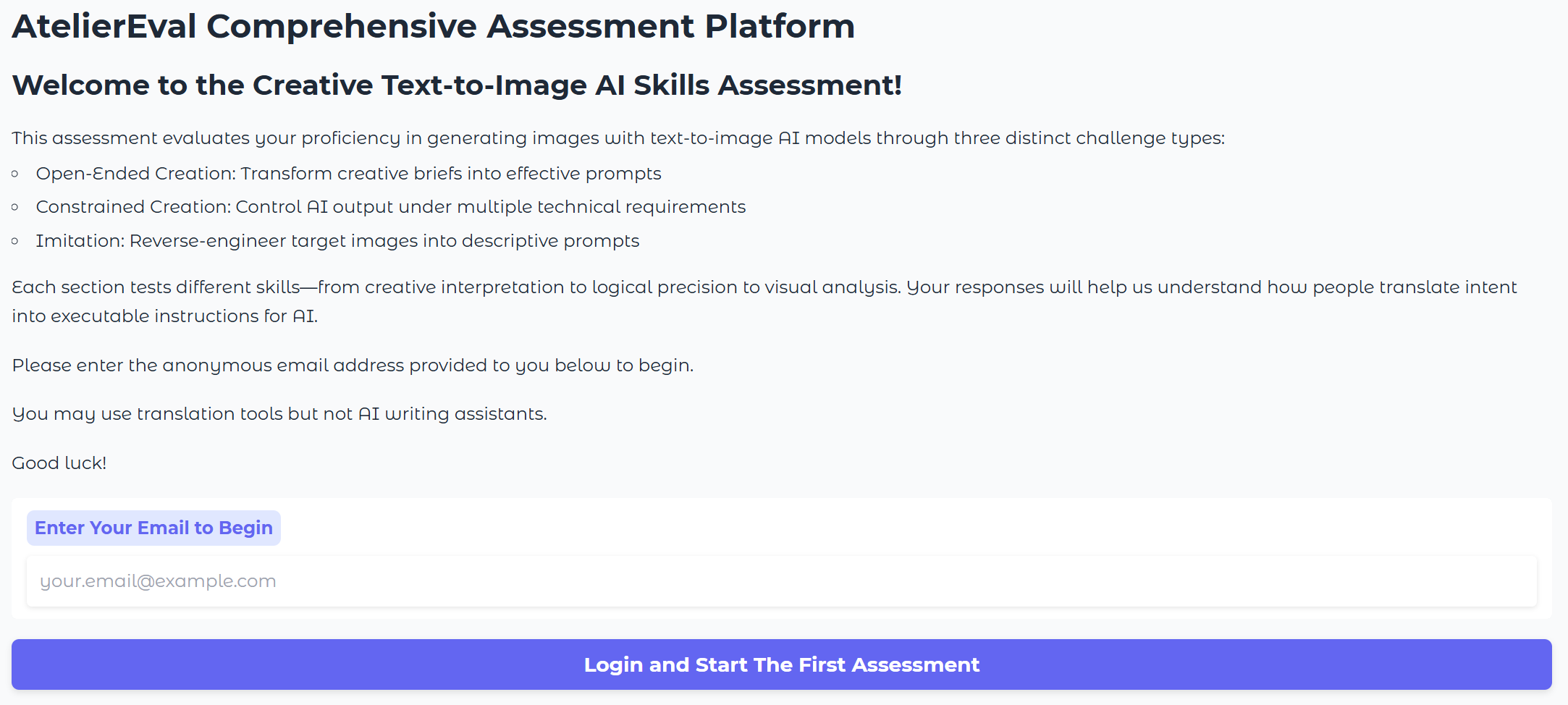}
    \caption{The welcome and login interface. Participants enter their anonymous email address and receive a clear overview of the three assessment modules before beginning.}
    \label{fig:welcome_page}
\end{figure}

Participants begin by accessing the assessment platform via a provided Hugging Face Space link using their assigned anonymous account credentials. Upon loading, they are greeted with a welcome screen that provides a concise overview of the assessment's structure and objectives, as shown in Figure~\ref{fig:welcome_page}. 

The interface explains that the assessment evaluates three distinct skill dimensions: (1) \textit{Open-Ended Creation} for testing creative interpretation and problem-solving, (2) \textit{Constrained Creation} for evaluating logical precision under multiple technical requirements, and (3) \textit{Imitation} for measuring visual analysis and reverse-engineering capabilities. 

Participants are instructed to enter their assigned anonymous email address into the text field and click the ``Login and Start Assessment'' button to proceed. This authentication step ensures proper data association while maintaining participant anonymity throughout the study. The clean, welcoming design with clear instructions minimizes cognitive load and technical barriers, allowing participants to focus on the prompting tasks themselves.

\subsection{Assessment Structure and Randomization}
\label{app:W2}
After successful authentication, participants are informed that they will complete \textbf{two independent assessment rounds}. Each round consists of 15 prompting tasks distributed across the three task types (5 tasks per type). Participants may take short breaks between rounds. To ensure robust evaluation and prevent order effects, the system implements two levels of randomization:

\begin{itemize}
    \item \textbf{Task Type Order Randomization:} Within each round, the three task types (Open-Ended, Constrained, Imitation) are presented in a randomized sequence. This means a participant might encounter Constrained tasks first in Round 1 but Imitation tasks first in Round 2.
    
    \item \textbf{Task Instance Randomization:} For each task type, the 5 specific task instances are randomly sampled from a curated pool of 360 validated problems (120 per task type). This ensures that no two participants receive identical task sets, while maintaining consistent difficulty and construct validity across all instances.
\end{itemize}

This randomization protocol serves dual purposes:

\begin{enumerate}
    \item \textbf{Methodological rigor:} Controls for learning effects and position bias in our experimental design.
    \item \textbf{Ecological realism:} Simulates the unpredictable variety of real-world creative demands that professional prompters encounter in practice.
\end{enumerate}

\subsection{Task Type Interfaces}
\label{app:W3}

Having authenticated and understood the overall assessment structure, participants proceed to the core evaluation component. The following subsections describe the three task types that participants encounter in randomized order. Each task type begins with a dedicated introduction screen explaining its specific objectives and evaluation criteria, followed by the task interface where participants construct their prompts. Navigation buttons (``Previous'' and ``Next'') allow participants to review and modify their responses or adjust task order if needed.

\subsubsection{Open-Ended Creation Module}

\begin{figure}[h]
    \centering
    \includegraphics[width=0.85\textwidth]{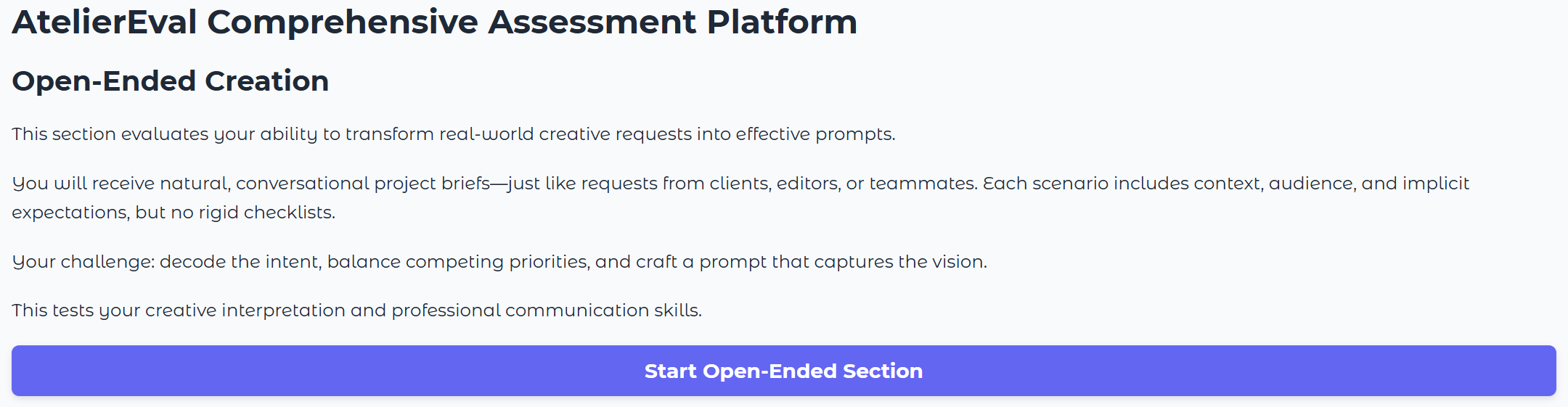}
    \caption{Introduction screen for the Open-Ended Creation module, explaining the conversational nature of creative briefs and the assessment objectives.}
    \label{fig:oe_page}
\end{figure}

When participants first encounter an Open-Ended Creation block, they are presented with an introductory screen (Figure~\ref{fig:oe_page}) that explains the purpose and nature of these tasks. The interface informs participants that they will receive natural, conversational project briefs similar to real-world creative requests from clients, editors, or teammates. Each scenario includes context, audience, and implicit expectations, but deliberately avoids rigid checklists to test the participant's ability to independently interpret requirements and translate abstract concepts into concrete visual specifications.

The screen emphasizes that this task type evaluates \textit{creative interpretation and professional communication skills}---the ability to decode intent, balance competing priorities, and craft prompts that capture the envisioned aesthetic without explicit guidance.

\begin{figure}[h]
    \centering
    \includegraphics[width=0.85\textwidth]{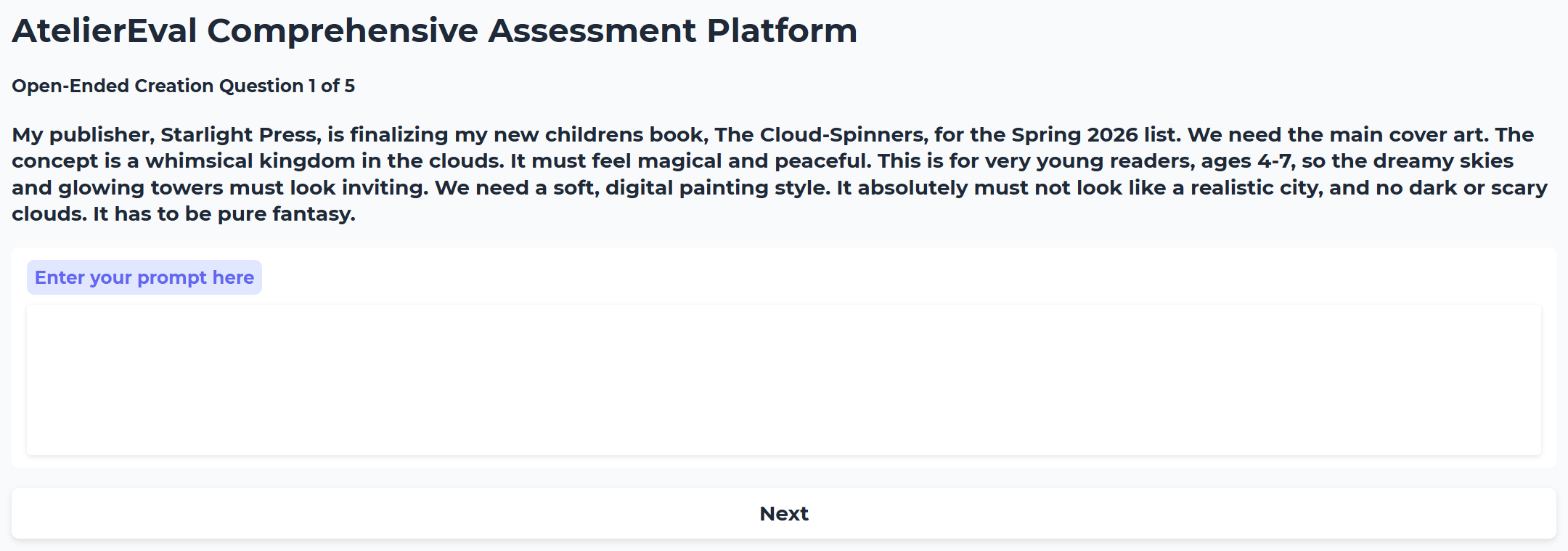}
    \caption{Example task interface for Open-Ended Creation. The scenario provides rich contextual information that needs to be translated into an effective prompt.}
    \label{fig:oe_task}
\end{figure}

For each task within this module, participants are presented with a detailed creative scenario. Figure~\ref{fig:oe_task} shows a representative example where a participant needs to create cover art for a children's book. The task description provides rich context including the target audience (ages 4--7), thematic requirements (whimsical, magical, peaceful), stylistic constraints (soft digital painting, not photorealistic), and technical specifications (no dark or scary clouds). 

Participants need to synthesize this multifaceted information into a single, coherent prompt entered in the text box labeled ``Enter your prompt here.'' Once satisfied with their response, participants click ``Next'' to proceed to the subsequent task.

\subsubsection{Constrained Creation Module}

\begin{figure}[h]
    \centering
    \includegraphics[width=0.85\textwidth]{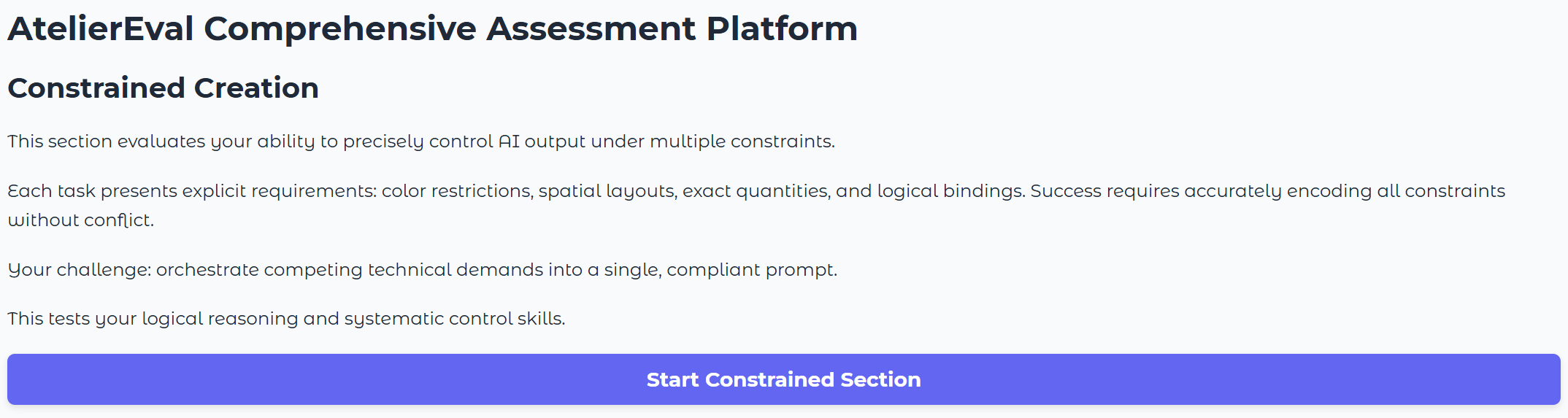}
    \caption{Introduction screen for the Constrained Creation module, highlighting the focus on technical precision and constraint satisfaction.}
    \label{fig:co_page}
\end{figure}

When participants enter a Constrained Creation block, the introductory screen (Figure~\ref{fig:co_page}) explains that this module evaluates their ability to precisely control AI output under multiple simultaneous constraints. Unlike the open-ended tasks, success here requires accurately encoding all explicit requirements---including color restrictions, spatial layouts, exact quantities, and logical bindings---without introducing conflicts or omissions.

The interface emphasizes that this module tests \textit{logical reasoning and systematic control skills}, challenging participants to orchestrate competing technical demands into a single, compliant prompt.

\begin{figure}[h]
    \centering
    \includegraphics[width=0.85\textwidth]{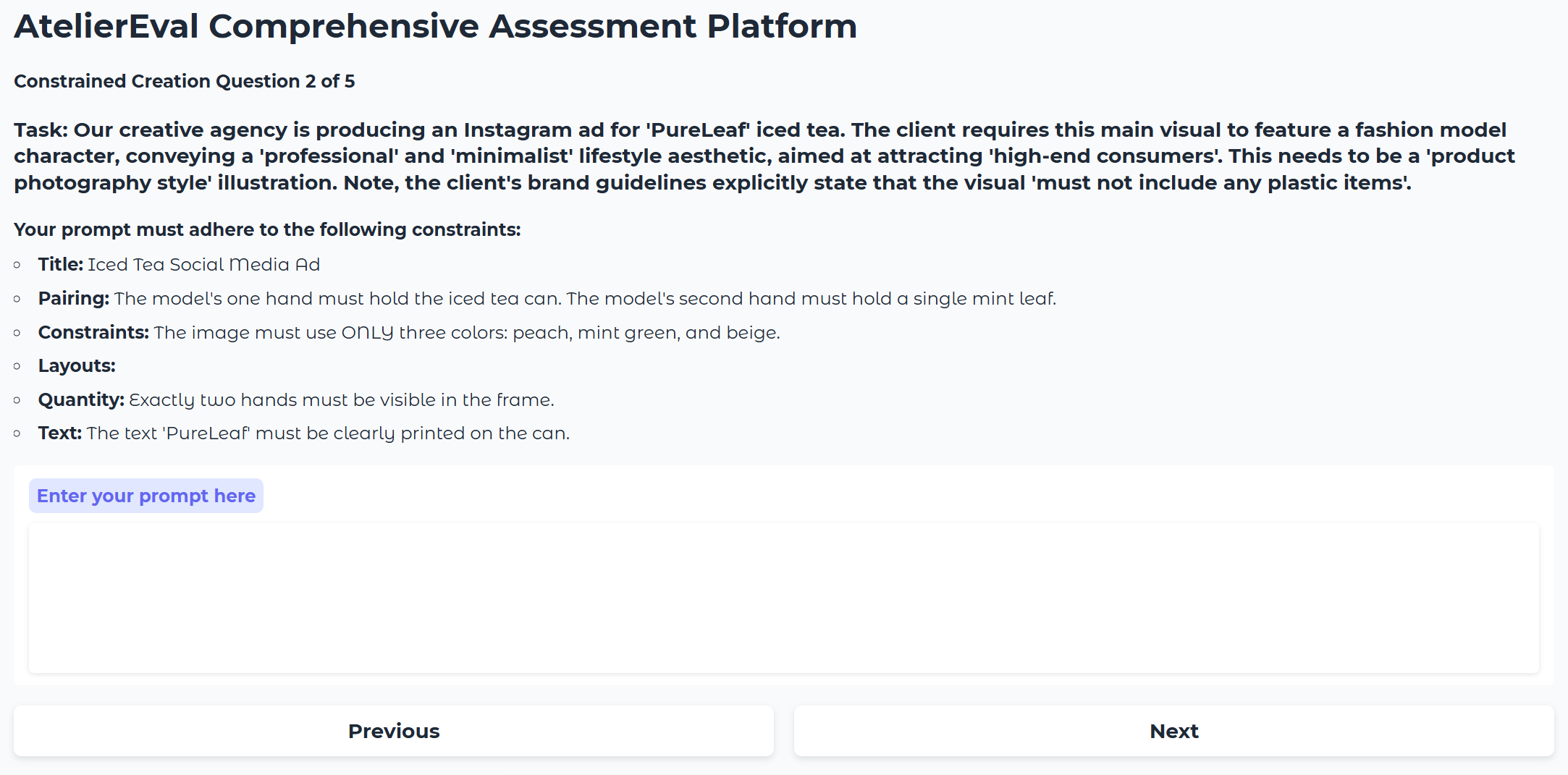}
    \caption{Example task interface for Constrained Creation. The bulleted list explicitly enumerates all requirements, testing the participant's ability to encode multiple constraints into a single compliant prompt.}
    \label{fig:co_task}
\end{figure}

Each Constrained Creation task presents a detailed list of explicit requirements in a structured, bulleted format. Figure~\ref{fig:co_task} illustrates a representative example: creating an Instagram advertisement for iced tea. The task specifies:

\begin{itemize}
    \item \textbf{Title:} Product name must appear as text overlay
    \item \textbf{Pairing:} Specific hand positions for two objects
    \item \textbf{Constraints:} Restricted color palette (peach, mint green, beige only)
    \item \textbf{Layout:} Product photography style illustration
    \item \textbf{Quantity:} Exact number of visible hands (two)
    \item \textbf{Text:} Brand name must be clearly printed on the can
    \item \textbf{Prohibitions:} No plastic items allowed
\end{itemize}

Participants need to construct a prompt that satisfies all constraints simultaneously without internal contradictions. The task description is displayed at the top, followed by the structured constraint list, with the prompt input area below.

\subsubsection{Imitation and Reproduction Module}

\begin{figure}[h]
    \centering
    \includegraphics[width=0.85\textwidth]{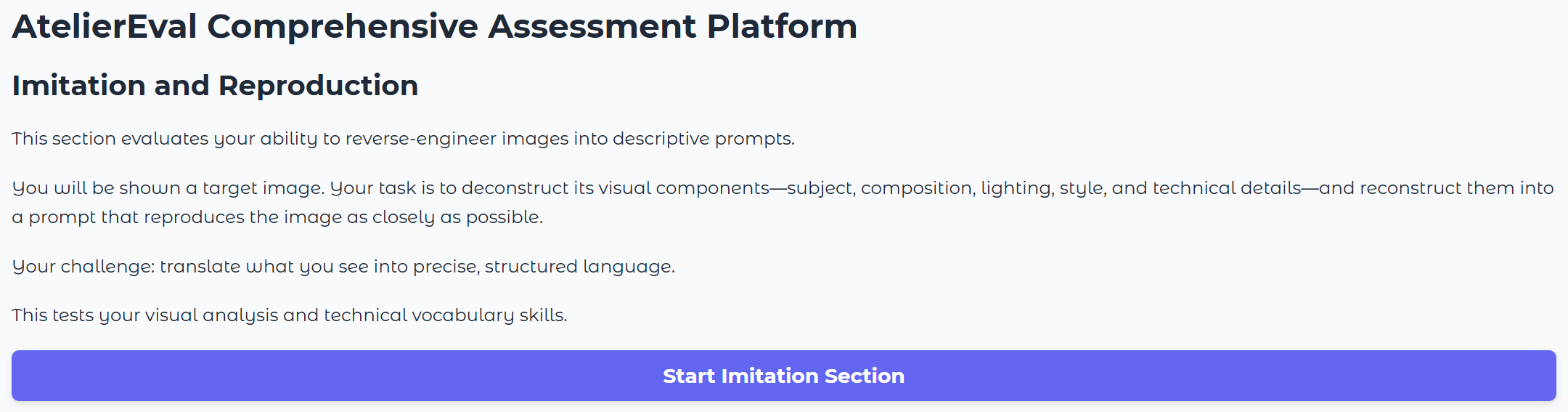}
    \caption{Introduction screen for the Imitation and Reproduction module, emphasizing the reverse-engineering challenge.}
    \label{fig:im_page}
\end{figure}

When participants encounter the Imitation module, the introductory screen (Figure~\ref{fig:im_page}) explains that they will be shown target images and need to deconstruct their visual components---including subject matter, composition, lighting, style, and technical details---into descriptive prompts that can reproduce the images as closely as possible.

This module tests \textit{visual analysis and technical vocabulary skills}, requiring participants to translate what they observe into precise, structured language that captures both obvious elements and subtle stylistic nuances.

\begin{figure}[h]
    \centering
    \includegraphics[width=0.85\textwidth]{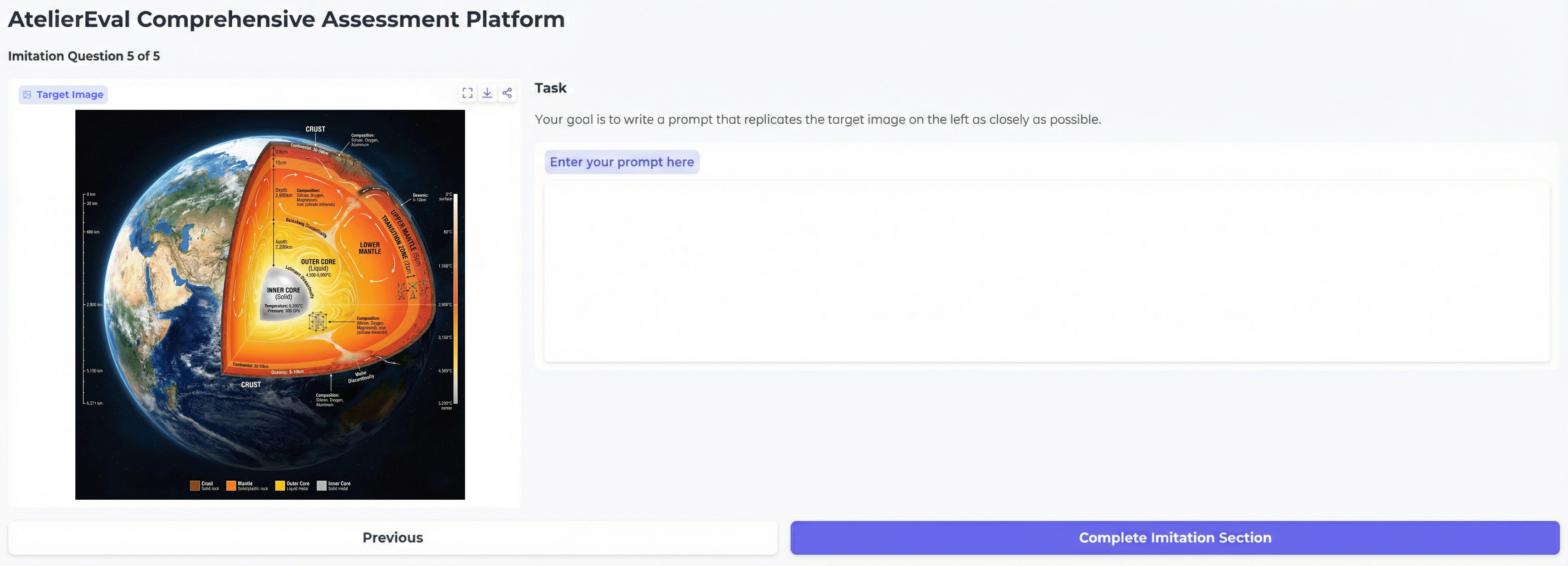}
    \caption{Example task interface for Imitation. The side-by-side layout presents the target image (left) alongside the prompt input area (right), enabling continuous visual reference during prompt construction.}
    \label{fig:im_task}
\end{figure}

For each Imitation task, the interface adopts a side-by-side layout (Figure~\ref{fig:im_task}). The target image is displayed prominently on the left side of the screen, while the prompt input area occupies the right side. This parallel presentation allows participants to continuously reference the target image while constructing their descriptive prompt, facilitating iterative refinement of their textual description.

The example shown in Figure~\ref{fig:im_task} presents a cross-sectional diagram of Earth's interior structure. Participants need to identify and describe not only the obvious elements (planetary sphere, labeled layers) but also technical details such as the cutaway visualization style, color-coding scheme, text annotations, and scientific illustration aesthetic. The task description simply states: ``Your goal is to write a prompt that replicates the target image on the left as closely as possible,'' providing no additional hints about which features to prioritize.

\subsection{Workflow Integration and Data Collection}
\label{app:W4}
Throughout the entire assessment, the system automatically records all participant interactions, including:

\begin{itemize}
    \item Submitted prompt text for each task
    \item Task type and instance identifier for each response
    \item Task progression order (reflecting the randomized sequence)
    \item Round completion status (Round 1 vs. Round 2)
    \item Operational logs including interface interactions
\end{itemize}

The two-round structure serves multiple purposes: it increases the sample size per participant for more robust statistical analysis, allows examination of within-subject consistency, and provides sufficient task diversity through the randomization mechanism. Each participant ultimately completes 30 tasks in total (15 tasks × 2 rounds), with 10 responses per task type collected across both rounds.

Upon completing all tasks, participants upload their mandatory screen recording to a secure cloud storage location (Google Drive or Baidu Netdisk). The screen recording is strictly for verification purposes and will be permanently deleted after the verification process is complete.

This self-contained, browser-based workflow ensures a standardized experience across all participants while maintaining \textbf{high ecological validity} through multiple design principles:

\begin{enumerate}
    \item \textbf{Realistic task scenarios:} All tasks are grounded in authentic professional use cases, from creative briefs to technical specifications to visual reproduction challenges.
    \item \textbf{Professional-grade interface:} The Gradio-based UI mirrors industry-standard text-to-image platforms, ensuring participants interact with familiar design patterns and workflows.
    \item \textbf{Authentic cognitive demands:} Time constraints, task complexity, and the need for multifaceted decision-making reflect real-world prompting scenarios.
    \item \textbf{Naturalistic interaction patterns:} Participants construct prompts without artificial restrictions, using their own vocabulary, style, and problem-solving approaches.
\end{enumerate}

The combination of randomized task presentation and comprehensive data collection enables rigorous assessment of prompting proficiency across multiple skill dimensions while controlling for potential confounds such as learning effects and order bias. This design allows our findings to generalize effectively to real-world text-to-image prompting contexts.

\section{Computational Resource Consumption}
\label{app:resource}

\paragraph{Benchmarking Consumption.}
The primary resource consumption stems from API invocations for three commercial backends (Gemini-3-Pro-Image-Preview, GPT-Image-1-All, and Flux.1 Pro), generating 7,200 images each, alongside MLLM overheads for prompting and evaluation. Given shared institutional access and pricing variability, we report usage via query volume rather than monetary expenditure. Locally, SDXL also generated 7,200 images on a workstation with NVIDIA RTX 5080 (16GB). Precise GPU hours are not isolated due to the mixed-workload nature of the deployment environment.

\paragraph{Evaluator Consumption.}
To assess evaluator efficiency, we conducted a controlled experiment using GPT-5.2 on 30 sampled tasks involving full objective and subjective cycles. Based on official pricing, the zero-shot QA/VQA baseline incurred \$0.35 versus \$0.89 for AtelierJudge—a margin we consider justifiable given the substantial reliability gains. Additionally, owing to engineering optimizations, the embedding calculation for each T2I-prompter pair requires less than 5 minutes on our local workstation, rendering this overhead negligible compared to the T2I generation costs.

\section{Detailed Ethical Considerations and Procedures}
\label{app:ethics}

This section summarizes the ethical considerations and related procedures of the user study of our work. The full informed consent form and all questionnaires are provided in Appendix~\ref{apx:human_study_materials}.

\subsection{Ethics Approval and Consent}
\label{app:Y1}
This study was reviewed and approved by the Institutional Review Board (IRB) at the first author's institution. Prior to any experimental tasks, all participants were provided with a comprehensive Informed Consent Form (Appendix \ref{app:consent}). This document detailed the research objectives, experimental procedures, potential risks, and compensation details. Participants were required to explicitly indicate their agreement to proceed. Participants were informed of their right to withdraw from the experiment at any stage without penalty. To ensure the capacity for informed consent, all participants were required to be at least 18 years of age.

\subsection{Participant Anonymity}
\label{apx:confidentiality}

This subsection outlines the data protection protocols enforced to guarantee participant confidentiality and data integrity throughout the user study component of our work.

\paragraph{Anonymity in Tests.}
The core strategy for maintaining anonymity was the utilization of pre-configured, anonymous Hugging Face accounts. The research team assigned a unique, anonymous account credential to each participant. These credentials were securely transmitted to participants via email before the session began, and users were required to use these credentials to access the tests. By adhering to this protocol, we ensured that all interactions within the Gradio-based UI in Hugging Face Space were completely decoupled from the participants' real-world identities.

\paragraph{Anonymity in Tool Usage.}
To further safeguard participant privacy, we implemented strict protocols for tool deployment. For VPN, we used Clash, an open-source VPN tool. Only generic config files were distributed without any user accounts or credentials. For DeepL, participants were required to utilize the non-registered version. For screen recording, participants were instructed to avoid displaying Personally Identifiable Information (PII) during the session. All recordings are subject to unified post-hoc deletion immediately after the validity verification of the submissions.

\paragraph{Participant Information Verification.}
To maintain the consistency of the skilled group and to validate self-assessments provided in the pre-test questionnaire (Appendix~\ref{apx:pre-test}), we conducted a limited eligibility verification for participants who self-identified as skilled users. This verification primarily involved reviewing publicly available portfolios or example works voluntarily provided by participants (e.g., prompts and generated images shared on content platforms). In a small number of cases, participants were invited to participate in a brief online discussion (with cameras turned off). All verification steps were performed solely to confirm eligibility for the skilled group. Any personal information accessed during this process was handled confidentially and was not retained beyond the verification stage.

\paragraph{De-identification and Aggregation.}
Following the conclusion of data collection, a comprehensive de-identification procedure was executed. For analysis purposes, only the anonymous Hugging Face IDs were retained. Identifying information was stored separately and never linked in the released dataset. Sensitive contact information required for compensation was isolated in encrypted storage with restricted access. The final research data and questionnaire responses were linked exclusively via anonymous IDs. All findings presented in this paper and any open-source data are fully anonymized and aggregated, ensuring that no individual participant can be re-identified.

\end{document}